\documentclass[examplefnt,biber]{nowfnt}

\usepackage[utf8]{inputenc}
\usepackage{amsmath}
\usepackage[capitalise]{cleveref}
\usepackage{algorithm, algorithmic}
\usepackage{mathtools}
\usepackage{bm,amsmath,amsfonts,amssymb}
\usepackage{tcolorbox}
\usepackage{booktabs}
\usepackage{hyperref}
\usepackage{amsmath}
\usepackage{graphicx}
\usepackage{wrapfig}
\usepackage{amssymb}
\usepackage{stackengine,xcolor}
\usepackage{caption}
\usepackage{subcaption}
\usepackage{tabularx}
\usepackage{mwe}

\DeclareMathOperator*{\argmax}{arg\,max}
\DeclareMathOperator*{\argmin}{arg\,min}

\DeclareMathOperator{\T}{\mathcal{T}}
\DeclareMathOperator{\D}{\mathcal{D}}

\title{An Introduction to Lifelong Supervised Learning}

\maintitleauthorlist{
Shagun Sodhani \\
FAIR, Meta AI \\
\and
Mojtaba Faramarzi \\
Universite de Montreal \\
\and
Sanket Vaibhav Mehta \\
Carnegie Mellon University \\
\and
Pranshu Malviya \\
Polytechnique Montréal \\
\and
Mohamed Abdelsalam \\
Universite de Montreal \\
\and
Janarthanan Rajendran \\
Universite de Montreal \\
\and
Sarath Chandar \\
Polytechnique Montréal \\
Canada CIFAR AI Chair \\
Quebec Artificial Intelligence Institute (Mila) \\
}

\makeatletter
\def\alignedspace@left{\null\,}
\makeatother

\DeclareUnicodeCharacter{0301}{\'{e}}

\begin{document}

\chapter{Introduction}

\section{Artificial Intelligence Systems}

\paragraph{Artificial Intelligence} Artificial Intelligence (AI) systems can be defined as systems that think and act rationally like humans~\citep{bellman1978introduction, kurzweil1990age, schalkoff1991artificial, rich1992artificial, winston1992artificial, haugeland1997mind, russell2005ai}. While the term was formally coined at the famous Dartmouth conference in $1956$~\citep{mccarthy2006proposal,Woo.2014}, philosophers dating back to Aristotle and Plato contemplated formulating the law governing the rational part of the mind. The idea of creating~\textit{intelligent} systems inspired myths like the story of Talos, a giant bronze robot created by gods that carried within it a mysterious life source and guarded the island of Crete~\citep{Shashkevich.2019}. Since then, psychologists, behaviorists, cognitive scientists, linguists, and computer scientists have championed various approaches for understanding intelligence and developing AI systems.

Early AI systems were often~\textit{rule-based}: given a collection of~\textit{rules of the world}, they would use approaches like search and symbol manipulation to solve a given task. These systems focused on (and generally performed well) on reasoning-related problems, like proving theorems (e.g., Logic Theorist~\citep{gugerty2006newell} and General Problem Solver~\citep{newell1959variety}) or focused on setups with few entities to interact with~\citep{minsky1972artificial}. Systems relying on these classical approaches enabled significant breakthroughs like IBM's Deep Blue system defeating the then world champion of Chess in 1997. However, these systems were often limited by how fast they could ~\textit{process} the rules. As a result, these systems do not work well when the number of combinations of rules becomes large. Another significant limitation of the rule-based systems is that they need a clean and well-curated collection of rules to start with. It is possible that one can define and describe these rules explicitly for a game like Chess, but this is often not feasible in real-life scenarios.

\paragraph{Machine Learning}
The over-reliance of early AI systems on hard-coded knowledge limited their scope and use for complex setups and real-world applications. Machine Learning (ML) is a sub-field of AI that aims to address this limitation by inferring knowledge from raw data using techniques like pattern mining, association rule mining, representation learning, classification, regression, etc. Machine Learning systems can be broadly categorized into two groups:

\begin{enumerate}
    \item~\textbf{Parametric models} are models that ``summarize'' (or encode) the knowledge in the given dataset/task\footnote{For simplicity, we use the terms dataset and task interchangeably in the introduction} using a set of parameters. These models generally assume that a function exists that explains the knowledge in the data and infer the parameters of that function. Once the parameters have been learned, the original data is no longer needed. Common examples of parametric models include logistic regression, linear discriminant analysis, and neural networks.
    \item~\textbf{Non-parametric models} do not infer any parameters from the given data, though they may infer some summary statistics, like mean, to speed up inference. Common examples of parametric models include k-Nearest Neighbors and Support Vector Machines. 
\end{enumerate}

In general, a machine learning system may or may not have to~\textit{learn} feature representations for a given dataset. For example, consider an email spam classifier where the input to the system is a set of features like ``is the email from an unknown user'' or if certain keywords are present or not. In this case, these features could be fed as input to a logistic regression classifier, and only the classifier needs to be trained. In general, we can not assume access to high-quality, informative features, and the machine learning system has to infer these features. For example, in the email spam classifier example, the system may only have access to the blob of email text. It would need to learn a good feature representation that can be used as input to the classifier. In this case, the system could use a non-parametric approach like ~\emph{term frequency-inverse document frequency} (TF-IDF)~\citep{ramos2003using} or use a parametric representation learning model like a recurrent neural network~\citep{hochreiter1997long,cho2014learning}

\paragraph{Deep Learning} Deep Learning is a sub-field within machine learning that focuses on representation learning (learning representation from the given data), usually using parametric models. The high-level idea behind deep learning is as follows: There are some base computational units called layers, like the convolutional neural network layer~\citep{lecun1989backpropagation}, which can be stacked over each other (or, in general, composed arbitrarily) to create powerful architectures. For example, the ResNet architecture is composed using a stack of convolutional layers, along with other layers like max-pooling layers.

As the feature representation passes through the subsequent layers, it is transformed into more complex features. The resulting feature could be used as input to a classifier system. The entire system, i.e., the representation learning system, and the classifier system, can be trained together end-to-end. Today, machine learning is one of the most popular AI paradigms, and deep learning is the most popular representation learning approach. It is worth noting that the current AI systems are often a combination of techniques from different sub-fields. For example, AlphaGo~\citep{silver2016mastering}, which defeated the world champion of Go, uses convolution networks, a deep learning approach, to learn feature representation, and Monte-Carlo Tree Search, a traditional AI approach, to search for the next action.

\section{Success Stories of Machine Learning}

Machine Learning Systems have come a long way since the McCulloch-Pitts Neuron, the first computational model of a neuron~\citep{mcculloch1943logical}. ML systems have shown impressive results in a number of problem settings where the previous AI approaches struggled: fundamental sciences~\citep{gemp2021eigengame, PhysRevResearch.2.033429, bapst2020unveiling}, bio-medicine~\citep{cirecsan2013mitosis, litjens2016deep}, life-sciences~\citep{senior2020improved, yim2020predicting, tomavsev2019clinically, leibo2018psychlab}, hardware design and manufacturing~\citep{schmidt2019recent, bhuvaneswari2021deep, mirhoseini2021graph}, graph analysis~\citep{10.1145/2736277.2741093, kipf2016semi, NIPS2017_5dd9db5e}, neuroscience~\citep{mathis2018deeplabcut, mathis2020deep} etc. 

Even for domains where traditional AI systems were used earlier, the current generation of ML systems have led to significant improvements. This includes areas such as image understanding~\citep{Krizhevsky2012-imagenet-classification-with-deep-convolutional-neural-networks, xie2017aggregated}, semantic segmentation~\citep{girshick2014rich, ren2015faster}, video processing~\citep{fan2021multiscale}, machine translation~\citep{bahdanau2014neural, cho2014learning}, question answering~\citep{lan2019albert, zhang2020retrospective}, text summarization~\citep{raffel2019exploring, lewis2019bart}, text generation~\citep{radford2019language_models_are_unsupervised_multitask_learners, kaplan2020scaling}, speech recognition~\citep{schneider2019wav2vec, baevski2020wav2vec}, textless NLP~\citep{lakhotia2021generative, kharitonov2021text,Polyak2021SpeechRF}, 

robotics~\citep{hadsell2008deep, koutnik2013evolving, chen2015deepdriving}, social network analysis~\citep{sodhani2019attending, tang2021graphbased}, 

etc. ML systems have reached super-human performance on several tasks~\citep{hochreiter1997long, bahdanau2014neural, graves2014neural, mnih2015human, he2016deep, miller2016key, vaswani2017attention, krizhevsky2017imagenet, silver2017mastering, silver2018general, devlin2018bert, vinyals2019grandmaster, zhang2020resnest, brown2020language, schrittwieser2020mastering, badia2020agent57}. These ML systems were already used in the digital world~\citep{Lewis-Kraus.2016, zhai2017visual, naumov2019deep} but are now being actively deployed in the physical world as well~\citep{Satariano.2020, Vincent.2021, Davies.2021}.

These advances are bringing the current generation of AI systems closer to the long-standing goal of AI practitioners - designing systems that can \textit{imitate} the behavior of humans or can demonstrate human-like general intelligence~\citep{10.1093/mind/LIX.236.433}. However, despite all the success and promising results, there are still significant gaps in the capabilities of even the most powerful AI systems when compared to humans. 

\section{Lifelong Learning Systems}

A key criticism of the current machine learning systems is that they tend to be~\textit{data-hungry}~\citep{marcus2018deep, ford2018architects}. Take the example of the~\textit{GPT-3} model~\citep{brown2020language}, a large scale language model that is trained with $300$B tokens from text data sources like Common Crawl corpus~\citep{raffel2019exploring} ($570$ GB of data after filtering and cleaning), WebText~\citep{radford2019language_models_are_unsupervised_multitask_learners}, two internet-based book corpora and Wikipedia pages. The datasets had to be curated and processed to provide meaningful learning signals to the training models. While recent advances in self-supervised learning have reduced the dependence on large-scale, clean and well-labeled datasets, we still need to account for the time and cost of pre-training large-scale models. For instance, the~\textit{GPT-3} model used compute equivalent to $3.14e^{23}$ flops\footnote{floating point operations} and it would take $355$ years to train GPT-3 on a single NVIDIA Tesla V100 GPU. The sample efficiency of ML systems significantly lags behind that of humans, making them expensive to develop and deploy.

A second key challenge is that standard AI paradigms are not good at transferring (or leveraging) knowledge across tasks. While it is possible to train systems that provide excellent performance on a specific task (or related distribution of tasks, in the case of multi-task learning), it is much harder to train general-purpose AI systems that can perform a diverse set of tasks. When AI systems are trained over a sequence of tasks, they tend to~\textit{forgets} the crucial knowledge they acquired from the previous tasks. This phenomenon is often referred to as catastrophic forgetting~\citep{mccloskey1989catastrophic,ratcliff1990connectionist} and affects all parametric AI systems. Sometimes, knowledge transfer even hurts the performance on the current task due to~\textit{negative interference} (a common challenge for multi-task learning) of knowledge across tasks~\citep{standley2020tasks, yu2020gradient, mansilla2021domain,chen2018gradnorm}. Even in the case of paradigms like transfer learning (which specifically emphasizes the transfer of knowledge across tasks), the knowledge transfer is often uni-directional, i.e., the knowledge from the previous tasks is used to improve the performance on the current task (and not all the tasks). The emphasis is on improving the performance of the current task, even if that hurts the performance of the previous tasks. In an ideal world, we would want the learning systems to perform both~\textit{forward} (training on the current task improves the performance on the future tasks) as well as~\textit{backward} transfer of knowledge (training on the current task improves the performance on the previous tasks).

These two challenges are related. AI systems need a lot of data to train on because they start training on every task from~\textit{scratch}. Imagine a system that has to learn the alphabet every time it reads a book. Such a system would have a poor sample complexity because it cannot transfer knowledge across tasks (of learning alphabets and reading books). In terms of learning strategy, the current AI systems are closer to this hypothetical system than humans. As the new data becomes available, the AI systems can not~\textit{incrementally} acquire new knowledge (without forgetting the prior knowledge). These challenges also make the AI systems harder to adapt to new tasks/datasets. Since these systems do not effectively transfer knowledge across tasks, they need a lot of data to adapt to the new task when they encounter a new task. These behaviors are in sharp contrast to how humans learn and behave. Humans do not need to \textit{train} over a stationary data distribution for multiple epochs. While they do not have perfect memory, they can incrementally acquire and update knowledge over their lifetime without catastrophically forgetting the knowledge relevant for the previous tasks. Moreover, humans can efficiently leverage experience across tasks and exhibit knowledge transfer to improve performance on new (forward transfer) and previous (backward transfer) tasks. Over time, humans learn how to quickly adapt to novel situations without learning everything from scratch. 

The~\textit{Lifelong Learning} paradigm is the branch of AI that focuses on developing lifelong learning systems - systems that keep accumulating new knowledge throughout their lifetime without forgetting the prior knowledge and use this accumulated knowledge to improve their performance on the different tasks. We highlight that the lifelong learning paradigm is not unique to the multi-task setup and applies to the single-task setup as well. Lifelong learning is a general setup since it makes fewer assumptions about the task (or tasks). Consider a standard single-task supervised learning setup where the learner can access the entire dataset before starting the training. In this case, the learner can perform multiple epochs over the dataset, shuffling the data in each epoch to keep the data distribution, i.i.d (independent and identically distributed). However, there are many implicit assumptions in this setup - since we have access to the dataset beforehand, we know how many unique classes exist in the dataset. We also have access to the class distribution and can weigh the classes differently. We can also over/under-sample the data. While these assumptions make the setup amenable for training, they also take the setup away from the more general open-ended learning setup. If we were not to assume access to the dataset (or even the number of unique classes), the AI system would have to address challenges like modifying the network architecture as it sees new classes, not forgetting the old data points as it trains on new data points and potentially increasing the capacity of the system as new data keeps coming in. All these challenges are studied under the paradigm of lifelong learning. 

\section{Outline}

This primer is an attempt to provide a detailed summary of the different facets of lifelong learning. We start with~\cref{sec::overview} which provides a high-level overview of lifelong learning systems. In this chapter, we discuss prominent scenarios in lifelong learning (\cref{sec:prominent_scenarios}), provide a high-level organization of different lifelong learning approaches (\cref{sec:lifeong_learning_strategies}), enumerate the desiderata for an ideal lifelong learning system (\cref{subsec:desidrata}), discuss how lifelong learning is related to other learning paradigms (\cref{sec:relation_to_other_areas}), describe common metrics used to evaluate lifelong learning systems (\cref{sec:metrics}).   This chapter is more useful for readers who are new to lifelong learning and want to get introduced to the field without focusing on specific approaches or benchmarks. 

The remaining chapters focus on specific aspects (either learning algorithms or benchmarks) and are more useful for readers who are looking for specific approaches or benchmarks.~\cref{sec:regularization} focuses on regularization-based approaches that do not assume access to any data from previous tasks.~\cref{sec:memory} discusses memory-based approaches that typically use a~\textit{replay buffer} or an \textit{episodic memory} to save subset of data across different tasks.~\cref{sec:architecture} focuses on different architecture families (and their instantiations) that have been proposed for training lifelong learning systems. Following these different classes of learning algorithms, we discuss the commonly used evaluation benchmarks and metrics for lifelong learning (\cref{sec_metrics_benchmarks}) and wrap up with a discussion of future challenges and important research directions in~\cref{sec:future_challenges}. 

\section{Scope}
\label{sec::scope}
The primer is designed to serve as an introduction to lifelong learning paradigm and address questions like ``what is lifelong learning'', ``why is it a relevant problem to work on'', ``what are some key desiderata of a lifelong learning system'', ``what are some common design decisions when developing lifelong learning system'', ``what are commonly used benchmarks in lifelong learning'' etc. While we include (and describe) several lifelong learning approaches and benchmarks and intend to keep the document updated over time, the primer is not an exhaustive literature survey by any means. The selection of work is based on the diversity of approaches and pedagogical reasons. We note that we are focusing on lifelong learning approaches in the context of supervised learning and do not cover work-related to lifelong reinforcement learning, which is an important and interesting topic on its own. We recommend the readers to refer~\citet{khetarpal2020towards} for a survey on lifelong reinforcement learning.

\section{Target Audience}

The target audience for this primer is both newcomers (people who are new to the field of lifelong learning or are just curious about lifelong learning) and practitioners (who are working on lifelong learning or related areas like meta-learning, transfer learning, multi-task learning, etc.). It should be useful for people across the spectrum - from researchers working on the fundamental problems to practitioners working on applications of ML.~\cref{sec::overview} is particularly useful for readers who are new to the area of lifelong learning. Readers already familiar with lifelong learning may benefit more from~\cref{sec:regularization},~\cref{sec:memory} and~\cref{sec:architecture} that focus on different classes of lifelong learning algorithms. Readers looking to evaluate their lifelong learning systems or create new evaluation benchmarks would benefit from a discussion on benchmarks and metrics~(\cref{sec_metrics_benchmarks}).

\chapter{Overview of Lifelong Learning}
\label{sec::overview}

\section{What is Lifelong Learning}
\label{sec::what_is_lifelong_learning}

Consider a setup where a machine learning model is trained over a sequence of tasks. Let us assume that the model has trained on the first $k$ tasks and is starting to train on the $k+1^{th}$ task. As the model trains on the $k+1^{th}$ task, a couple of scenarios are possible: (i) the model learns to solve the current task at the expense of performance on the previous tasks, (ii) the model fails to learn the new tasks though it retains its performance on the previous tasks, (iii) the model learns the new tasks while retaining its performance on the previous tasks, or (iv) the model does not learn the new task while forgetting its knowledge on the previous task. While the ideal outcome is the one where the model learns the new tasks while retaining its performance on the previous tasks, in practice, the model would likely forget some of the previous knowledge and may not be able to learn the new task. 

This setup can be viewed from the lens of~\textbf{stability-plasticity dilemma} \citep{mermillod2013stability}. Here,~\textit{plasticity} refers to the ability to integrate new knowledge, and ~\textit{stability} refers to the ability to retain previous knowledge~\citep{mirzadeh2020dropout}. Too much plasticity will likely lead to forgetting previous knowledge, while too much stability will hurt learning on the current task. Any learning system, biological or artificial, needs to balance plasticity with stability to ensure continued learning without catastrophic forgetting.

Much work in machine learning looks at the~\textit{stability-plasticity dilemma} as two separate problems and puts more emphasis on one of the two aspects. For example, transfer learning approaches focus exclusively on the plasticity aspect, while approaches to alleviate catastrophic forgetting focus more on the stability aspect. The~\textit{Lifelong Learning} paradigm focuses on both the challenges at once, with the goal of developing lifelong learning systems - systems that keep accumulating new knowledge throughout their lifetime (plasticity) without catastrophically forgetting the prior knowledge (stability) and use this accumulated knowledge to improve their performance on the different tasks.

As discussed in~\cref{sec::scope}, in this primer, we focus on lifelong learning paradigm in context of supervised learning. We briefly recap the supervised learning setup~(\cref{sec:supervised_learning}), describe the lifelong supervised learning paradigm~(\cref{sec:lifelong_supervised_learning}) and discuss three prominent scenarios in lifelong supervised learning~(\cref{sec:prominent_scenarios}). For the sake of simplicity, we drop the term~\textit{supervised} when referring to lifelong learning and make it explicit when we are referring to lifelong reinforcement learning.

\section{Background: Supervised Learning}
\label{sec:supervised_learning}
In supervised learning, we want to learn a function $f: \mathcal{X} \rightarrow \mathcal{Y}$ that is able to predict a target vector $y \in \mathcal{Y}$, when given an input sample $x \in \mathcal{X}$ (where $x$ can be in raw form, or in the form of a curated set of features for the raw input). To do so, we have access to some training data $D = \{(x_i, y_i)_{i=1}^n\}$, which consists of $n$ pairs of input samples $x_i \in \mathcal{X}$ and their corresponding target vectors $y_i \in \mathcal{Y}$. We assume data is drawn i.i.d. from a fixed distribution $P(x,y)$.

In order to train this function $f$, we use some loss function $L$ that captures how much the function's prediction $\hat{y} = f(x)$ is different from the ground truth $y$ given a sample $x$. The risk associated with this function becomes:
\begin{equation}
    R(f) = \mathbb{E}_{x,y \sim P} [L (f(x), y)]\,,
\end{equation}
and hence the optimal function $f^*$ is the function that minimizes this risk:
\begin{equation}
    f^* = \argmin_f R(f)\,.
\end{equation}

However, since the distribution $P$ is unknown, the risk $R$ cannot be computed. As an alternative, the Empirical Risk Minimization (ERM) principle \citep{risk_minimization} is usually used, which seeks to obtain the optimal function $\hat{f}$ that minimizes the empirical risk $\hat{R}$
\begin{gather}
    \hat{R}(f) = \frac{1}{n} \sum_{i=1}^n L(f(x_i), y_i) \,,\\
    \hat{f} = \argmin_f \hat{R}(f)\,.
\end{gather}

\section{Lifelong Learning Formulation}
\label{sec:lifelong_supervised_learning}

In the lifelong learning setup, there exists a sequence of tasks, where each task $t$ represents a set of unique classes $\mathcal{C}^{(t)}$, where $\mathcal{C}^{(t)} \subseteq \mathcal{Y}$ (the set of all possible classes). The tasks come in a sequence one by one and each task $t$ comes with its set of data $D^{(t)} = \{(x_i, y_i)_{i=s}^{s + n_t}\}$, where $x_i \in \mathcal{X}$ and $y_i \subseteq \mathcal{C}^{(t)}$. 

The output space $\mathcal{Y}^{(\T)}$ keeps expanding whenever a new task $\T$ is introduced $\mathcal{Y}^{(\T)} = \bigcup_{t=1}^{\T} \mathcal{C}^{(t)}$, the goal is still to learn the function that maps the input to output space across all seen tasks $f_{\T}: \mathcal{X} \rightarrow \mathcal{Y}^{(t)}$. Applying the ERM principle as is would lead us to the following equation:

\begin{gather}
    \hat{R}_{\T}(f) = \frac{1}{\T} \sum_{t=1}^{\T} \frac{1}{|D^{(t)}|} \sum_{(x_i, y_i) \in D^{(t)}} L(f(x_i), y_i)\,, \label{eqn_per_task_risk} \\
    \hat{f}_{\T} = \argmin_f \hat{R}_{\T}(f)\,.
\end{gather}

However, as the data from older tasks $t < \T$ is not available anymore, calculating the risk this way becomes infeasible. On the other hand, minimizing the risk on only the currently available data will lead to good performance on the current task and potential catastrophic forgetting of the previous tasks. We shall explain in the next chapter how the different existing methods try to deal with this issue.

\section{Prominent Scenarios in Lifelong Learning}
\label{sec:prominent_scenarios}
There are three prominent scenarios in lifelong learning: Domain-incremental Learning, Task-incremental Learning, and Class-incremental Learning. These scenarios assume that during training, there are clear and well-defined boundaries between the tasks to be learned \citep{vandeven2019scenarios} (though the learning system may not have access to these task boundaries). These scenarios are distinguished by whether task identity $t$ is provided during evaluation and, if it is not, whether task identity must be inferred.

\begin{figure}[htbp!]
\centering
\begin{subfigure}{.33\textwidth}
  \centering
  \includegraphics[width=1.\linewidth]{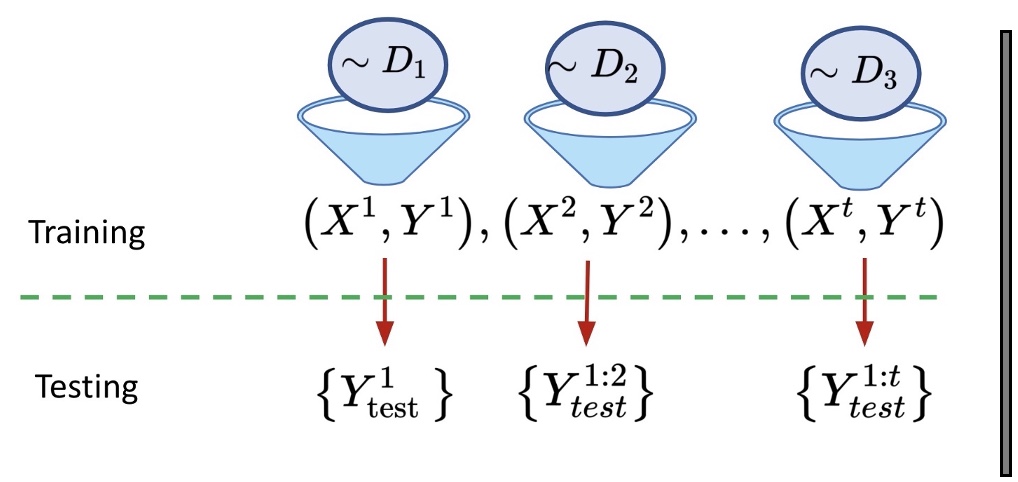}
  \caption{Domain-Incremental}
  \label{fig:sub1}
\end{subfigure}%
\begin{subfigure}{.33\textwidth}
  \centering
  \includegraphics[width=1.\linewidth]{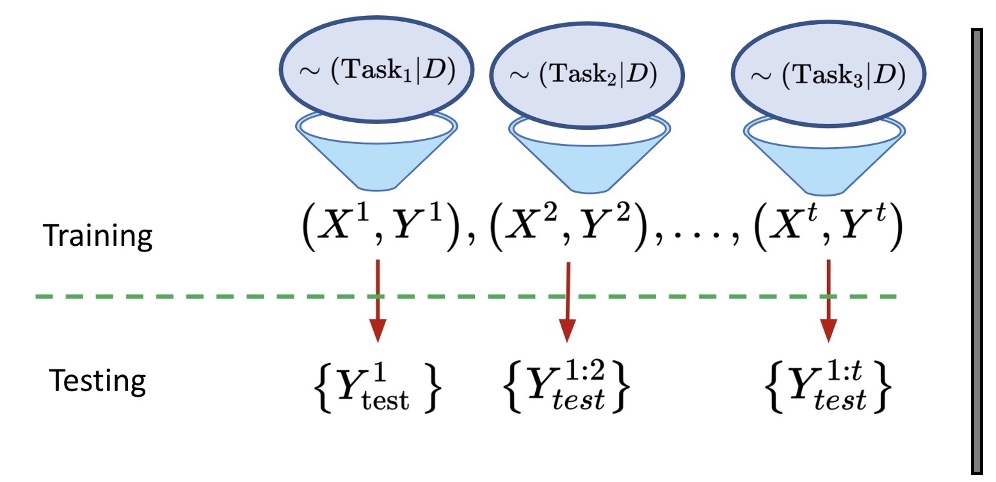}
  \caption{Task-Incremental}
  \label{fig:sub2}
\end{subfigure}%
\begin{subfigure}{.33\textwidth}
  \centering
  \includegraphics[width=1.\linewidth]{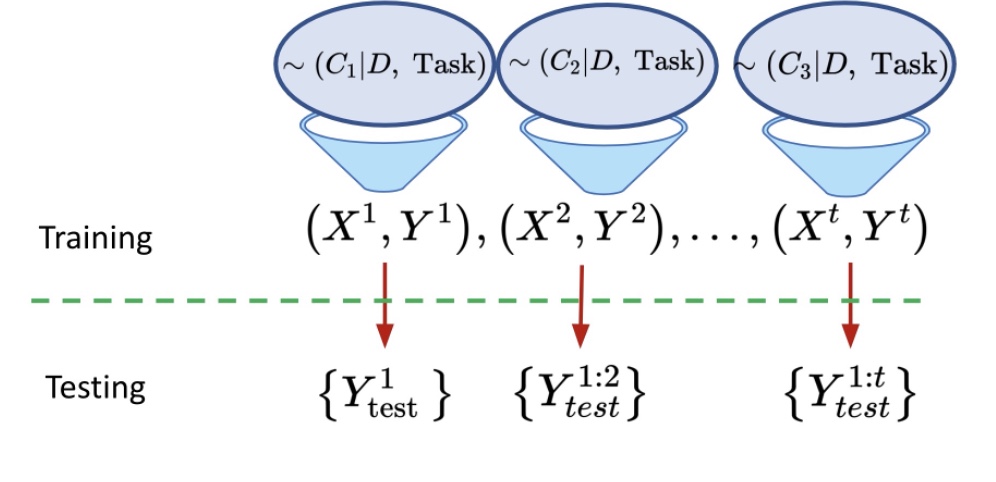}
  \caption{Class-Incremental}
  \label{fig:sub3}
\end{subfigure}
\caption{Overview of the three lifelong learning scenarios}
\label{fig:test}
\end{figure}

\subsection{Domain-incremental Learning}
In the domain-incremental learning scenario~(Figure \ref{fig:sub1}), the system does not need (and does not have) access to the task identity $t$ during evaluation. In this setup, the input distributions are different, while the output distribution is the same, i.e., $P(x^{(a)}) \neq P(x^{(b)})$ and $\mathcal{C}^{(a)} = \mathcal{C}^{(b)}  ~\forall a, b \in W \text{ if } a \neq b$ where $W$ is the set of whole numbers.
In this setup, the models have a single-headed output layer, and each class has the same semantic meaning across all the tasks. Since the system does not have to choose an output head, it does not need to infer the task identity. 

\subsection{Task-incremental Learning}
In the task-incremental learning scenario, the model is trained on a sequence of tasks with known task identities. Since task identity is always provided, it is possible to train models with task-specific components using a \textit{multi-headed} output layer (for deep neural networks) \citep{ewc,chaudhry2019tiny,mirzadeh2020understanding}. 
The output classes are disjoint between tasks, $P(x^{(a)}) \neq P(x^{(b)})$, $P(y^{(a)}) \neq P(y^{(b)}),~\mathcal{C}^{(a)} \cap \mathcal{C}^{(b)} = \Phi \quad \forall a, b \in W \text{ if } a \neq b$ in the task-incremental scenario and models are evaluated by their average final performance across all tasks after being trained on all tasks sequentially (see Figure \ref{fig:sub2}). Here, when evaluating on a given task, the model's predictions for only the classes corresponding to the given task are considered~\citep{vandeven2019scenarios}. 

\subsection{Class-incremental Learning}
In the class-incremental learning scenario, the model must infer the task identity and solve the tasks seen so far. It is, by far, the most challenging setting in lifelong learning, and many existing methods fail in this setting \citep{Rebuffi_2017,aljundi2019online}. This scenario employs a single-head architecture where the output space is the same for all distributions, and the model needs to classify all labels without a task-ID (Figure \ref{fig:sub3}). Here, $P(x^{(a)}) \neq P(x^{(b)})$ and $P(y^{(a)}) \neq P(y^{(b)}) ~\forall a, b \in W \text{ if } a \neq b$.  For instance, considering a classification task using deep neural networks, the units of all the classes seen so far are active in this scenario. 

\section{An overview of Lifelong Learning strategies}
\label{sec:lifeong_learning_strategies}
A wide range of methods have been proposed in the past years to tackle the challenges in lifelong learning. However, each method makes assumptions that are not consistent due to the presence of different settings defined above. In particular, a few methods require fewer supervisory signals during both training and inference times and hence generalize better to different lifelong learning settings. Such signals can be a natural number for task identity, a natural language descriptor, a vector representation of data describing a task, etc. However, there is a clear trend in recent works to simultaneously apply multiple techniques to tackle this problem. 

Methods proposed in lifelong learning are broadly categorized into the following three categories: Regularization-based, Memory-based, and Architecture-based methods \citep{de2019continual,masana2020class}.

\begin{figure}[htbp!]
    \centering
    \includegraphics[width=.7\linewidth]{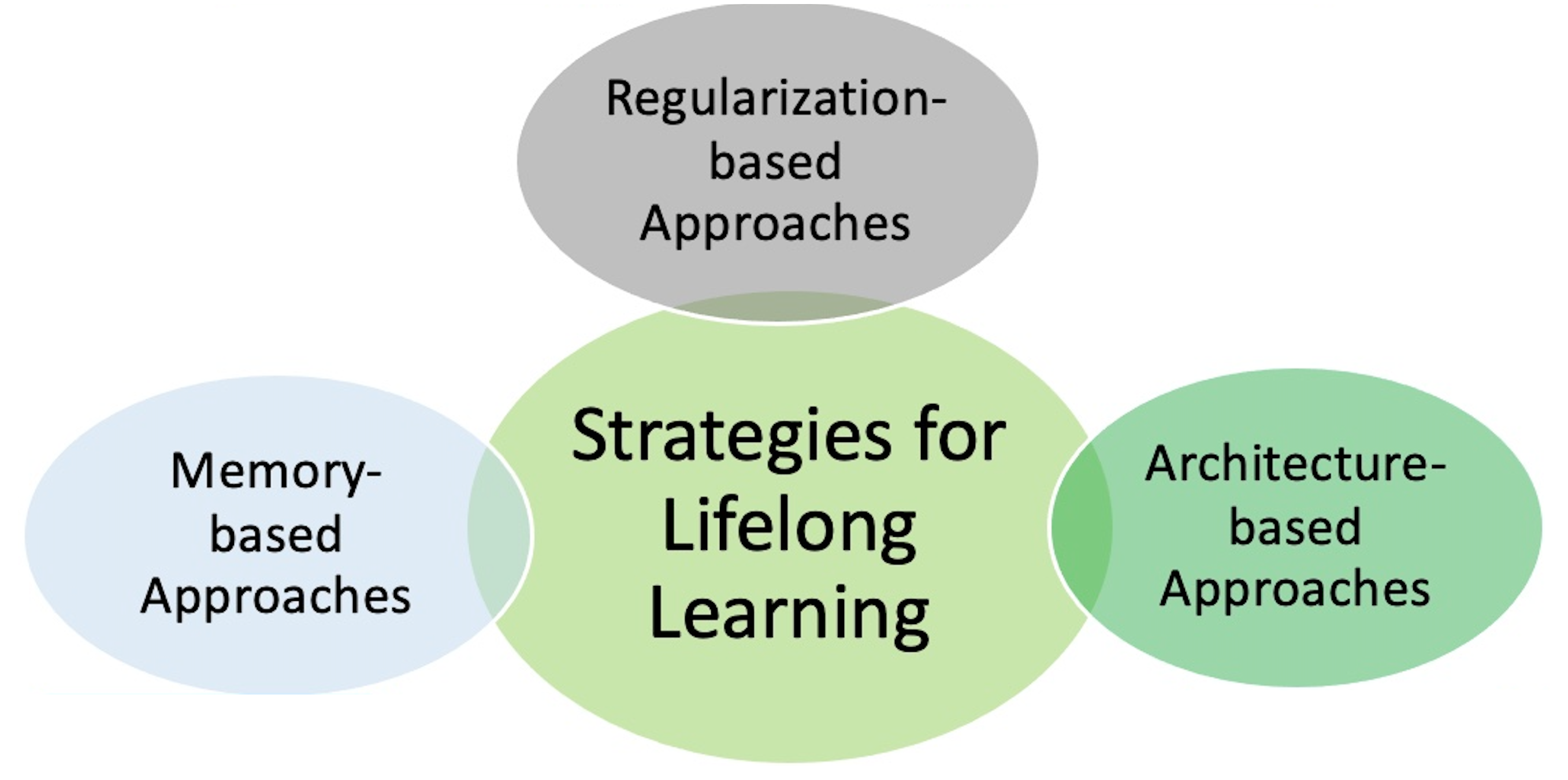}
    \caption{Common strategies in Lifelong Learning}
    \label{approaches}
\end{figure}

\subsection{Regularization-Based Methods}
Regularization-based methods prevent a drastic change in the network parameters as the new task arrives to mitigate forgetting. These methods are further classified as importance-based, Bayesian-based, distillation-based, and optimization trajectory-based. Here, the importance-based methods regularize the loss function to minimize changes in the parameters important for previous tasks. Distillation-based methods transfers knowledge from the model trained on the previous task to the model being trained on the new data. On the other hand, optimization trajectory-based methods exploit the geometric nature of the local minima to prevent catastrophic forgetting. These methods are shown to be vulnerable to domain shift between tasks \citep{2017_expert_gate_lifelong_learning_with_a_network_of_experts}. We discuss regularization-based methods in Chapter \ref{sec:regularization}.

\subsection{Memory-Based Methods}
Memory-based methods maintain an `episodic memory', containing a few examples from past tasks that are revisited while learning a new task. These methods apply gradient-based updates that facilitate a high-level transfer across different tasks through the examples from the past tasks that are simultaneously available while training on the current new task. For instance, Averaged Gradient Episodic Memory (\textit{A-GEM}) \citep{chaudhry2019efficient} uses the episodic memory to project the gradients based on hard constraints defined using the episodic memory and the current mini-batch. Experience Replay (\textit{ER}) \citep{chaudhry2019tiny} uses both replay memory and input mini-batches in the optimization step by averaging their gradients to mitigate forgetting. However, selecting which examples to store is also a significant challenge that has been the focus of various research works. Instead of storing raw samples, Generative Replay trains a deep generative model such as GAN \citep{goodfellow2020generative} to generate data that mimic past data for replay. However, it takes a long time to train such generative models and hence is not a viable option for complex datasets in terms of computational cost. We discuss memory-based methods in Chapter \ref{sec:memory}.

\subsection{Architecture-Based Methods}

Architecture-based methods either freeze or add a set of parameters with the idea that different tasks should have their own set of isolated parameters. These methods alleviate catastrophic forgetting in general, but they rely on a strong base network and work on a small number of tasks. For example,  \citet{2017_expert_gate_lifelong_learning_with_a_network_of_experts} assigns a model copy to every new task that arrives. Similarly, there are expansion-based methods that handle the lifelong learning problem by expanding the model capacity in order to adapt to new tasks \citep{2018_on_training_recurrent_neural_networks_for_lifelong_learning,rao2019continual}. We discuss architecture-based methods in Chapter \ref{sec:architecture}.

\section{Desiderata of Lifelong Learning Systems}
\label{subsec:desidrata}
Several works have outlined useful properties and open challenges for lifelong learning systems, both in the context of supervised learning~\citep{sodhani2021multi, veniat2021efficient, cl_dnn_hadsell} and reinforcement learning~\citep{schaul2018barbados}. We compile these properties into a list of desired properties of a model suitable for lifelong learning settings:

\begin{enumerate}
    \item~\textbf{Knowledge Retention} - As the model trains over the new tasks, it should not forget the knowledge from the previous tasks. Learning new tasks should not happen at the expense of the knowledge from the previous tasks. Much of the existing literature focuses on this problem (under the name of Catastrophic Forgetting~\citep{ mccloskey1989catastrophic, french1999catastrophic}). Due to this problem, conventional deep learning tends to focus on offline training, with i.i.d. sampling of mini-batches with multiple epochs over the training data. Therefore, the model requires a significant amount of previous tasks data to learn and accumulate explicit knowledge. Some works refer to this knowledge retention property as~\textit{plasticity} or~\textit{stability}.

    \item~\textbf{Knowledge Transfer} - The model should be able to reuse the knowledge across tasks. This includes both~\textit{forward} transfer of knowledge where the knowledge acquired during previous tasks is used to solve the subsequent tasks, and~\textit{backward} transfer where the knowledge acquired in the current/future tasks is used to improve performance on the previous tasks. The underlying premise is, if the tasks are related, this knowledge transfer could lead to faster learning and better generalization. Most current approaches for knowledge transfer focus on the forward transfer of knowledge.
    \item~\textbf{Model Expansion} - As the model trains over a sequence of tasks, the model should be able to~\textit{expand} itself or increase its learning~\textit{capacity}. This could mean that the model can introduce new trainable parameters in practice. Also, training a separate model entirely for each task discounts the possibility of transferring knowledge forward and backward directions when the tasks are related. This further discounts better generalization or faster learning. In~\cref{sec:expansion}, we discuss expanding networks that aim to tackle these problems by increasing the model capacity and reusing learned representations.
    \item~\textbf{Parameter Efficiency} - While increasing the model's capacity, we would also want the computational and memory costs of the model to increase only sub-linearly (or to be bounded) as the model trains on new tasks to avoid computational performance degradation. The model expansion property comes with additional constraints: In the true lifelong learning setting, the model would experience a continual stream of training data that can not be stored. Hence the model would, at best, have access to only a small sample of the historical data. We can not rely on past examples to train the expanded model from scratch in such a setting, and a zero-shot knowledge transfer is desired. 

\end{enumerate}

\section{Relation to Other Areas}
\label{sec:relation_to_other_areas}

Lifelong learning is referred by different names in the literature: incremental learning \citep{solomonoff1989system}, continual learning \citep{de2019continual}, explanation-based learning \citep{thrun1996explanation,thrun2012explanation}, never-ending learning \citep{carlson2010toward}, etc. The underlying idea in all these works is that lifelong learning systems would be more effective at learning and retaining knowledge across different tasks. In principle, the ability to generalize is one of the most important characteristics of a machine learning model. If tasks are related, then knowledge transfer between tasks should lead to a better generalization, and faster learning \citep{Biesialska_2020}. 

Lifelong learning also bears some resemblance to other dominant research areas. It is closely related to areas like Multitask Learning~\citep{caruana1997multitask_learning}, Meta Learning~\citep{schmidhuber1987evolutionary, Thrun1998}, Transfer Learning~\citep{pan2009survey}, Online Learning~\citep{Shalev-shwartz07onlinelearning:,MAL-018}, and Curriculum Learning~\citep{bengio2009curriculum}.

\subsection{Multitask Learning}  

The paradigm of multitask learning focuses on improving the performance of a single model on multiple tasks by sharing knowledge across tasks~\citep{caruana1997multitask_learning, zhang2014facial_landmark_detection_by_deep_multitask_learning, ruder2017overview, radford2019language_models_are_unsupervised_multitask_learners, sodhani2021multi}. This goal is quite similar to the goal of lifelong learning systems, with one major difference - multitask learning approaches generally assume that information about all the tasks is known when the training starts. In practice, this means that the learning system has access to all the tasks, and in some cases, the system can even choose the ordering of the tasks~\citep{bengio2009curriculum, pentina2015curriculum} as done in curriculum learning. This assumption is generally not valid for lifelong learning setups where neither the number of tasks nor the nature of tasks is assumed to be known when starting the training.

Since the learning system has upfront knowledge about all the tasks, catastrophic forgetting is not usually studied in multitask learning. However, a closely related challenge that is well-studied in the context of multitask learning is the problem of negative interference~\citep{adapting_auxiliary_losses_using_gradient_similarity, regularizing_deep_multi_task_networks_using_orthogonal_gradients, yu2020gradient} where the gradients corresponding to the different tasks interfere negatively with each other, thus slowing down (or completely inhibiting) training on multiple tasks. Negative interference is related to catastrophic forgetting as it can cause the learning system to forget (or~\textit{unlearn}) knowledge from one or more tasks. 

In some cases of multitask learning, referred to as sequential multitask learning~\citep{zhang2017survey, xiong2018guided}, the different tasks may be introduced sequentially, over a period of time. This setup is closer to the lifelong learning setup (as compared to the general multitask learning setup) but even in the case of sequential multitask learning, the information about all the tasks is generally assumed to be known upfront.

Multitask Learning also shares several similarities with lifelong learning in terms of inductive biases and architecture choices. For example, modular networks are a common design choice for both multitask learning~\citep{end_to_end_multi_task_learning_with_attention,learning_modular_neural_network_policies_for_multi_task_and_multi_robot_transfer, chang2018automatically,sodhani2021multi} and lifelong learning (\cref{subsec:modular_networks}). In both the cases, the inductive bias of compositionality and learning expert~\textit{knowledge} (or~\textit{skills}) is seen as a useful property for the learning model.

\subsection{Meta Learning}

Meta Learning, also known as~\textit{learning to learn}~\citep{Thrun1998, bengio2013optimization}, is the machine learning paradigm that focuses on enabling the training system to learn aspects of the learning process itself. This paradigm can be seen as a natural extension from learning features and models to learning~\textit{algorithms}. Meta Learning comprises of three broad family of approaches: (i) Metric-based~\citep{Koch2015SiameseNN, Vinyals2016MatchingNF, Sung2018LearningTC, Snell2017PrototypicalNF}, (ii) Model-based~\citep{Santoro2016MetaLearningWM, Munkhdalai2017MetaN}, and (iii) Gradient-based~\citep{mishra2018a, Ravi2017OptimizationAA, Finn2017ModelAgnosticMF}.

Meta-learning and lifelong learning approaches have similar motivation - train on the distribution of tasks to improve performance on new, potentially unseen tasks. Similar to lifelong learning, several meta-learning approaches generally do not assume access to all the training tasks at the start of the training. Several works have started focusing on the intersection of lifelong learning and meta-learning~\citep{AlShedivat2018ContinuousAV, Ritter2018BeenTD, Nagabandi2019LearningTA, Javed2019MetaLearningRF, wang2020efficientML}. However, the two paradigms also have some differences. Unlike lifelong learning methods, Meta-learning approaches generally do not focus on challenges like catastrophic forgetting or capacity saturation. On the other hand, meta-learning approaches use an explicit objective function that incentives faster training on the new tasks while lifelong learning implicitly optimizes for accelerating training.

\subsection{Transfer Learning}

The paradigm of transfer learning~\citep{dai2009eigentransfer, pan2009survey, torrey2010transfer, bengio2012deep, weiss2016survey, ying2018transfer, tan2018survey, zamir2018taskonomy, zhuang2020comprehensive} focuses on transferring knowledge from one or more~\textit{source} tasks to one or more~\textit{target} tasks. It is related to the lifelong learning paradigm as the learning system is trained over multiple tasks with the hope of doing a forward knowledge transfer to the subsequent tasks. 

Transfer learning faces several challenges similar to lifelong learning when considering to transfer a trained model to a new task:~(i) Should the model's architecture be changed (for example by adding more parameters \citep{2016_progressive_neural_networks} or modules \citep{1999_modular_neural_networks_a_survey})?~(ii) Should some parts of the model be frozen (if yes, which parts?) or should the entire network be finetuned?~(iii) How should we set the learning rate on the new tasks?~(iv) How to~\textbf{infer} the~\textit{relatedness} between the tasks. Interestingly, some of the architecture choices and inductive biases (like modular architectures) that are useful for lifelong learning \citep{veniat2021efficient} is useful for transfer learning as well~\citep{houlsby2019parameter, stickland2019bert}.

However, transfer learning is also different from lifelong learning in several ways:~(i) Transfer learning generally focuses on~\textit{one-way} transfer of knowledge, from the older to the newer task. In contrast, lifelong learning focuses on~\textit{two-way} transfer of knowledge, from both old tasks to new tasks and vice-versa.~(ii) Transfer learning focuses primarily on the performance of the current task. Catastrophic forgetting is not seen as a problem and is often not even measured. On the other hand, lifelong learning aims to improve performance over all the tasks.~(iii) Transfer learning is often used to~\textit{initialize} a model such that it can perform well on the target task. Hence, many approaches for transfer learning involve pre-training on a large corpus. This is not the case with Lifelong learning.

\subsection{Online Learning}

Standard machine learning paradigms (especially in the context of supervised learning and unsupervised learning) use the~\textit{batch} (or~\textit{offline}) learning approach where any given data point can be used for training any number of times. While this approach often works well in practice, it may be infeasible in certain setups. For example, there may be privacy-related restrictions for storing the data, making it infeasible to use it for offline training. In other cases, the size of the training data could be unbounded (as in the case of click-stream data). In such a case, storing (and training over) all the historical data is not practical.~\textit{Online learning}~\citep{Shalev-shwartz07onlinelearning:, MAL-018,10.5555/3041838.3041955} is a paradigm in machine learning that aims to address some of these limitations. 

Online learning techniques have been used in conjunction with other machine learning paradigms like multi-task learning~\citep{dekel2006online, agarwal2008matrix, li2013collaborative, wang2016large}, metric learning~\citep{shalev2004online, jain2008online, 7244184}, transfer learning~\citep{ZHAO201476, 10.5555/1953048.2021051,10.1145/2505515.2505603,6467144} etc.
The connection between lifelong learning and online learning is less obvious as the majority of works in lifelong learning focus on the~\textit{batch} (samples within a task) setup. However, several recent lifelong learning works are starting to focus on the online setup and have argued in favor of online lifelong learning setup to be closer to real-life learning as compared to the offline counterpart~\citep{pmlr-v119-chrysakis20a, sodhani2020toward, 2020arXiv200309114P, 2021arXiv210110423M, NEURIPS2019_15825aee, Aljundi_2019_CVPR, pham2021contextual, Liu_2020, kruszewski2021evaluating, malviya2021tag}.

\subsection{Curriculum Learning}

Humans and animals learn more efficiently when starting with simpler~\textit{concepts} (or tasks) and progressively learning more complex concepts (or tasks)~\citep{skinner1958reinforcement, peterson2004day}. For example, the human education system is designed as a curriculum where new concepts build on (and leverage) previous concepts. Curriculum learning~\citep{elman1993learning, bengio2009curriculum, pmlr-v97-hacohen19a,wang2020survey} is the machine learning paradigm that aims to leverage insights about the importance of curriculum and use these insights to improve the training of machine learning models. Given the generic nature of curriculum learning, it has been used in conjunction with other machine learning paradigms like multi-task learning~\citep{pentina2015curriculum, sarafianos2017curriculum, murugesan2017self}, reinforcement learning~\citep{narvekar2017curriculum,narvekar2018learning,narvekar2020curriculum, portelas2020automatic}, transfer learning~\citep{dong2017multi, weinshall2018curriculum}, etc.

Curriculum learning and lifelong learning have several commonalities. Both the paradigms can be motivated from the perspective of human cognition and involve a notion of~\textit{continuous} learning - over a sequence of datasets in lifelong learning and over a sequence of splits of one (or more) datasets in curriculum learning. However, there are notable differences as well. In the general lifelong learning setup, the learning system has no control over the sequence of data points. In contrast, curriculum learning focuses on the most optimal sequence of data points (optimal to learning). In curriculum learning, information about the different datasets/data points is available beforehand, while in the lifelong learning setup, this information becomes available during training. Despite these differences, curriculum learning can be a helpful technique in the context of lifelong learning, and the general idea of selecting data points, based on their estimated~\textit{hardness}, has been used in several approaches for experience replay \citep{andrychowicz2017hindsight,li2021parallel}.

\section{Common Metrics in Lifelong Learning}
\label{sec:metrics}
Lifelong learning differs from supervised learning in terms of how the systems are trained and evaluated. These differences imply that the use of the traditional, ~\textit{single-task} performance metrics like top-$1$ or top-$5$ error rates is not suitable for lifelong learning systems. As discussed in the previous sections, alleviating catastrophic forgetting and knowledge transfer are the crucial challenges that the methods should focus on in lifelong learning. Therefore, we need metrics to measure the models' performance appropriately in a lifelong learning setup. The metrics should evaluate lifelong learning methods to assess their performance through time, including how much the model forgets or gains on the previously learned knowledge. In this section, we explain some of the most popular metrics in lifelong learning, including the average accuracy for overall performance~\citep{chaudhry2019efficient}, the average forgetting~\citep{chaudhry2018riemannian}, the forward and the backward knowledge transfer that assesses the ability of the models to transfer knowledge~\citep{lopezpaz2017gradient,lesort2019continual}.

\subsection{Performance Metrics}
\label{sec:LL_performance_metrics}
In lifelong learning settings, a system learns from the dataset \((x_{1},\,y_{1}), \ldots, (x_{\T},\,y_{\T}),\) at each episode where $x_{i}$ denotes input variable and $y_{i}$ denotes target/output variable belonging to training set of a task $i$. However, the system's performance is reported based on \(\{x^{test}_{1}, y^{test}_{1}\},\ldots, \{x^{test}_{1:\T}, y^{test}_{1:\T}\}\).

In the class incremental learning setting, the system incrementally learns a set of classes. The model also incrementally learns a new task at each time in task incremental learning. The model aims to have less forgetting through time and better performance. The \textbf{average accuracy} ~\citep{chaudhry2018riemannian} is computed as follows:
\begin{align}
    A_{\mathcal{T}}=\frac{1}{\mathcal{T}} \sum_{i=1}^{\mathcal{T}} a_{\mathcal{T}, i},
\end{align}
where $A \in[0,1]$, $\mathcal{T}$ is the total number of tasks seen so far,  and $a_{n, i}$ is the test classification accuracy on task $i$ after sequentially learning the $n^{\text {th }}$ task. Forgetting Measure is the another metric that is very crucial in the lifelong learning model performance report. \citet{chaudhry2018riemannian} introduce the \textbf{forgetting} measure, formally defined as follows:

\begin{align}
    F_{\mathcal{T}}=\frac{1}{\mathcal{T}-1} \sum_{i=1}^{\mathcal{T}-1} f_{{\mathcal{T},i}},
\end{align}
where $F \in[-1,1]$, $f_{j, i}$ is a measure of forgetting on task $i$ after training up to task $j$. $f_{j, i}$ is defined as the difference between best accuracy achieved on task $i$ in the past and the final accuracy of task $i$ after training on task $j$:
\begin{align}
    f_{j, i}=\max _{k \in\{1, \cdots, j-1\}} a_{k, i}-a_{j, i}.
\end{align}

The \textbf{average forgetting ratio} is another metric introduced by~\citet{2018_overcoming_catastrophic_forgetting_with_hard_attention_to_the_task}. It measures the amount of forgetting over time and studies the effectiveness of the lifelong learning method in multiple datasets relatively. After training on task $t$, it computes the accuracy on all testing sets of tasks $\tau \leq t$. This process is repeated multiple times using different seeds for uniformly randomized task-order. Then, the forgetting ratio is defined as follows:
\begin{align}
    \rho^{\tau \leq t}=\frac{A^{\tau \leq t}-A_{\mathrm{R}}^{\tau}}{A_{\mathrm{J}}^{\tau \leq t}-A_{\mathrm{R}}^{\tau}}-1,
\end{align}
where $A^{\tau \leq t}$ is the accuracy measured on task $\tau$ after sequentially learning task $t, A_{\mathrm{R}}^{\tau}$ is the accuracy of a random multi-layer linear classifier using the class information of task $\tau$ and $A_{\mathrm{J}}^{\tau \leq t}$ is the accuracy measured on task $\tau$ after jointly learning $t$ tasks in a multitask learning manner ~\citep{2018_overcoming_catastrophic_forgetting_with_hard_attention_to_the_task}. 
To compute the average ratio, we can simply compute the average as follows:
\begin{align}
    \rho^{\leq t}=\frac{1}{t} \sum_{\tau=1}^{t} \rho^{\tau \leq t}.
\end{align}

\begin{figure}[htbp!]
    \centering
    \includegraphics[width=.6\textwidth]{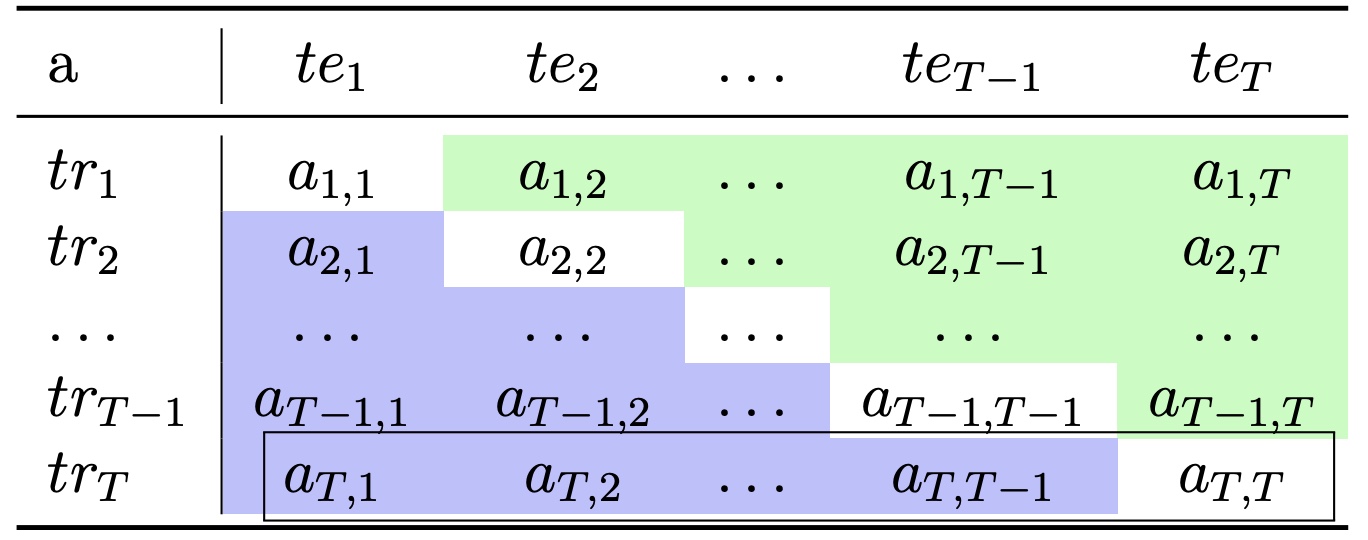}
    \caption{$t r_{i}$ and $t e_{i}$ denote training and test set of task i. $B W T^{+}$ is the average of the difference between accuracies in purple and accuracies in the diagonal. $F W T$ is the average of accuracies in green. $A_{T}$ is the average of accuracies in the box in the last row~\citep{2021arXiv210110423M}.}
    \label{fig:benchmarks_ac_bwt_fwt}
\end{figure}

The Positive Backward Transfer and Forward Transfer metrics are two more important metrics in lifelong learning. \citet{2021arXiv210110423M} visually shows how we can compute these metrics and what they measure. Figure~\ref{fig:benchmarks_ac_bwt_fwt} illustrates their explanation. 

The Positive Backward Transfer metric measures the positive influence of learning a new task on preceding tasks' performance. Positive Backward Transfer metric is denoted as $B W T^{+}$ and computed as follows~\citep{2021arXiv210110423M}:
\begin{equation}
    \begin{gathered}
    B W T=\frac{\sum_{i=2}^{T} \sum_{j=1}^{i-1}\left(a_{i, j}-a_{j, j}\right)}{\frac{T(T-1)}{2}}
    \\
    B W T^{+}=\max \left(\frac{\sum_{i=2}^{T} \sum_{j=1}^{i-1}\left(a_{i, j}-a_{j, j}\right)}{\frac{T(T-1)}{2}}, 0\right)
    \end{gathered}
\end{equation}
where Backward Transfer and Positive Backward Transfer are denoted as $B W T$ and $B W T^{+}$ respectively. 
As Figure~\ref{fig:benchmarks_ac_bwt_fwt} shows, the purple area corresponds to the area used to compute Positive Backward Transfer. $BWT < 0$ indicates catastrophic forgetting, and $BWT > 0$ indicates that learning new tasks has helped with the preceding tasks~\citep{ebrahimi2020adversarial}. 
The Forward Transfer metric denoted as $F W T$ measures the positive influence of learning a task on future tasks' performance. We can compute $F W T$ as follows: 
\begin{align}
    F W T=\frac{\sum_{i=1}^{j-1} \sum_{j=1}^{T} a_{i, j}}{\frac{T(T-1)}{2}}
\end{align}

In Figure~\ref{fig:benchmarks_ac_bwt_fwt}, $F W T$ is the average of accuracies in green.

Learning Curve Area (LCA $\in[0,1])$ is another performance metric proposed by~\citet{chaudhry2019efficient}. To explain the LCA metric, we need to define an average $b$-shot performance after the model has been trained for all the $T$ tasks as:
\begin{align}
    Z_{b}=\frac{1}{T} \sum_{k=1}^{T} a_{k, b}\,,
\end{align}
where $b$ is the number of  mini-batches. $\mathrm{LCA}$ at $\beta$ is defined as the area of the convergence curve $Z_{b}$ as a function of $b\in[0, \beta]$:
\begin{align}
    \mathrm{LCA}_{\beta}=\frac{1}{\beta+1} \int_{0}^{\beta} Z_{b} d b=\frac{1}{\beta+1} \sum_{b=0}^{\beta} Z_{b}
\end{align}
It is worth mentioning that $\mathrm{LCA}_{0}$ is the average zero-shot performance and is considered the same as the forward transfer performance. $\mathrm{LCA}_{\beta}$ and the area under the $Z_{b}$ curve will be high when the zero-shot performance is good, and it shows how quickly the model learns new tasks. This metric is valuable when two models have the same $Z_{\beta}$ or $A_{T}$, but very different $\mathrm{LCA}_{\beta}$ where one learns much faster than the other with same final accuracy~\citep{chaudhry2019efficient}.

Aside from the metrics discussed here, there are other useful metrics that can reveal the potential weakness or strength of the methods. Following is a list of some of the traditional performance metrics that can be used in the lifelong learning domain.
\begin{itemize}
    \item  Throughput (images/sec) at train and test time.
    \item  Mean and standard deviation of top-1 and top-5 error rates for each individual task in task incremental learning.
    \item  Comparison of the method's performance considering the replay buffer size or the memory overhead.
    
    \item Confusion matrix comparison. Since it is tough to observe the differences between
    the reported confusion matrix from different approaches, we have to find a nice and
    cheap way to compare two very similar confusion matrix~\citep{wu2019large, Hou_2019_CVPR, Abdelsalam_2021_CVPR}. 
    \item Similarity Measurement of the classes that the model should learn continually. In this case, the order of the tasks will be more meaningful in the experiment result. Since the forgetting of the model correlates with the order of the task, considering the tasks sequence and measuring their similarity will be essential to have a fair comparison with other methods.

\end{itemize}

\subsection{Time}

The time-related metrics that can be reported either task by task or as a historical average are task's training time comparison, testing time comparison, and validation time comparison.

\subsection{Memory}

Replay-based methods are the most popular methods in lifelong learning. These methods do not assume any limitations to access to data from the previous tasks. In practice, samples of data, from the previous tasks, are retained into a memory bank called the~\textit{replay buffer}. The performance of these methods depends on the size of the replay buffer and the strategy of selecting, editing, and removing samples from the replay buffer. Indeed, the size of the memory or replay buffer is the most critical parameter that should be reported and included in the model performance evaluation process. Following is a list of simple metrics and parameters that we should consider when comparing methods. 
\begin{itemize}
    \item Replay buffer size and the strategy for constructing and updating the replay buffer. 
    \begin{itemize}
        \item Fixed memory size should be reported if a fixed window is used to keep samples in the replay. 
        \item Some methods use a replay such that a fixed window per class is reserved to keep track of replayed samples. In this case, there is no fixed memory size for the lifelong learning method, but instead, a fixed window is reserved for each class. Therefore as the model learns new classes or tasks, we expand the memory size for new tasks or set of classes. Reporting the strategy that is used to construct the memory and the size of the memory is important.   
    \end{itemize}
    \item A metric to show how much the memory consumption grows as the model learns a new concept. This is different from the size of the replay that is explained above. Some approaches need more memory to alleviate the catastrophic forgetting and use the parameter isolation method (discussed in \cref{sec:isolation}). In such cases one needs to keep track of the memory consumption overhead as models learn new concepts through time. 
    
    \item A metric to evaluate the memory update rates when we train a lifelong learning model in a large-scale memory-based distributed computing cluster. The memory update rates are important in such cases because of the network communication overhead that might decrease the computation speed up either at training or testing time.
    
    \item The number of memory components also should be reported if the proposed method uses a different type of extra memory component for different purposes such as keeping previous task models or hidden representation or extracted features of some samples from the previous task.
\end{itemize}

\chapter{Regularization-based Approaches}
\label{sec:regularization}
In this chapter, we shall explore how the different methods try to overcome the challenges of lifelong learning without having access to any data from previous tasks nor having the ability to expand the network to learn new tasks. It should be noted that this chapter and the following two chapters are, for the most part, orthogonal to each other in their approaches, which means that the different approaches can, in practice, be combined as done in previous works like~\citet{sodhani2020toward}.

\begin{figure}[!b]
    \centering
    \includegraphics[width=0.8\textwidth]{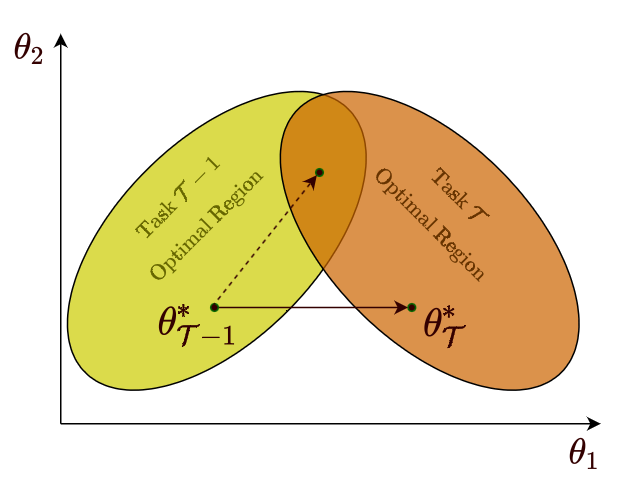}
    \caption{After learning the task $\T-1$, the parameters are at $\theta^*_{\T-1}$. While learning the task $\T$, if the model follows the solid line to reach $\theta^*_{\T}$, it may incur a significant loss on task $\T-1$, i.e., it suffers from catastrophic forgetting. On the other hand, the goal is to follow the dashed line to reach an optimal parameter setting to incur a minimal loss on both tasks.}
    \label{fig:catastrophic_forgetting}
\end{figure}

When a neural network is trained sequentially on a series of tasks, the network parameters try to minimize the objective on the current task irrespective of what happens to the performance of the previous tasks. In other words, the network does not see except what we allow it to see, and it does not solve except what we ask it to solve. When the network is trained on task 1, we ask it to minimize the objective on task 1. However, when it is trained on task 2, if we only ask it to minimize the objective on task 2 (as we do not have access to the data of task 1 anymore), this can lead to the deterioration of the network performance on task 1. The reason behind this is that task 1 is not incorporated in the objective function when training on task 2. This kind of behavior is known as catastrophic forgetting (see Figure~\ref{fig:catastrophic_forgetting}), which takes place when the parameters of the network change sufficiently across the tasks, such that their performance on previous tasks degrades significantly.

\section{Definition}
Let us consider a setup where we receive a continuum of data from different tasks in a sequential manner: $(x_1, d_1, y_1), \cdots, (x_t, d_t, y_t), \cdots, (x_{\T}, d_{\T}, y_{\T})$ where $x_{t}$ is the input data, $d_t$ is the task descriptor and $y_t$ is the target variable of task $t$. Let the current task be $\T$, and the set of weights after training on task $\T$ be $\theta_{\T}$. $\theta$ consists of $N$ parameters $\theta \in R^N$, 
with $\theta_{t}$ denoting the parameters after training on task $t$. The training objective $\tilde{L}$ would normally be our original objective $L_{\T}$ (for example, cross entropy, in the case of classification, on the data that belongs to task $\T$).
\begin{equation}
    \tilde{L}(\theta) = L_{\T}(\theta) := \frac{1}{|D_{\T}|} \sum_{(x,y) \sim D_{\T}} l(\theta; x,y) \,,
\label{objective_fun}\end{equation}
where $l$ is the per sample loss, and $D_{\T}$ is the set of samples that belong to task $\T$.

Regularization-based approaches constrain the update of neural networks to prevent catastrophic forgetting by adding a penalty term ($R_{\T}$) such that the new objective function looks like:
\begin{equation}
    \tilde{L}(\theta) = L_{\T}(\theta) + R_{\T} \,.
\label{regularize}\end{equation}

Regularization-based approaches can be roughly categorized into four main types: (i) importance-based regularization, (ii) Bayesian-based regularization, (iii) distillation-based regularization, and (iv) optimization trajectory-based regularization. Out of these four categories, the first three categories explicitly define $R_{\T}$ using old parameters, training examples/outputs or parameter distribution, etc. On the other hand, optimization trajectory-based regularization exploits the geometric nature of the local minima, and the corresponding trajectories followed to prevent catastrophic forgetting. We discuss each of these categories in detail in this chapter.

\section{Importance-Based Regularization}

Importance-based Regularization tries to introduce solutions that do not require access to samples from previous tasks to alleviate catastrophic forgetting. The first such solution is to try to make the network parameters after training on task 2 (let us call them $\theta_2$) as close as possible to the parameters of the network trained on task 1 ($\theta_1$). One way to do so would be by applying a quadratic constraint between each pair of parameters in $\theta_1$ and $\theta_2$. Although this solution might seem to significantly limit the capacity of the model to solve the different tasks, it can be supported by the over-parametrization of neural networks, where many configurations of $\theta$ can still lead to the same performance \citep{hecht1992theory,sussmann1992uniqueness}. However, doing the regularization this way can still significantly reduce the network's capacity, especially when training takes place sequentially. It is not guaranteed that there exists another solution in the vicinity of $\theta_1$ that can still perform well on both tasks.

More sophisticated solutions under the importance-based regularization umbrella try to solve this problem by making the regularization more selective, giving the network some freedom to change some parameters while limiting this freedom for other parameters, based on the importance of each parameter with respect to the previous tasks. Here the central question becomes how to measure the importance of each parameter in an efficient and tractable way.

Let us make the above ideas more concrete. The objective function defined in Eq. \ref{objective_fun} has no guarantees on the performance over the data of the previous tasks $1:\T-1$. Hence, a naive solution would be to add a quadratic constraint so that the parameters when training on task $\T$ do not deviate from the parameters of task $\T-1$, i.e.,
\begin{equation}
    R_{\T} = \alpha \sum_{i=1}^N (\theta_{i} - \theta_{\T-1,i})^2 \,,
\end{equation}
where $\alpha$ is a regularization weight. The idea here is to find a new set of weights $\theta^*_{\T}$ that can perform well on task $\T$, while at the same time is close enough to $\theta^*_{\T-1}$ so that the performance on task $\T-1$ is preserved. However, adding this constraint this way might be too limiting to the network's capacity. Given that not all the parameters $\theta_i$ contribute equally to the performance, a more selective approach can be used so that the parameters that are more important for previous tasks are regularized more than the other parameters:
\begin{equation}
    R_{\T} = \alpha \sum_{i=1}^N \Omega^{\T-1}_{i} (\theta_i - \theta_{\T-1,i})^2
    \label{eqn:regularization_function} \,,
\end{equation}
where $\Omega_{i}^{\T-1}$ represents the importance of each parameter $\theta_i$ after training on task $\T-1$. Given that $\Omega_{i}^{\T-1}$ is the only measure of importance used, it should also be a function of $\Omega_{i}^{t}\ \forall\ t<\T-1$, so that the performance is preserved on all the previous tasks $t < \T$. In the next section, we shall explore how Elastic Weight Consolidation (\textit{EWC}) \citep{ewc} tries to estimate these importance parameters $\Omega_i$.

\subsection{Elastic Weight Consolidation (EWC)}
\label{sec:regularizzation_ewc}
\label{sec:ewc}

\textit{EWC} \citep{ewc} uses the diagonal terms of the Fisher information matrix as a proxy for the importance of the parameters ($\Omega_i$). However, it motivates it from a probabilistic perspective. Given a set of labelled data $\mathcal{D}$, the optimum set of weights $\theta$ would be the weights that maximize the posterior probability $p(\theta | \mathcal{D})$, where $p(\theta | \mathcal{D})$ is:
\begin{equation}
    p(\theta | \D) = \frac{p(\D|\theta) p(\theta)}{p(\D)}\,.
\end{equation}

Assuming that the dataset $\mathcal{D}$ is divided into two parts, $D_A$ and $D_B$, where both are independent, then:
\begin{multline}
    p(\theta | \D) = \frac{p(\D|\theta) p(\theta)}{p(\D)} = \frac{p(D_B, D_A|\theta) p(\theta)}{p(D_B, D_A)} 
    = \frac{p(D_B|\theta) p(D_A|\theta) p(\theta)}{p(D_B) p(D_A)} \\ = \frac{p(D_B|\theta)}{p(D_B)}\frac{p(D_A|\theta)p(\theta)}{p(D_A)} = \frac{p(D_B|\theta) p(\theta | D_A)}{p(D_B)}\,
\end{multline}
By applying $log$ to the previous function we get:
\begin{equation}
    \log{p(\theta | \D)} = \log{p(D_B|\theta)} + \log{p(\theta | D_A)} - \log{p(D_B)} \,.
    \label{eqn:log_posterior}
\end{equation}
This means that maximizing the posterior on the dataset $\D$ is equivalent to maximizing the posterior on the subset $D_A$ while minimizing the Negative Log-Likelihood (NLL) of $D_B$ ($-\log{p(D_B | \theta)}$).

Since calculating the posterior $p(\theta | D_A)$ is intractable, \textit{EWC} approximates the posterior as a Gaussian distribution with the mean given by $\theta^*_A$ (the parameters that minimize the NLL on $D_A$), and the diagonal precision given by the diagonal of the Fisher information matrix $F$, which is known as the Laplace approximation \citep{mackay1992practical}. Hence, optimizing the parameters $\theta$ for the posterior in Eq. \eqref{eqn:log_posterior} would be equivalent to minimizing the loss:
\begin{equation}
    \tilde{L}(\theta) = L_B(\theta) + \alpha \sum_{i=1}^{N} F_i (\theta_i - \theta_{A,i}^{*})^2 \,,
    \label{eqn:ewc}
\end{equation}

where $N$ is the total number of parameters ($|\theta|$) and $L_B(\theta)$ is the loss on task $B$ data. \textit{EWC} uses the diagonal terms of the Fisher information matrix, which is equivalent to the positive semi-definite second-order derivative of the loss near a minimum, as a proxy for the importance of each of the weight. The Fisher information matrix is:
\begin{equation}
    F = E[\nabla_{\theta} \log f(x; \theta) \nabla_{\theta} \log f(x;\theta)^T]\,.
\end{equation}
From Eq. \eqref{eqn:regularization_function}, we have:
\begin{equation}\label{ewc_imp}
    \Omega_i^{(\T)} = F_i^{(\T)} = E[\nabla_{\theta_i} \log f(x;\theta)^2|_{\theta^{*}_{\T,i}}] \,,
\end{equation}
where the Fisher information matrix $F$ is calculated as a point estimate at the end of each task. 

When training on subsequent tasks, \textit{EWC} tries to minimize the distance from previous optimal parameters that correspond to each task $t < \T$, where the objective function becomes:
\begin{equation}
    \tilde{L}(\theta) = L_{\T}(\theta) + \alpha \sum_{t < \T} \sum_{i} F_i^{t} (\theta_i - \theta_{t,i}^{*})^2 \,.
    \label{eqn:ewc_afterwards}
\end{equation}
Therefore \textit{EWC} requires keeping all the Fisher matrices from previous tasks $F^{t}$, as well as the optimal parameters for these tasks $\theta_{t}^*$, which gives it a memory complexity that increases linearly with the number of tasks $\T$ ($\mathcal{O}(N\T)$). Algorithm \ref{algo:reg} shows the pseudo code for EWC.

\begin{algorithm}[H]
    \caption{EWC}
    \label{algo:reg}
    \begin{algorithmic}[1]
        \REQUIRE $\alpha, \eta, \theta_{0}, \mathcal{D}_{1} \ldots \mathcal{D}_{T}$
        \ENSURE $\theta^*_{\T}$
        
        \STATE $\theta \leftarrow \theta_0$
        \FOR{$t \leftarrow 1 \ldots \T$}
            \WHILE{Until Convergence}
                \STATE $(X, Y) \leftarrow \text{ Batch from }D_t$
                \STATE $L \leftarrow loss(f(X), Y ; \theta)$
                \IF{$t > 1$}
                    \STATE $\tilde{L} \leftarrow L + \alpha \sum_{t' < t} \sum_{i} F^{t'}_{i} (\theta_i - \theta_{t', i}^*)^2$
                \ELSE
                    \STATE $\tilde{L} \leftarrow L$
                \ENDIF
                \STATE $g \leftarrow \nabla_{\theta} \tilde{L}$
                \STATE $\theta \leftarrow \theta - \eta g$
            \ENDWHILE
            \STATE $\theta^*_t \leftarrow \theta$
            \STATE $F^{t}_{i} \leftarrow \frac{1}{|D_t|} \sum_{(x, y) \sim D_t} \nabla_{\theta} f(x; \theta) \nabla_{\theta} f(x; \theta)^T$
        \ENDFOR
    \end{algorithmic}
\end{algorithm}

The derivation of \textit{EWC} provided in \citet{ewc} uses a two task case. \citet{huszar2017quadratic} extends the Laplace approximation in \textit{EWC} to the multiple tasks leading to the following objective function:
\begin{equation}
    \tilde{L}(\theta) = L_{\T}(\theta) + \alpha \sum_{i} \left[\sum_{t < \T} \lambda^{(t)} F_i^{t}\right] \left(\theta_i - \theta_{\T-1,i}^{*}\right)^2 \,.
\end{equation}
This objective function depends only on the last set of optimal parameters $\theta_{\T-1,i}^{*}$ and the sum of the previous Fisher matrices $\sum_{t < \T} [\lambda^{(t)} F_i^{t}]$. If the above modification is made, its memory complexity is constant in the number of tasks. 

Approximating the Fisher information matrix using only the diagonal terms means that the interactions between the parameters are ignored. \citet{ritter2018online} tries to obtain a better approximation by using the Kronecker product approximation of the Fisher information matrix, and it uses a block diagonal Fisher information matrix rather than a diagonal one. This translates to ignoring the interactions between the parameters across the layers but taking into account the interactions within the same layer at the expense of having a higher computational complexity.

\subsection{EWC++}
\citet{chaudhry2018riemannian} introduce \textit{EWC++}, which is a more efficient version of \textit{EWC}.~\textit{EWC++} uses the KL-divergence between the conditional probabilities $p_{\theta}(y|x)$ and $p_{\theta + \Delta \theta}(y|x)$ as a regularization loss. The conditional probability is represented by the neural network function $f(x; \theta) := p_{\theta}(y|x)$. 

Here we follow the derivation for the $D_{KL}(p_{\theta} || p_{\theta + \Delta \theta})$ provided in \citet{chaudhry2018riemannian} that shows the relationship between the KL divergence and the Fisher information matrix.

Let's first start by defining the shorthand notations $p_{\theta}(z) = p_{\theta}(y|x)$ and $\mathbb{E}_z[.] = \mathbb{E}_{x \sim \mathcal{D}, y \sim p_{\theta}(y | x)}[.]$, then we have:
\begin{equation}
    D_{KL}\left(p_{\theta}(z) || p_{\theta + \Delta \theta}(z)\right) = \mathbb{E}_z \left[\log{p_{\theta}(z)} - \log{p_{\theta + \Delta \theta}(z)}\right] \,.
    \label{eqn:kl}
\end{equation}
Using the second order Taylor expansion ($z$ is omitted for brevity), we have:
\begin{equation}
    \log{p_{\theta + \Delta \theta}} \approx \log{p_{\theta}} + \Delta \theta ^{\top} \frac{\partial \log p_{\theta}}{\partial \theta} + \frac{1}{2} \Delta \theta ^{\top} \frac{\partial^2 \log p_{\theta}}{\partial \theta^2} \Delta \theta \,.
\end{equation}
By substituting this in Eq. \eqref{eqn:kl}, we get:

\begin{align}
    D_{KL}(p_{\theta} || p_{\theta + \Delta \theta}) &= \mathbb{E}_z \left[\log{p_{\theta}}\right] - \mathbb{E}_z\left[\log{p_{\theta + \Delta \theta}}\right] \nonumber \\
    &\approx \mathbb{E}_z \left[\log{p_{\theta}}\right] - \mathbb{E}_z\left[\log{p_{\theta}}\right] - \Delta \theta ^{\top} \mathbb{E}_z\left[\frac{\partial \log p_{\theta}}{\partial \theta}\right] \\
    & - \frac{1}{2} \Delta \theta ^{\top} \mathbb{E}_z\left[\frac{\partial^2 \log p_{\theta}}{\partial \theta^2}\right] \Delta \theta  \nonumber \\
    &= - \Delta \theta ^{\top} \mathbb{E}_z \left[ \frac{\partial \log p_{\theta}}{\partial \theta} \right]  - \frac{1}{2} \Delta \theta ^{\top} \mathbb{E}_z \left[ \frac{\partial^2 \log p_{\theta}}{\partial \theta^2} \right] \Delta \theta \, \label{eqn:dkl_approx}
\end{align}

Here, the first term cancels out:
\begin{align}
    \mathbb{E}_z\left[\frac{\partial \log p_{\theta}(y|x)}{\partial \theta}\right] &= \mathbb{E}_{x \sim \mathcal{D}}\left[\sum_{y} p_{\theta}(y|x)\frac{\partial \log p_{\theta}(y|x)}{\partial \theta}\right]  \nonumber \\
    &=\mathbb{E}_{x \sim \mathcal{D}}\left[\sum_{y} p_{\theta}(y|x) \frac{1}{p_{\theta}(y|x)} \frac{\partial p_{\theta}(y|x)}{\partial \theta}\right]  \nonumber \\
    &= \mathbb{E}_{x \sim \mathcal{D}}\left[\sum_{y} \frac{\partial p_{\theta}(y|x)}{\partial \theta}\right] \, \nonumber \\
    &= \mathbb{E}_{x \sim \mathcal{D}}[\frac{\partial}{\theta} \sum_{y} p_{\theta}(y|x)] =  \mathbb{E}_{x \sim \mathcal{D}}[0] \nonumber \\
    \mathbb{E}_z\left[\frac{\partial \log p_{\theta}(y|x)}{\partial \theta}\right] &= 0 \,.
\end{align}
This means that Eq. \eqref{eqn:dkl_approx} simplifies to:
\begin{equation}
    D_{KL}(p_{\theta} || p_{\theta + \Delta \theta}) \approx - \frac{1}{2} \Delta \theta ^{\top} \mathbb{E}_z[\frac{\partial^2 \log p_{\theta}}{\partial \theta^2}] \Delta \theta] = - \frac{1}{2} \Delta \theta ^{\top} H \Delta \theta \approx \frac{1}{2} \Delta \theta ^{\top} F \Delta \theta \,,
\end{equation}
where $F$ is the empirical Fisher information matrix and $F = -H$ at the maximum likelihood estimate. As $F$ is prohibitively expensive, the diagonal Fisher is used instead (which assumes that the parameters are independent), which gives:
\begin{equation}
    D_{KL}(p_{\theta} || p_{\theta + \Delta \theta}) \approx \sum_{i} F_i \Delta \theta_i^2\,.
\end{equation}
\citet{chaudhry2018riemannian} proposes \textit{EWC++} which uses the $D_{KL}$ as the regularization term. After task $\T$ is introduced, we have:
\begin{equation}
    \tilde{L}(\theta) = L_{\T}(\theta) + \alpha D_{KL}(p_{\theta} || p_{\theta_{\T-1}^*}) \approx L_{\T} + \alpha \sum_{i} F_i (\theta_i -\theta^*_{\T-1, i})^2 \,.
    \label{eqn:ewcplus}
\end{equation}
It has to be noted that many of the approximations that were performed were based on the assumption that $\Delta \theta$ is small. If the new $\theta$ diverges away from the old $\theta$, these approximations become very imprecise, but that should not happen given that $\theta$ is already constrained in the objective function to stay as close to the older $\theta$ as possible.

We can see from Eq. \eqref{eqn:ewcplus} that \textit{EWC++} is equivalent to \textit{EWC} (Eq. \eqref{eqn:ewc}) when $\T$ represents the second task, while they start to diverge afterwards (Eq. \eqref{eqn:ewc_afterwards}). \textit{EWC++} is more efficient than \textit{EWC} as it only needs to keep a single Fisher matrix and a single set of parameters $\theta_{\T-1}$, which gives it a constant memory complexity.

Another difference between \textit{EWC} and \textit{EWC++} is that in \textit{EWC++}, the Fisher matrix is updated in an online fashion, and hence no extra pass over the dataset is needed after the task $\T$ is done. After each minibatch, the diagonal Fisher matrix $F$ is updated as follows:
\begin{equation}
    F = \gamma F_{new} + (1 - \gamma) F_{old}
\end{equation}

The pseudo code for \textit{EWC++} is given in Algorithm \ref{algo:ewcplusplus}.
\begin{algorithm}[H]
    \caption{EWC++}
    \label{algo:ewcplusplus}
    \begin{algorithmic}[1]
        \REQUIRE $\alpha, \eta, \gamma, \theta_{0}, \mathcal{D}_{1} \ldots \mathcal{D}_{T}$
        \ENSURE $\theta^*_{\T}$
        
        \STATE $\theta \leftarrow \theta_0$
        \STATE $F \leftarrow I$
        \FOR{$t \leftarrow 1 \ldots \T$}
            \WHILE{Until Convergence}
                \STATE $(X, Y) \leftarrow \text{ Batch from }D_t$
                \STATE $L \leftarrow loss(f(X), Y ; \theta)$
                \STATE $F_{old} \leftarrow F$
                \STATE $F_{new} \leftarrow \frac{1}{|X|} \sum_{x \in X} \nabla_{\theta} f(x; \theta) \nabla_{\theta} f(x; \theta)^T$
                \STATE $F = \gamma F_{new} + (1 - \gamma) F_{old}$
                \IF{$t > 1$}
                    \STATE $\tilde{L} \leftarrow L + \alpha \sum_{i} F_{i} (\theta_i - \theta_{t-1, i}^*)^2$
                \ELSE
                    \STATE $\tilde{L} \leftarrow L$
                \ENDIF
                \STATE $g \leftarrow \nabla_{\theta} \tilde{L}$
                \STATE $\theta \leftarrow \theta - \eta g$
            \ENDWHILE
            \STATE $\theta^*_t \leftarrow \theta$
        \ENDFOR
    \end{algorithmic}
\end{algorithm}

\subsection{Synaptic Intelligence} 
Synaptic Intelligence (\textit{SI}) \citep{pmlr-v70-zenke17a} tries to measure the importance of the parameters by estimating how much each parameter affects the loss trajectory. In more concrete terms, the contribution of each parameter $\theta_i$ to the loss during the current task $\T$ is defined as:
\begin{equation}
    \omega_i^{(\T)} = \int_{t'_{\T-1}}^{t'_{\T}} \frac{\partial L_{\T}}{\partial \theta_{i}} \theta_i'(t') dt'\,,
\end{equation}
where $t'$ is the time step and $\theta_i'(t')dt'$ contributes to the the parameter change from the initial point (at time $t'_{\T-1}$) to the final point (at time $t'_{\T}$). Since gradient descent uses discrete time steps to perform the updates, the effect of each parameter during task $\T$ ($\omega_i^{(\T)}$) is, in practice, calculated as an online running sum of the product of the gradient with the parameter update $\theta_{\T, i}'(t')$, where $\omega_i^{(\T)}$ is initialized to zero for each new task $\T$.

The parameter importance $\Omega_i^{(\T)}$ from ~\cref{eqn:regularization_function} is calculated to be directly proportional to the contribution $\omega_i^{(\T)}$ of each parameter to the loss during the task $\T$, normalized with the square of final change $\theta_i$ needed to make during $\T$ to avoid large changes to important parameters (similar to Eq. \eqref{ewc_imp}):
\begin{equation}
    \Omega_i^{(\T)} = \sum_{t < \T} \frac{\omega_i^{(t)}}{(\theta^*_{t,i} - \theta^*_{t-1, i})^2 + \epsilon}
\end{equation}
where $\epsilon$ is the damping parameter used to bound the expression in case ($\theta^*_{t,i} - \theta^*_{t-1, i}) \xrightarrow{} 0$.

\subsection{Memory Aware Synapses (MAS)}
We have seen in Section~\ref{sec:ewc} (Elastic Weight Consolidation) that the importance of the parameter is inversely related to its uncertainty, for which the Fisher information matrix acts as a proxy. Hence a parameter with a higher precision is a more important parameter (from a probabilistic point of view as in Eq. \eqref{eqn:log_posterior}). \citet{mas}, on the other hand, try to estimate the importance of a parameter based on how sensitive the learned function $F_{\T-1}$ is to a change in that parameter. This removes the dependence on a labeled set to estimate the importance, and hence an unlabeled set can be used to estimate the parameter importance. Hence for a specific output class $c$, and a set of unlabeled dataset (or just the new task set) $D$, the sensitivity would be measured using the following equation:
\begin{equation}
    \Omega_i = \frac{1}{|D|} \sum_{x \sim D} \frac{\partial F_c(x; \theta)}{\partial \theta_i} \,.
\end{equation}
As an alternative to computing the importance per output/class, \citet{mas} propose using the $L_2$ norm of all the output nodes $||F(x;\theta)||_2^2$ as a representative for all the outputs. Hence, the alternative equation would be (similar to Eq. \eqref{ewc_imp}):
\begin{equation}
    \Omega_i = \frac{1}{|D|} \sum_{x \sim D} \frac{\partial ||F(x; \theta)||_2^2}{\partial \theta_i} \,.
\end{equation}
Calculating the importance $\Omega_i$ this way allows for decoupling the updates of the importance parameter $\Omega$, and the training on the task itself, since there is no dependence of $\Omega$ on the task data. As mentioned earlier, $D$ can be an unlabeled set of samples that are fixed across the tasks, but it also allows the use of an online stream of samples.

\citet{benzing2021unifying} shows that although \textit{SI} and \textit{MAS} are motivated differently than \textit{EWC}, as elaborated earlier, they both approximate the square root of the Fisher Information Matrix, which gives a unified picture for the three importance-based methods.

There have been some other recent advances in importance-based regularization techniques in lifelong learning.  For example, building upon MAS \citep{mas},  Importance Driven Continual Learning approach \citep{ozgun2020importance} defines a parameter-specific learning rate such that the learning rate becomes a function of the parameter’s importance.

\citet{jung2020continual} proposed Adaptive Group Sparsity based lifelong learning that introduced a loss function based on group-sparsity norms for parameter-wise importance regularization in neural networks. 

\section{Bayesian-Based Regularization}
Bayesian-based regularization can be considered a special type of importance-based regularization.  some scenarios, we don't have access or are not allowed to store previously seen data due to privacy or security restrictions. In such scenarios, the lifelong learning algorithm's goal is to train a model at the current task using training data related to the current task without revisiting the training data from the previous task and reducing catastrophic forgetting through time. In this section, we revisit this scenario from a Bayesian inference perspective. The goal of the model for the first task is to predict a set of targets denoted as $Y^{(t_1)}$, where $t_1$ is the task ID, in a supervised manner using parameters $\theta$, with a group of hyperparameters that the model uses to reach its highest performance. For a set of $n$ i.i.d. samples used for training in the first task, the joint probability distribution of $Y^{(t_1)}$ and model's parameters, given the hyperparameters used in the training procedure, can be formalized as follows:
\begin{equation}
    p\left(\mathrm{Y}^{(t_1)}, \theta \mid \alpha, X\right)\,,
\end{equation}
where $X$ is the set of observation for task $t_1$ and $\alpha$ represents the hyperparameters. Computing the integral over $\theta$ gives the desired marginal distribution $\int p\left(\mathrm{Y}^{(t_1)}, \theta \mid \alpha, X\right)\mathrm{d} \theta = p\left(\mathrm{Y}^{(t_1)}\mid \alpha, X\right)$. By dividing and normalizing the joint distribution, we can also get the posterior distribution as $p\left(\theta \mid Y^{(t_1)}, \alpha, X\right)$. The model can be trained to predict both the posterior distribution and marginal distribution: (i) predicting the targets given by the best hyperparameters and (ii) having the best distribution of the model parameters given by targets and the set of observations. For lifelong learning, the posterior distribution can be obtained by multiplying the previous posterior by the likelihood of the dataset belonging to the new task and the regularization term can be interpreted as a prior.

We cannot compute the posterior distribution directly since we do not have any knowledge of the model parameters. However, we can expand the probability distribution using Bayesian inference and, from that, try to find the best distribution for model parameters. So, according to the Bayesian inference for $n$ i.i.d. training samples:
\begin{equation}
    p\left(\mathrm{Y}^{(t_1)}, \theta \mid \alpha, X\right)=\left[\prod_{i=1}^{n} p\left(y_{i}^{(t_1)} \mid \theta, \alpha, x_{i}\right)\right] p(\theta \mid \alpha, X)\,,
    \label{v_bayesian_inference}
\end{equation}
where $(x_{i}, y_{i})$ is a input and and target pair from the observation set. 

\subsection{Approximate Inference in Lifelong Learning}
In Eq. \eqref{v_bayesian_inference}, calculating $p(\theta \mid \alpha, X)$ is intractable in most cases i.e., it is not computationally possible to perform exact inference when the dimension of $\theta$ is high. This motivates us to approximate the exact posterior distribution by another distribution that is computationally easier to handle. 
We can approximate $p(\theta \mid \alpha, X)$ with a posterior distribution using prior knowledge that we have over $\theta$ by assuming that it comes from a Gaussian distribution. With this assumption, we can get $q^{*}(\theta \mid \alpha, X)$ as a posterior of the $p(\theta \mid \alpha, X)$ using the Hidden Markov Model (HMM) or Gaussian processes approaches. Therefore, we can define the joint probability distribution for the second task as follows:
\begin{equation}
    p\left(\mathrm{Y}^{(t_2)}, \mathrm{Y}^{(t_1)}, \theta \mid \alpha, X\right)=\left[\prod_{i=n+1}^{n+m} p\left(y_{i}^{(t_2)} \mid \theta, \alpha, x_{i}\right)\right]p\left(\mathrm{Y}^{(t_1)}, \theta \mid \alpha, X\right)
    \label{v_task_1}
\end{equation}
where $n$ and $m$ refer to number of samples in tasks $t_1$ and $t_2$ respectively. Substituting $p\left(\mathrm{Y}^{(t_1)}, \theta \mid \alpha, X\right)$ according to Eq. \eqref{v_bayesian_inference} we have:
\begin{equation}
\begin{aligned} p\left(\mathrm{Y}^{(t_2)}, \mathrm{Y}^{(t_1)}, \theta \mid \alpha, X\right) &=\left[\prod_{i=n + 1}^{n+m} p\left(y_{i}^{(t_2)} \mid \theta, \alpha,x_{i}\right)\right]\left[\prod_{i=1}^{n} p\left(y_{i}^{(t_1)} \mid \theta, \alpha,x_{i} \right)\right] \\ 
&~~~~ \times p(\theta \mid \alpha, X) \\ & \approx\left[\prod_{i=n+1}^{n+m} p\left(y_{i}^{(t_2)} \mid \theta, \alpha, x_{i}\right)\right] q_{1}^{*}(\theta, X) \\ & \approx q_{2}^{*}(\theta, X) \end{aligned}
\end{equation}
The approximating distribution is usually chosen to be a product of several independent distributions, one for each parameter or a set of similar parameters. Such methods have been used for solving various inference problems in machine learning. 

There are several approaches for approximate inference, including Moment matching \citep{li2015generative}, variational KL minimization \citep{hershey2007variational}, Taylor expansion, Importance Sampling \citep{tokdar2010importance} and Laplace’s approximation \citep{friston2007variational}. 

Approximate inference can be used to alleviate catastrophic forgetting in the lifelong learning setup. Proposed lifelong learning methods using approximate inference can be categorized as prior-focused methods. To this end, the main focus is on approximating the model's parameters distribution. In a lifelong learning setup, we can approximate the model's parameters recursively using the optimal parameters at previous tasks. This intuition and approach to find the best model parameters can show the importance of Approximate Inference to propose new methods in this setup. Both Taylor expansion and variational KL minimization can be used to alleviate catastrophic forgetting in lifelong learning methods. Next, we describe the proposed lifelong learning methods that employ these concepts in detail.

\subsection{Variational KL Minimization}
As we discussed in the EWC method in Section \ref{sec:regularizzation_ewc}, the goal of regularization techniques is to tune the model parameters such that the parameters do not deviate from model parameters in the previous task. EWC alleviates the catastrophic forgetting by forcing model parameters to move around the last task parameters space considering the parameters' importance using Fisher Information Matrix. To reach the same goal, Variational Continual Learning (VCL) method proposed an alternative way by using Approximate inference. To not deviate too much from previously learned parameters' space, we can use the variational KL minimization to reduce the parameters' distribution distance from what the model learned in the previous task~\citep{nguyen2017variational}.  

Before showing the advantage of using variational KL minimization to approximate the parameters' posterior distribution, we review some basic divergence measures such as the Kullback-Leibler (KL) divergence that is used to measure the closeness of the two distributions. It is defined as: 

\begin{equation}
    KL(q||p) = \mathbb{E}_q[log(\frac{q(\theta)}{p(\theta)})]\,,
\end{equation}

where $KL(q||p)$ indicates $p$'s divergence from $q$. Intuitively, when $p(\theta)$ is large, but $q(\theta)$ is small, there is a large divergence. When $p(\theta)$ is small and $q(\theta)$ is large, there will a again be a large divergence, but not as large as the previous case.

VCL, as an influential method in prior-focused approaches, computes the posterior distribution of the parameters given the previous examples and keeps changing it over time. Computing the posterior distribution can be simplified using the mean-field approximation. VCL finds the new posterior for task $t$ by that minimizes the KL-divergence with the old posterior at time step $t-1$ as follows:

\begin{align}
    \tilde{q}_{t}(\theta)= \arg \min _{q \in Q} KL(q(\theta) \| \frac{1}{\tilde{Z}_t} \tilde{q}_{t-1}(\theta)p\left(\mathcal{D}_{t}   \mid \theta\right))
\end{align}
where ${D}_{t}$ is the training data at time $t$, $Z_t$ represents the intractable normalizing constant ~\citep{nguyen2017variational}. VCL predicts the targets for the test set inputs denoted as $x^{*}$ as follow:
\begin{align}
    p\left(y^{*} \mid \boldsymbol{x}^{*}, \mathcal{D}_{1: t}\right)=\int q_{t}(\theta) p\left(y^{*} \mid \theta, \boldsymbol{x}^{*}\right) \mathrm{d} \theta,
\label{equ:vcl_prediction}
\end{align}
where $\mathcal{D}_{1: t}$ represents the data from the beginning to the end of time $t$. 
\begin{figure}
    \centering
    \includegraphics[width=\textwidth]{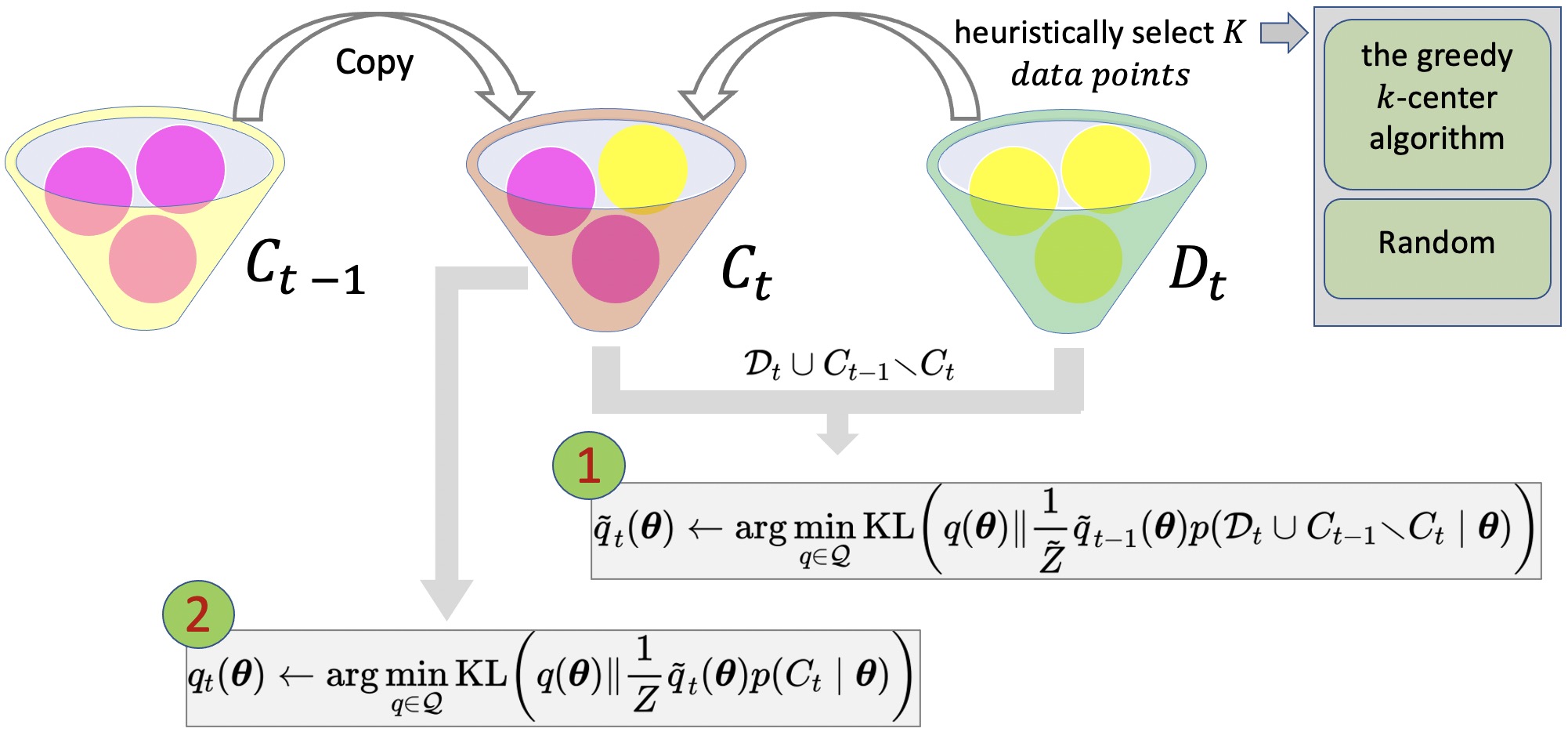}
    \caption{The summary of the  Variational Continual Learning (VCL) algorithm.}
    \label{fig:vcl_summary}
\end{figure}
To get the advantage of keeping some samples from the previous task in the replay, VCL proposes using a replay buffer called Coreset in the VCL-Coreset version. Figure~\ref{fig:vcl_summary} illustrates the VCL-Coreset algorithm. The VCL-Coreset observes the current task data denoted as $D_t$. It updates the coreset combining the information currently existing in the Coreset and $D_t$ denoted as $C_t$. Then VCL updates the variational distribution for all samples in $\mathcal{D}_t \cup C_{t-1} \setminus C_t$ as marked $1$ in the Figure~\ref{fig:vcl_summary}. VCL uses the sample in the $C_t$ to compute the final variational distribution, which is used only for prediction and not propagation as marked by $2$ in the figure~\ref{fig:vcl_summary}. VCL uses the Eq. \eqref{equ:vcl_prediction} to perform prediction at test time. 

To create a new Coreset $C_{t}$ at time $t$,  VCL selects new data points from the current task and a selection from the old coreset $C_{t-1}$ as shown in figure~\ref{fig:vcl_summary}. To select samples, any heuristic, including greedy approaches or simple random selection, can be used to select $K$ data points from $\mathcal{D}_{t}$ and added to $C_{t-1}$. It helps to have an unbiased coreset for computing the model parameters for the current task.

Following the VCL, Functional Regularisation for Continual Learning with Gaussian Processes (FRCL)~\citep{titsias2019functional} uses Gaussian process for a function family ${f}=\left(f_{1}, \ldots, f_{n}\right)$ such that functions sampled from $\mathcal{N}({\mu},\;\Sigma)$ where ${\mu}$ and $\Sigma$ is defined by a mean function \(\operatorname{\textit{mean({x})}}\) and covariance function \(\mathrm{K}\left(\mathrm{x}, \mathrm{x}^{\prime}\right)\) such that \(f({x}) \sim \mathcal{G} \mathcal{P}\left(\operatorname{mean}({x}), \;\mathrm{K}\left({x}, \;{x}^{\prime}\right)\right)\). In FRCL each function $f_{i}$ defined as follow: 
\begin{align}
    & f_{i}\left(x ; w_{i}\right) \equiv f_{i}\left(x ; w_{i},\; \theta\right)=w_{i}^{\top} \phi(x ;\;\theta) \\ 
    & \text{\quad and \quad} w_{i} \sim \mathcal{N}\left(w_{i} \mid 0,\;\sigma_{w}^{2} I\right), 
\end{align}
where $\Omega_{i}$ is the task-specific weights and $\phi(x ;\;\theta)$ represents the shared feature vector. FRCL should maximize $\mathcal{F}\left(\theta,\;q\left(w_{k}\right)\right)$ as a learning objective that is computed as follow:
\begin{align*}
    \mathcal{F}\left(\theta,\; q\left(w_{k}\right)\right)&=\sum_{j=1}^{N_{k}} \mathbb{E}_{q\left(w_{k}\right)}\left[\log p\left(y_{k, j} \mid w_{k}^{\top} \phi\left(x_{k, j} ; \;\theta\right)\right)\right]\\
&-\operatorname{KL}\left(q\left(w_{k}\right) \| \;p\left(w_{k}\right)\right) \\
&-\sum_{i=1}^{k-1} \operatorname{KL}\left(q\left({u}_{i}\right) \|\;p_{\theta}\left({u}_{i}\right)\right) 
\end{align*}
where,
\begin{align*}
    & f_{k}\left(x ; w_{k}\right)=w_{k}^{\top} \phi(x ;\;\theta),\\
    & w_{k} \sim \mathcal{N}\left(0,\;\sigma_{w}^{2} I\right) \\ 
    & q\left(w_{k}\right)=\mathcal{N}\left(w_{k} \mid \mu_{w_{k}},\;\Sigma_{w_{k}}\right)
\end{align*}
The $-\sum_{i=1}^{k-1} \operatorname{KL}\left(q\left({u}_{i}\right) \| \;p_{\theta}\left({u}_{i}\right)\right)$ term is also considered as the regularisation term ($R_{\T}$ in Eq. \eqref{regularize}) that is computed for the previous tasks~\citep{titsias2019functional}. FRCL uses the $KL$ term to distinguish the task boundaries such that if $\mathrm{KL} \gg 0$ shows the task shift and $\mathrm{KL} \approx 0$ shows that model is still in the same task. 

Recently, Uncertainty-regularized Continual Learning proposed by \citep{ahn2019uncertainty} builds on a Bayesian learning framework with variational inference with the notion of node-wise uncertainty. The authors perform an interpretation of the closed-form of the KL-divergence term for the Gaussian mean-field approximation and the Bayesian neural network pruning that reduces the number of additional parameters for implementing per-weight regularization. On the other hand, Uncertainty- guided Continual Bayesian Neural Networks \citep{ebrahimi2019uncertainty} introduced a learning rate that adapts according to the uncertainty defined in the probability distribution of the weights in networks and retains task performance after pruning weights by saving binary masks per task. \citet{kumar2021bayesian} applies variational Bayesian-based regularization for both discriminative and generative settings by learning priors from previous tasks.

\section{Distillation-based Regularization}
\label{sec:distill}

Distillation-based regularization is mainly based on the following premises: if the network has access to the samples of task 1, and was forced to produce the same output on these samples while training on task 2, then the performance on task 1 would be preserved and no catastrophic forgetting would take place. However, the samples of task 1, evidently, are not there anymore, but since images lie on a low dimensional manifold \citep{pless2009survey}, images of the new task may provide some sort of sampling of the images of task 1, and hence, preserving the network output on these images for the classes of task 1 may keep the performance on task 1 from degrading, depending on how similar the images are between the two tasks. As this mechanism is similar to knowledge distillation \citep{gou2021knowledge}, where the teacher network is the network trained on task 1 and the student network is the network being trained on task 2, we shall call it distillation-based regularization.

\subsection{Learning Without Forgetting}
\label{sec:lwf}

The distillation based regularization in the context of lifelong learning was introduced by Learning Without Forgetting (LWF)~\citep{2017_learning_without_forgetting}. A copy of the model that was trained on the last task ($f_{\T-1}$) is saved. When a new task $\T$ is introduced, the output of $f_{\T-1}$ is used as a soft target for the new model $f_{\T}$ to imitate. This happens for the classes that are shared between the two models (the classes that belong to the previous tasks). This idea is inspired from knowledge distillation \citep{hinton2015distilling}, where the model trained on the previous task $f_{\T-1}$ is the teacher model, and the model being trained on the current task $f_{\T}$ is the student model. Since the student model is only trained on the current task data, LWF is making the assumption that the samples of the current task might provide a poor sampling for the older tasks. Hence, if there is visual similarity between the previous tasks and the new tasks, this mechanism can help in alleviating catastrophic forgetting. iCaRL \citep{Rebuffi_2017} extends LWF to the case when there is a replay buffer that provides samples for the older tasks (\cref{sec:memory}). 

In more concrete terms, the objective function that LWF tries to minimize when training on task $\T$ is:
\begin{equation}
    \tilde{L}(\theta) =  \underbrace{-\sum_{c \in \mathcal{C}^{(\T)}} \delta_{c = y} \log{f_c(x; \theta)}}_{L_{\T}(\theta) \text{ in Eq. \eqref{regularize}}} ~~ \underbrace{-\sum_{c \in \mathcal{Y}^{(\T-1)}} f_c(x; \theta^{*}_{\T-1}) \log{f_c(x; \theta)}}_{R_{\T} \text{ in Eq. \eqref{regularize}}} \,
\end{equation}\label{lwf}
Along with the cross-entropy loss for the new task (first part), the knowledge distillation loss (second part) is incorporated to impose output stability of old tasks with new data. A temperature term might be multiplied to the distillation loss in order to control how sensitive is the distillation loss to the difference between $f_c(x; \theta^{*}_{\T-1})$ and $f_c(x; \theta)$.

\subsection{Learning without Memorization}
\label{sec:lwm}
Learning without Memorization (LWM)~\citep{dhar2019learning} claims that imitating the output of the previous model is not enough, but also the model has to remember where to look at; what are the regions in the image that it used to look at before. The intuition is that the new model $f_{\T}$ has no extra information about the older classes than the old model $f_{\T-1}$, and hence it makes no sense to look elsewhere when evaluating the older classes. In addition to the distillation on the output of the teacher model, it applies a distillation loss on the attention maps of the teacher model for the classes that belong to the old tasks as following:
\begin{equation}
    \tilde{L}(\theta) = L_{\T}(\theta) + \beta L_D(\theta) + \gamma L_{AD}(\theta)
\end{equation}
where $\beta$ and $\gamma$ are regularization parameters, $L_D(\theta)$ is the distillation loss as used in LWF \eqref{lwf} and $L_{AD}(\theta)$ is the attention distillation loss that is defined as the sum of element wise $L_1$ difference of the normalized, vectorized attention map (generated using Grad-CAM \citep{selvaraju2017grad}). Therefore, the attention maps represent the regions in the image which resemble the base classes. 

Deep Model Consolidation (DMC) \citep{zhang2020class} leverages the unlabeled auxiliary data instead of old training data to ensure the student model absorbs the knowledge in an unbiased way. Essentially, it learns a new network for each new task and then trains a new model on the outputs of the old and new networks on some unlabeled data to promote symmetric knowledge transfer. The premise is that if natural images lie on a low-dimensional manifold, then the unlabeled data from a similar domain will provide some representation for the datasets of the learned tasks.

Recent knowledge distillation-based regularization methods in lifelong learning includes Batch-level Distillation (BLD) \citep{fini2020online} that adopts a dynamic weighting strategy while minimizing the memory overhead. Similarly, \citet{zhong2021discriminative} proposed a discriminative distillation approach by adding an expert classifier whose knowledge is distilled to the new classifier to discriminate features between confusing classes in lifelong learning. \citet{douillard2021plop} proposed a multi-scale pooling distillation approach to preserve long and short-range spatial relationships at the feature level. To avoid catastrophic forgetting of the old classes, they perform an entropy-based pseudo-labeling of the background for the classes predicted by the old model.

\section{Optimization Trajectory based Regularization}
Another direction in the lifelong learning research is to look at the problem from the perspective of the optimization. \citet{mirzadeh2020understanding} has shown that the geometric nature of the local minima reached by the learning algorithm affects how much the model is affected by catastrophic forgetting.  Let $w_i^*$ be the minima obtained after sequential training on the $i^{th}$ task and $L_j(w_i)$ as the loss of $j^{th}$ task with parameters $w_i$. Then forgetting of the first task $F_1$ after training on the second task is defined as:
\begin{equation}
	F_1 \coloneqq L_1(w_2^*) - L_1(w_1^*) \label{eq:forgetting}
\end{equation}
According to the second-order Taylor expansion of $L_1(w_2^*)$ around $w_1^*$:
\begin{align}
	L_1(w_2^*) &\approx L_1(w_1^*) + \frac{1}{2} (w_2^* - w_1^*)^{\top} \nabla^2 L_1(w_1^*)(w_2^*-w_1^*) \label{eq:taylorapprox}
\end{align}
By using the above approximation to $L_1(w_2^*)$, \citet{mirzadeh2020understanding} derived an upper bound to $F_1$ in terms of the maximum eigen value ($\lambda_1^{max}$) of $\nabla^2 L_1(w_1^*)$:
\begin{align}
    F_1 &\approx \frac{1}{2} (w_2^* - w_1^*)^{\top} \nabla^2 L_1(w_1^*)(w_2^*-w_1^*) \nonumber \\
    &\leq \frac{1}{2} \lambda_1^{max}||w_2^*-w_1^*||_2^2, \label{eq:forgettingbound}
\end{align}.

According to the bound in \cref{eq:forgettingbound}, smaller $\lambda_1^{max}$ means lesser forgetting. 
$\lambda_1^{max}$ has been used to characterize the width of the local minima -- small values correspond to flat minima, and large values correspond to sharp/ narrow minima \citep{hochreiter1997flat}. Based upon the above analysis and empirical results on existing benchmarks, \citet{mirzadeh2020understanding} conclude that flatter minima lead to lesser forgetting. Building on this result, \citet{mehta2021empirical} has shown that models initialized with pre-trained weights undergo lesser forgetting than random weights as pre-trained models have an inductive bias towards flat task minima.

\subsection{Sharpness Aware Minimization (SAM)}
To promote flat minima during lifelong learning, \citet{mirzadeh2020understanding} suggests modifying training dynamics by varying hyper-parameters like batch size, learning rate, and dropout regularization. It is well known that these hyper-parameters influence the optimization trajectory and loss curvature around minima \citep{xie2020diffusion}, also the variance of the gradients \citep{jastrzebski2019break}, implicitly leading to flatter minima for certain values. However, searching for appropriate hyper-parameters for lifelong learning is ill-defined as one does not know task sequence apriori. To address these limitations, \citet{mehta2021empirical} proposes to explicitly optimize for the loss sharpness (alternatively flatter minima) during lifelong learning. Specifically, they employ a Sharpness-Aware Minimization (SAM) procedure \citep{foret2020sharpness} to simultaneously minimize task loss value and loss sharpness. 

SAM searches for parameters that lie in neighborhoods with uniformly low loss regions by minimizing the following loss sharpness (for model $f$ with parameters $w$):
\begin{align}
\max_{||\epsilon||_2 \leq \rho} f(w+\epsilon) - f(w),
\end{align}
where the maximization region is defined to be $\ell^2$ ball with radius $\rho$ around $w$. Formally, the SAM procedure comprises solving the following minimax optimization problem: 
\begin{align}
& \min_w \underbrace{f(w)}_{\text{task loss}} +  \underbrace{\max_{||\epsilon||_2 \leq \rho} f(w+\epsilon) - f(w)}_{\text{loss sharpness}}  + \underbrace{\lambda ||w||_2^2}_{\text{L2 regularization}} \nonumber \\
& \min_w \max_{||\epsilon||_2 \leq \rho} f(w+\epsilon) + \lambda ||w||_2^2 \label{eq:minimaxsam}
\end{align}
Let $\hat{\epsilon}(w)$ denotes the solution to the inner maximization problem in \cref{eq:minimaxsam}. By using first-order Taylor expansion of $f(w+\epsilon)$ w.r.t $\epsilon$ around $0$ and solving for dual norm problem, \citet{foret2020sharpness} derives $\hat{\epsilon}(w)$ to be:
\begin{align}
    \hat{\epsilon}(w) = \rho \frac{\nabla_w f(w)}{||\nabla_w f(w)||_2} \label{eq:sameps}
\end{align}
By using the value of $\hat{\epsilon}(w)$ from \cref{eq:sameps}, the gradient for the minimax problem in \cref{eq:minimaxsam} is approximated as:
\begin{align}
    \nabla_w \max_{||\epsilon||_2 \leq \rho} f(w+\epsilon) & \approx \nabla_w f(w) |_{w + \hat{\epsilon}(w)} + \frac{\partial \hat{\epsilon}(w)}{\partial w} \nabla_w f(w)|_{w+\hat{\epsilon}(w)} \nonumber \\
    & \approx \nabla_w f(w)|_{w + \hat{\epsilon}(w)} \text{ (dropping the second order gradient term) }
\end{align}
\subsection{Orthogonal Gradient Descent (OGD)} 
\begin{figure}[!tb]
    \centering
    \includegraphics[width=0.8\textwidth]{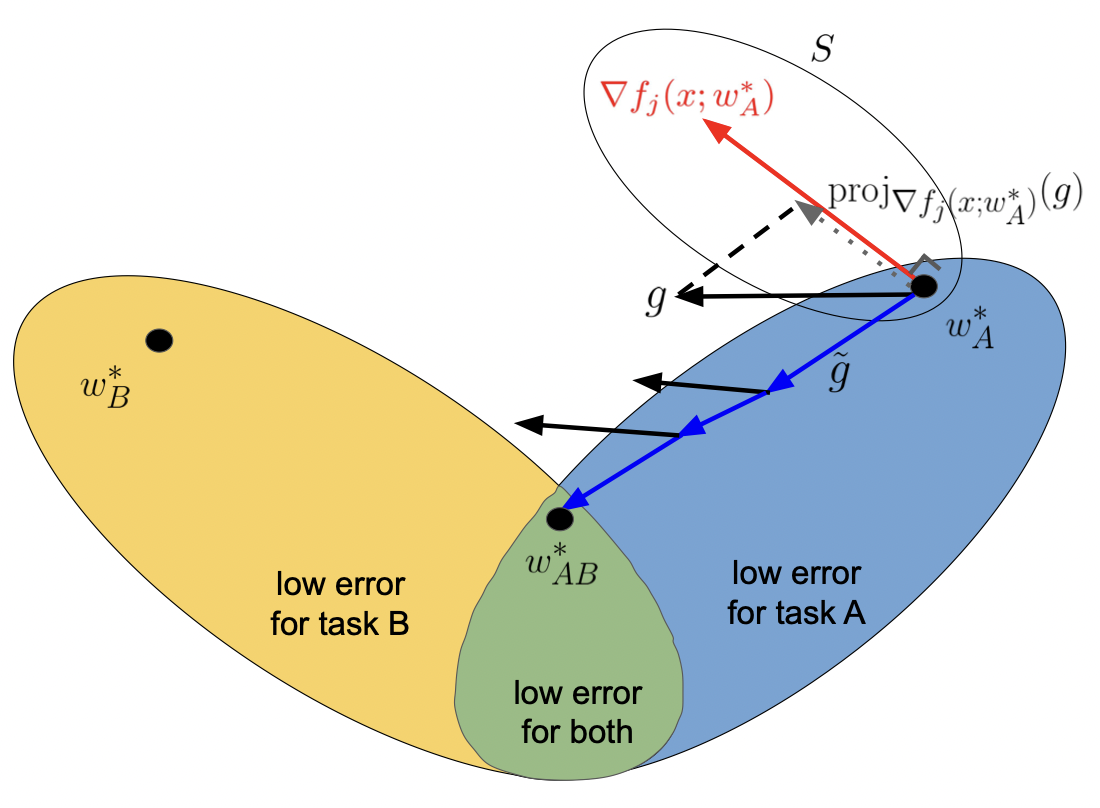}
    \caption{Orthogonal Gradient Descent}
    \label{fig:ogd}
\end{figure}

OGD proposed by \citet{farajtabar2020orthogonal} works on the optimization trajectory by projecting the gradients of the new task $\T$ in a direction that is still useful for learning the task $\T$, but that does not change the predictions on the older tasks $t < \T$. In other words, it tries to maintain a space of the gradient directions of the predictions in the previous tasks, and project the gradients in the new task in a direction perpendicular to that space (Figure~\ref{fig:ogd}).

In more concrete terms, given a task $A$ which contains $c$ classes and a dataset $D_A$ containing $n_A$ samples, we would have $n_A \times c$ gradient directions $\nabla_{\theta} f_j(x; \theta) \forall j \leq c$, where $f_j$ is the model output for class $j$. During training on task $B$ when obtaining the gradient $g$, \textit{OGD} tries to find the gradient direction $\tilde{g}$ such that:
\begin{equation}
    \tilde{g} \perp \nabla_{\theta} f_{j}(x ; \theta) \quad \forall \quad j <= c, x \in D_A. 
\end{equation}

In practice, instead of keeping $\nabla_{\theta} f_{j}$ for all classes, the average of the labels can be kept, or only the ground truth class can be kept as well (which is what is mainly adopted in \citet{farajtabar2020orthogonal}). Moreover, since obtaining the exact  $\nabla_{\theta} f(x ; \theta)$ for any $\theta$ requires having the whole dataset $D_A$, only the gradients at the optimum $\theta_{A}^*$ are kept, assuming that the optimization of the subsequent tasks will let the parameters $\theta$ stay in the vicinity of $\theta_A^*$. Finally, not all the gradients are kept for the whole $D_A$, but rather a subset of these gradients that is chosen randomly.

As mentioned before, when training on task $\T$, the gradient $\tilde{g}$ should be perpendicular to the space of the gradients of the ground truth logits for the previous tasks $t < \T$. This space is defined as:

\begin{equation}
    S = span\{\nabla f_{k}(x, \theta^*_{t}) | (x, y) \in D_{t}, t < \T, y_k = 1\}.
\end{equation}

The orthogonal basis for $S$ are obtained using the Gram-Schmidt procedure, where they are computed iteratively as follows:
\begin{align}
    v_1 &= \nabla f_{k_{1,1}} (x_{1,1}; \theta^*_{1}) \nonumber\\
    v_i &= \nabla f_{k_{t,l}} (x_{t,l}; \theta^*_{t}) - \sum_{j < i} proj_{v_j}(\nabla f_{k_{t,l}} (x_{t,l}; \theta^*_{t})) \\
    v_n &= \nabla f_{k_{\T-1, n_{\T-1}}} (x_{\T-1, n_{\T-1}}; \theta^*_{t}) - \sum_{j < n} proj_{v_j}(f_{k_{\T-1, n_{\T-1}}} (x_{\T-1, n_{\T-1}}; \theta^*_{t})) \nonumber \,
\end{align}

Where $n_{t} = |D_t|$, $n = \sum_{t<T} n_t$. $k_{t, l}$ represents the ground truth index for sample $l$ in $D_t$ such that $y_{t, l_{k_{t, l}}} = 1$. Finally, $proj_v(u) = \frac{\langle u, v \rangle}{\langle v, v \rangle}v$. After obtaining the space $S$, the gradient $\tilde{g}$ is obtained using:
\begin{equation}
    \tilde{g} = g - \sum_{v \in S} proj_v(g) \,
\end{equation}
where it is proven in \citet{farajtabar2020orthogonal} that $\tilde{g}$ is still a descent direction for task $\T$, where there exists a non zero learning rate $\eta$, such that taking a step in the direction $\eta \tilde{g}$ will reduce the loss on task $\T$.

\begin{algorithm}[H]
    \caption{Orthogonal Gradient Descent}
    \label{algo:ogd}
    \begin{algorithmic}[1]
        \REQUIRE $\theta_{0}, \mathcal{D}_{1} \ldots \mathcal{D}_{T}$
        \ENSURE $\theta^*_{\T}$
        
        \STATE $S \leftarrow \{\}$
        \STATE $\theta \leftarrow \theta_0$
        \FOR{$t \leftarrow 1 \ldots \T$}
            \WHILE{Until Convergence}
                \STATE $X \leftarrow \text{Batch from }D_t$
                \STATE $g \leftarrow \nabla_{\theta} \text{Gradient for }X \text{ at } \theta$
                \STATE $\tilde{g} \leftarrow g - \sum_{v \in S} proj_v (g)$
                \STATE $\theta \leftarrow \theta - \eta \tilde{g}$
            \ENDWHILE
            \STATE $\theta^*_t \leftarrow \theta$
            \FOR{$(x, y)$ in $D_t$}
                \STATE $v \leftarrow \nabla f(x ; \theta^*_t) - \sum_{v \in S} proj_v (\nabla f(x ; \theta^*_t))$
                \STATE $S \leftarrow S \cup v$
            \ENDFOR
        \ENDFOR
    \end{algorithmic}
\end{algorithm}

\subsection{Task-based Accumulated Gradients (TAG)} 
TAG proposed by \citet{malviya2021tag} uses an adaptive learning rate based on the relatedness between tasks in the incremental task setup. \textit{TAG} is applied mainly to the \textit{RMSProp} optimization method, but \citet{malviya2021tag} offers other versions for how \textit{TAG} can be applied to other adaptive optimization methods as well.

\textit{RMSProp} update rule works as follows:
\begin{align}\label{eqn:rms}
    V_n = \beta V_{n-1} + (1 - \beta) g_n^2 \\
    \theta_{n+1} = \theta_{n} - \frac{\eta}{\sqrt{V_n + \epsilon}}g_n
\end{align}

Where $n$ represents the update step, $V_n$ is the moving average of the square of the gradients, $\eta$ is the original learning rate, and $\epsilon$ is a very small value to avoid degenerate cases. Although adaptive optimization methods tend to perform well in the normal supervised learning setup, they tend to perform poorly in the lifelong learning setup \citep{mirzadeh2020understanding}.

\textit{TAG} tries to adapt \textit{RMSProp} in a way that is more aligned with the lifelong learning setup. First, it captures the first moment of the gradient for each task $t$:
\begin{equation}
    M_n^{(t)} = \gamma M_{n-1}^{(t)} + (1 - \gamma) g_n \,
\end{equation}
This moment acts as a proxy for the adaptation trajectory that the model followed for task $t$, and hence the correlation $\alpha_n(\T, t)$ between the different tasks $t$ and $\T$ can be measured by using this moment as follows:
\begin{equation}
    \alpha_n(\T, t) = \exp{\left(-b \frac{M_n^{(\T)^{T}} M_N^{(t)}}{|M_n^{(\T)}||M_N^{(t)}|}\right)} \,
\end{equation}
where $M_n^{(\T)}$ is the moment for the current task $T$ after the $n$-th step, and $M_N^{(t)}$ is the final moment computed on a previous task $t$. $b$ is a hyperparameter. 

They argue that if a task $t$ is correlated with a previous task $\T$, the learning rate in the parameter update step would be higher to encourage the transfer of knowledge between task $t$ and $\T$. Whereas if the current task $t$ is uncorrelated or negatively correlated to a previous task $\tau$, the new updates over parameters may cause catastrophic forgetting and hence the learning rate should adapt to lessen the effects of the new updates. By using $\alpha_n(\T, t)$ as a proxy for the correlation between tasks, \textit{TAG} modifies Eq. \eqref{eqn:rms} to take this correlation into account for any current task $\T > 1$ as follows:
\begin{equation}
    \theta_{\T, n+1} = \theta_{\T, n} - \frac{\eta}{\sqrt{\alpha_n(\T,\T)V_n^{(\T)} + \sum_{t=1}^{\T-1} \alpha_n(\T, t) V_N^{(t)} + \epsilon}} g_n \,
\end{equation}
Using the update rule this way ensures that the second moment of the gradients from the previous tasks is taken into account when adapting the learning rate, and that the more similar tasks have a stronger effect than the less similar tasks. With the exponential term, $\alpha_n(t,\tau)$ will attain a higher value for uncorrelated tasks and will minimize the new updates (hence prevent forgetting).

\section{Summary}
To overcome catastrophic forgetting, the regularization-based methods presented in this chapter add a penalty term in the objective function,  to constrain the drastic changes in the model parameters when data from a new task arrives. Based on the motivation for defining the penalty terms and storing the past knowledge, the regularization-based methods are further categorized.  

The first type of approach involves quantifying the importance of each parameter with respect to the previous tasks and using this knowledge to control the new changes in the parameters. These approaches require storing the model parameters before it started learning the new task. We present several proposed methods that compute the importance-term in different ways. 

Another closely related methods apply concepts of Bayesian inference in lifelong learning. These methods approximate the model's parameters distribution using the optimal parameters at previous tasks.  In particular, we expand upon the two prominent approximate inference techniques employed to alleviate catastrophic forgetting: 1) Taylor expansion and 2) Variational KL minimization. 

The importance-based approach and the Bayesian-based approach rely on the prior knowledge based on model parameters. The third type of approach presented in this chapter is data-focused since it involves knowledge distillation to preserve knowledge by imposing output stability of past tasks with new data. 

The final type of approach utilizes the optimization trajectories to impose hard constraints and apply learning rates that adapt for each task in lifelong learning. These approaches essentially require storing knowledge in the form of gradients computed during the parameter updates. 

Since the regularization-based methods generally require storing model parameters from previous tasks, they are computationally expensive and often depend on the choice of prior. Moreover, when the model needs to adapt to a large number of tasks, the interference between task-based knowledge is inevitable with fixed model capacity. In the next chapter, we explore memory-based methods that do not require storing the model and comprise several state-of-the-art techniques in lifelong learning with fixed model capacity. We shall discuss methods with dynamic model capacity in the chapter after that.

\chapter{Memory-based Approaches}
\label{sec:memory}

The vanilla approach for lifelong learning is to fine-tune the model parameters on a new task ($t$) starting from the previous task parameters ($\theta_{t-1}$). Chapter \ref{sec:regularization} focuses on parameter-based regularization approaches, which prevent the model parameters from deviating too far from their initialization while optimizing the current task loss. On the other hand, in this chapter, we discuss data-based regularization approaches that hope to induce a similar behavior as parameter-based approaches. Specifically, we focus on memory-based approaches for lifelong learning, and the main feature of such approaches is an \textit{episodic memory}, $\mathcal{M}_t$. Episodic memory retains a subset of the observed examples from task $t$, and memory-based approaches use it to regularize the learning of the future tasks to alleviate forgetting of previous tasks.

Classical cognitive science studies view human memory as a single system, i.e., ``memory is memory." However, recent studies show that human memory consists of several components, each performing varied functionalities and operating under different principles. In one such study, \citet{tulving1985many} proposes a ternary classification scheme of memory constituting procedural, semantic, and episodic memories. These three memories form a hierarchical arrangement - procedural memory at the lowest level, followed by specialized semantic memory and episodic memory at the top level. Procedural memory helps deal stimulus patterns with response chains; semantic memory enables humans to construct mental models of the world (based on the capability of internally representing states of the world); episodic memory handles acquisition and retention of individual experiences and allows visiting (replaying) them again. In another work, \citet{mcclelland1995there} proposes a theory of Complementary Learning Systems (CLS) which advocates that humans rely on episodic memory to store past experiences and conduct experience rehearsal to retain previously learned knowledge. Specifically, there are two complementary systems: one that allows for the gradual accumulation of knowledge and another that allows rapid adaptation to individual experiences. 

Motivated by the above studies, a plethora of works \citep{mitchell2018never, chen2015lifelong, lopez2017gradient, chaudhry2019efficient, sprechmann2018memorybased, de2019episodic, wang2020efficientML, riemer2019learning, guo2020improved} employ memory modules for lifelong learning, particularly to alleviate the catastrophic forgetting phenomena. These works differ along multiple dimensions: 
\begin{itemize}
    \item What type of memory system is used? \textit{Episodic memory} retains a subset of the observed examples for replay \citep{lopez2017gradient, de2019episodic}, \textit{Semantic memory} retains the structured knowledge \citep{mitchell2018never, chen2015lifelong, schwarz2018progress}, \textit{Generative memory} learns the parametric model of the data and reconstructs past task examples for replay \citep{shin2017continual, sun2020lamol}.
    \item How is memory employed? \textit{Explicit} constraints on the current task gradients \citep{lopez2017gradient, chaudhry2019efficient, guo2020improved}, \textit{Implicit} constraints on the current task gradients \citep{chaudhry2019tiny, riemer2019learning, sprechmann2018memorybased}.
    \item How is memory populated? \textit{Random} examples \citep{chaudhry2019tiny}, \textit{Uncertain}  examples \citep{aljundi2019gradient}, \textit{Forgettable} examples \citep{wang2020efficientML}.
\end{itemize}

\section{A Unified View of Episodic Memory for Lifelong Learning}
\label{sec:unifiedview_er}
Memory-based approaches use episodic memory ($\mathcal{M} = \cup_{k < t} \mathcal{M}_k$) to replay the examples from previous tasks while updating the model with the current task $t$. While several methods have been developed, here we abstract away from their specific implementations and instead focus on a unified view of episodic memory-based approaches for lifelong learning. 

\paragraph{Problem Definition.} 

We consider a setup with continuum of task data:

\noindent $(x_1, t_1, y_1), \cdots, (x_i, d_i, y_i), \cdots, (x_n, t_n, y_n)$. Each triplet $(x_i, d_i, y_i)$ consists of a task descriptor $d_i \in \mathcal{T}$, input data $x_i \in \mathcal{D}_{d_i}$ and target labels $y_i \in \mathcal{Y}_{d_i}$. Here we assume that an explicit task descriptor $d_i$ is available. Further, we assume that the continuum is locally i.i.d., i.e., triplet $(x_i, d_i, y_i)$ satisfies $(x_i, y_i) \stackrel{iid}{\sim} \mathcal{P}_{d_i}(X, Y)$. Overall, the goal is to learn a predictor $f_{\theta}: \mathcal{X} \times \mathcal{T} \rightarrow \mathcal{Y}$ such as a neural network, parameterized by $\theta \in \mathbb{R}^P$, to minimize the average expected risk of all $T$ tasks:
\begin{equation}
    R(f_{\theta}) \coloneqq \frac{1}{T} \sum_{t=1}^{T} \mathbb{E}_{x, y \sim P_t} \left[ \ell(f(x, t; \theta), y) \right], 
\label{eq:e_risk}
\end{equation}
with $\ell(\cdot,\cdot)$ being the specific task loss. While the average risk is commonly evaluated after the model has seen all tasks, one can also evaluate test pairs $(x, t)$ from previously observed tasks at different stages to demonstrate the model's training behavior, and evaluate its robustness against catastrophic forgetting in terms of backward and forward transfer. While different methods have been developed to optimize \cref{eq:e_risk}, in this chapter we focus on memory based approaches for lifelong learning. The main feature of these approaches is an \textit{episodic memory}, $\mathcal{M}_t$, which retains a subset of the observed examples from each task $t$.

\begin{algorithm}[H]
    \caption{Unified View of Episodic Rehearsal for Lifelong Learning}
    \label{algo:erlll}
    \begin{algorithmic}[1]
        \REQUIRE $\theta_{0}, \mathcal{M}, \mathcal{D}_{1} \ldots \mathcal{D}_{T}$
        \ENSURE $\theta_{T}, \mathcal{M}$
        
        \STATE $\mathcal{M} \leftarrow \{\}$
        \FOR{$t \leftarrow 1 \ldots T$}
            \STATE $\theta_t \leftarrow \theta_{t-1}$
            \FOR{$k \leftarrow 1 \ldots |\mathcal{D}_{t}|$}
                \STATE $b_{t}^{k} \leftarrow \mathcal{D}_t[k]$ \COMMENT{Sample mini-batch for current task}
                \STATE $r_{t}^{k} \leftarrow \{\}$
                \IF{$\mathcal{M} \neq \{\}$}
                    \STATE $r_{t}^{k} \leftarrow \text{MEMREAD}(\mathcal{M})$ \COMMENT{Sample replay examples from the episodic memory}
                    \STATE ER: Set $\alpha_1(\theta_t^k) =1$ and $\alpha_2(\theta_t^k) = 1$
                    \STATE GEM: Set $\alpha_1(\theta_t^k)$ and $\alpha_2(\theta_t^k)$ based on \cref{eq:unified_gem_soln}
                    \STATE A-GEM: Set $\alpha_1(\theta_t^k)$ and $\alpha_2(\theta_t^k)$ based on \cref{eq:unified_agem_soln}
                    \STATE MEGA-I: Set $\alpha_1(\theta_t^k)$ and $\alpha_2(\theta_t^k)$ based on \cref{eq:unified_mega1_soln}
                \ELSE
                    \STATE Set $\alpha_1(\theta_t^k) =1$ and $\alpha_2(\theta_t^k) = 0$
                \ENDIF
                \STATE $\theta_{t}^{k+1} \leftarrow \text{UPDATE}(\theta_{t}^{k}, b_{t}^{k}, r_{t}^{k})$ \COMMENT{Based on \cref{eq:unified_er_update}}
                \STATE $\mathcal{M}_t \leftarrow \text{MEMWRITE}(\mathcal{M}, b_{t}^{k})$
            \ENDFOR
            \STATE $\mathcal{M} \leftarrow \mathcal{M} \cup \mathcal{M}_t$
        \ENDFOR
    \end{algorithmic}
\end{algorithm}

Formally, given a task $t$, $b_t$ denotes a mini-batch sampled from $\mathcal{D}_t$, \cref{eq:task_loss} defines the task loss on $b_t$ and \cref{eq:replay_loss} defines the replay loss on $\mathcal{M}_t$. 
\begin{align}
L_{TASK}(\theta; b_t) &= \frac{1}{|b_t|} \sum_{(x, t, y) \in b_t} \ell(f(x, t; \theta), y)\,, \label{eq:task_loss} \\
    L_{REP}(\theta; \mathcal{M}_t) &= \frac{1}{|\mathcal{M}_t|} \sum_{(x, t, y) \in \mathcal{M}_t} \ell(f(x, t; \theta), y)\,. \label{eq:replay_loss}\
\end{align}

Algorithm \ref{algo:erlll} outlines the overall training procedure corresponding to a unified approach. There are three main routines in Algorithm \ref{algo:erlll} - \textit{MEMREAD}, \textit{UPDATE} and \textit{MEMWRITE} and most of the memory-based approaches differ in terms of a specific implementation of these. MEMREAD routine implements a strategy to sample examples from the memory while updating the model with current task gradients. MEMWRITE routine implements a strategy to select a subset of the observed examples to write to the memory for replay. UPDATE routine implements how two objectives - task loss \cref{eq:task_loss} and replay loss \cref{eq:replay_loss} are combined to alleviate forgetting and enable backward/ forward transfer. In this section, we discuss different realizations of the UPDATE routine, and Section \ref{sec:mem_read_write} discusses the rest of the two routines in detail.

\paragraph{UPDATE.} Let $\theta_t^k$ denote the model parameters when training on $k$-th mini-batch of the task $t$. Under the unified view, the joint optimization problem covering the task loss and replay loss is defined as follows:
\begin{align}
    \min_{\theta} \alpha_1(\theta_t^k) L_{TASK}(\theta) + \alpha_2(\theta_t^k) L_{REP}(\theta) \,. \label{eq:unified_er_objective}
\end{align}
Using stochastic gradient descent method to solve \cref{eq:unified_er_objective}, one-step gradient descent update for model parameters starting with $\theta_t^k$ is defined as follows:
\begin{align}
    \theta_t^{k+1} \leftarrow \theta_t^k - \eta \big(\alpha_1(\theta_t^k) \nabla_\theta L_{TASK}(\theta) + \alpha_2(\theta_t^k) \nabla_\theta L_{REP}(\theta)\big)\,, \label{eq:unified_er_update}
\end{align}
 where $\alpha_1(\theta), \alpha_2(\theta)$ are real-valued functions controlling the relative importance of $L_{TASK}(\theta)$ and $L_{REP}(\theta)$ in each mini-batch. Now we deep-dive into existing replay-based methods and see how they all are different manifestations of the update \cref{eq:unified_er_update}.

\subsection{Experience Replay (ER)}
\label{sec:naive_er}
The most basic type of update strategy is to replay examples, $M_k$, for each task $k$ learned so far while learning the current task $t$. Under the unified view, for the update \cref{eq:unified_er_update} we have $\alpha_1(\theta_t^k) = \alpha_2(\theta_t^k) = 1$ in \cref{eq:unified_er_update}. \citet{chaudhry2019tiny} shows that as simple as this strategy seems, it performs exceedingly well compared to other more sophisticated algorithms. 

\subsection{Gradient Episodic Memory (GEM)}
\label{sec:gem}

Gradient Episodic Memory (GEM) \citep{lopez2017gradient} has a constrained optimization-based update strategy. While updating the current task loss \cref{eq:task_loss}, GEM ensures that losses on the episodic memory (of $k < t$ tasks) \cref{eq:replay_loss} does not increase in comparison to the previous task model ($\theta_{t-1}$). Formally, the constrained objective is defined as follows:
\begin{align}
    \min_{\theta} L_{TASK}(\theta; \mathcal{D}_t) \quad \text{s.t.} \quad L_{REP}(\theta; \mathcal{M}_k) \leq L_{REP}(\theta_{t-1}; \mathcal{M}_k) \quad \forall k < t. \label{eq:gem_org_obj}\
\end{align}
To inspect the episodic memory loss increase, GEM computes the angle between the loss gradient vectors of previous tasks $g_k$ and the proposed gradient update on the current task $g$. When the angle between $g$ and any of the $ g_k$'s is greater than $90^{\circ}$, $g$ is projected to the closest in $\ell2-$norm gradient $\tilde{g}$ such that it avoids the increase in losses but allows their decrease. Formally, the modified objective is defined as follows:
\begin{align}
    \min_{\tilde{g}} \frac{1}{2}||g-\tilde{g}||_2^2 \quad \text{s.t.} \quad \langle \tilde{g}, g_k \rangle \geq 0 \quad \forall k < t \,. \label{eq:gem_mod_obj}
\end{align}

GEM solves the above optimization problem \cref{eq:gem_mod_obj} via quadratic programming in the dual space with $t-1$ variables ($v \in \mathbb{R}^{(t-1) \times 1}$):
\begin{align}
\min_{v} \frac{1}{2} v^{\top}GG^{\top}v + g^{\top}G^{\top}v \quad \text{s.t.} \quad v \geq 0 \label{eq:gem_dual},
\end{align}
where $G = - (g_1, \cdots, g_{t-1}) \in \mathbb{R}^{(t-1) \times P}$, $g \in \mathbb{R}^{P\times 1}$, $P$ is the number of model parameters. Notice that $G$ is computed at each gradient step of training. Let $v^*$ denote the solution of \cref{eq:gem_dual}, then the projected gradient used for updating the model is computed as $\tilde{g} = G^{\top}v^* + g$. 
Under the unified framework \cref{eq:unified_er_objective}, GEM algorithm reduces to setting the relative importance weights $\alpha_1(\theta_t^k)$ and $\alpha_2(\theta_t^k)$ as follows:
\begin{align}
    \alpha_1(\theta_t^k) = 1, \quad \alpha_2(\theta_t^k) = v^* \label{eq:unified_gem_soln}    
\end{align}

\subsection{Averaged Gradient Episodic Memory (A-GEM)}
\label{sec:agem}

GEM constraints the current task gradient such that the episodic loss on each of the previous tasks $k < t$ \cref{eq:replay_loss} does not increase. To enforce these constraints, it requires computing the gradient using the whole replay buffer $\mathcal{M}$, as well as solving a quadratic programming problem \cref{eq:gem_mod_obj}. However, the expensive nature of these computations limits the scalability of the GEM to a large number of tasks.

Averaged GEM (A-GEM) provides a more efficient version of GEM by relaxing the constraints as it only requires that the \textit{average} episodic memory loss over the previous tasks does not increase, which reduce the constraints from $t-1$ constraints to a single constraint:
\begin{align}
    \min_{\theta} L_{TASK}(\theta; \mathcal{D}_t) \quad \text{s.t.} \quad L_{REP}(\theta; \mathcal{M}) \leq L_{REP}(\theta_{t-1}; \mathcal{M}) \quad \text{where } \mathcal{M} = \cup_{k<t}\mathcal{M}_k \,. \label{eq:agem_org_obj}\
\end{align}
The optimization problem corresponding to \cref{eq:agem_org_obj} is defined as:
\begin{align}
    \min_{\tilde{g}} \frac{1}{2}||g - \tilde{g}||_2^2 \quad \text{s.t.} \quad \tilde{g}^{\top}g_{REP} \geq 0, \label{eq:agem_mod_obj}
\end{align}
where $g_{REP}$ is a gradient computed using batch of replay examples, sampled randomly over all previously seen tasks from the episodic memory. \cref{eq:agem_mod_obj} can be solved using just an inner product between the gradients of $L_{TASK}$ ($g$) and $ L_{REP}$ ($g_{REP}$) instead of a quadratic program. When the current task gradient $g$ violates the constraint, the project gradient $\tilde{g}$ is computed as:
\begin{align}
    \tilde{g} = g - \frac{g^{\top}g_{REP}}{g^{\top}_{REP}g_{REP}}g_{REP}\,. \label{eq:agem_soln}
\end{align}
For the composite objective \cref{eq:unified_er_objective}, A-GEM algorithm reduces to setting the importance weights $\alpha_1(\theta_k^t)$ and $\alpha_2(\theta_k^t)$ as follows:
\begin{align}
    \alpha_1(\theta_k^t) = 1,  \quad \alpha_2(\theta_k^t) = \mathbf{I}_{\langle g, g_{REP} \rangle \leq 0} \bigg( -\frac{g^{\top} g_{REP}}{g_{REP}^{\top} g_{REP}} \bigg), \label{eq:unified_agem_soln}
\end{align}
where $\mathbf{I}_A$ is the indicator function which evaluates to $1$ if A holds and otherwise to 0.
\subsection{Mixed Stochastic Gradient (MEGA)}
\label{sec:mega}
Under the unified framework, one can see that GEM and A-GEM put the same weight on the current task loss regardless of how the loss changes over time ($\alpha_1(\theta_k^t) = 1$ in \cref{eq:unified_gem_soln}, \cref{eq:unified_agem_soln}). \citet{guo2020improved} argues that such a strategy does not capture a good balance between current task loss \cref{eq:task_loss} and replay loss \cref{eq:replay_loss}. For example, if the current task loss is small ($ L_{TASK} < \epsilon$), then the model performs well on the current task, and the model should focus on previous tasks in the episodic memory. On the other hand, if the current task loss is larger ($ L_{TASK} > \epsilon$), then the algorithm should weigh the current task loss relatively higher compared to the replay loss. Based upon this intuition, \citet{guo2020improved} proposes Mixed Stochastic Gradient (MEGA-I), which adaptively balances two losses by leveraging the loss information available during training. For the unified update \cref{eq:unified_er_update}, MEGA-I sets
\begin{align}
    &\alpha_1(\theta_k^t) = 1, \quad \alpha_2(\theta_k^t) = \frac{ L_{REP}(\theta_k^t)}{ L_{TASK}(\theta_k^t)} &\text{  if  }  L_{TASK}(\theta_k^t) > \epsilon \nonumber \\
    &\alpha_1(\theta_k^t) = 0, \quad \alpha_2(\theta_k^t) = 1 &\text{  if  }  L_{TASK}(\theta_k^t) \leq \epsilon \label{eq:unified_mega1_soln}
\end{align} 
where $\epsilon$ is a pre-defined threshold parameter.

\subsection{Meta-Experience Replay (MER)}
\citet{riemer2019learning} proposed learning to learn technique using gradient alignment (similar to GEM \citep{lopez2017gradient}) to reduce backward interference with a possibility of future transfer. MER also maintains an experience replay style memory with reservoir sampling. They optimize for the following objective to maximize transfer and minimize interference in lifelong learning:
\begin{align}
    \min_{\theta} \mathbb{E}_{B_t \sim \mathcal{M}} \left[\sum_{t=1}^T \left[ L(\theta;B_t) - \sum_{p,q\leq t} \alpha \frac{\partial  L(\theta;B_p)}{\partial \theta} \frac{\partial  L(\theta;B_q)}{\partial \theta}\right]\right]
\end{align}

where $B_t$ is the batch of data points belonging to a task $t$, $ L$ is the loss function, $\mathcal{M}$ is the memory and $T$ is the total number of tasks. In other words, the authors integrate the Reptile algorithm \citep{nichol2018first} (that was defined in meta-learning context) with an experience replay module to help in lifelong learning by discovering notions of tasks without supervision.

Few recent papers that use experience replay in the lifelong learning domain include Look-ahead MAML (La-MAML) \citep{gupta2020maml}, another meta-learning method that modulates per-parameter learning rates to pace the learning of a model across tasks and time. Batch-level Experience Replay \citep{mai2020batch} modifies ER mainly by performing a review step before the final testing to remind the model of the knowledge it has learned during the whole training. Apart from that, \citet{buzzega2021rethinking} modify ER by applying several tricks like performing augmentation, adding a bias correction layer in the model, decaying the learning rate exponentially, and sampling examples greedily using the training loss value.

\section{Test-time use of Episodic Memory}

Section \ref{sec:unifiedview_er} focuses on approaches that use episodic memory during training. However, lifelong learning is a continuous process, and there might not be a clear delimitation between training and evaluation. So one can assume access to the episodic memory even during evaluation. Following this assumption, several works propose to use episodic memory during evaluation \citep{Rebuffi_2017, sprechmann2018memorybased, de2019episodic, wang2020efficientML} and we will review canonical methods in detail here.

\subsection{Incremental Classifier and Representation Learning (iCaRL)}
\label{sec:icarl}

iCaRL \citep{Rebuffi_2017} is among the first methods to use episodic memory during test-time. iCaRL introduces three components to alleviate the catastrophic forgetting in the class incremental learning setup: (i) representation learning using knowledge distillation \citep{hinton2015distilling} and experience replay (section \ref{sec:naive_er}), (ii) herding based example selection for MEMWRITE, and (iii) test-time classification using nearest mean-of-features from episodic memory. 

\paragraph{Training.} The idea of using knowledge distillation is similar to the Learning without Forgetting (LwF) approach we discussed in Section \ref{sec:lwf}. Basically, after introducing a new set of classes, the distillation loss is evaluated on the older classes to ensure that the outputs of the current model are close to the output of the previous model, and the classification loss is evaluated only on the new classes. Next, the iCaRL uses a herding-based example selection strategy to write examples to the episodic memory. Instead of choosing the examples randomly, herding chooses them such that their mean approximates the class mean in the feature space. Formal discussion about the herding approach is deferred to Section \ref{sec:mem_read_write}.

\paragraph{Inference.} Let us denote $f_{emb}$ to be a feature extractor and $g_w$ to be an output layer (e.g., linear layer followed by a Softmax). Under the parametric model, the prediction rule is given as $y^* = \argmax_{y \in 1 \cdots C} g_w(f_{emb}(x)) = \argmax_{y \in 1 \cdots C} w^{\top}_yf_{emb}(x)$. This prediction rule can be viewed as a linear classifier (with weights $w_1, \cdots, w_C$) on top of non-linear features (from $f_{emb}$). Now in the context of class-incremental learning setup, our weights $w$ are decoupled from the underlying feature extractor $f_{emb}$. As a result of this, whenever $f_{emb}$ changes, then corresponding predictions can go unchecked, leading to a severe performance drop (interpreted as forgetting). To overcome this issue, iCaRL suggests using episodic memory for classification. Specifically, they use the nearest mean-of-exemplars classification strategy to predict a label $y^*$ for an example $x$. First, a prototype vector for each class is computed, and then it is used to predict the class label with the most similar prototype with $f_{emb}(x)$. The prototype for class $y$ is defined to be the average feature vector of all examples with class label $y$ in the episodic memory. Formally, the prediction rule is as follows:
\begin{align}
y^* = \argmin_{y=1,\cdots,C} \bigg|\bigg|f_{emb}(x) - \frac{1}{|M^{(y)}|}\sum_{x' \in M^{(y)}} f_{emb}(x')\bigg|\bigg|, \label{eq:nearest_mof}
\end{align}
where $M^{(y)}$ denotes all examples with class label $y$ in the episodic memory. As the $f_{emb}$ changes, the class prototypes update accordingly, therefore, the prediction rule \cref{eq:nearest_mof} does not suffer from decoupled weights issue.

\subsection{Memory-based Parameter Adaptation (MbPA)}

Inspired from CLS theory, \citet{sprechmann2018memorybased} introduce a method that consists of two components: a parametric component (neural network) that learns slowly and a non-parametric component (episodic memory with instances from previous tasks) that rapidly adapts to the parametric component. Particularly, episodic memory is used for instance-based (local) adaptation of the parametric network at inference time. Hence, \citet{sprechmann2018memorybased} term their approach as Memory-based Parameter Adaptation (MbPA).  iCaRL proposes a nearest-mean-of-exemplars classifier to overcome issues relating to decoupled weights. MbPA can be viewed as an alternate way of updating the classifier at the test time to address the same issue.

The parametric component consists of an embedding network $f_{emb}$ and a task network $g_{w}$. The embedding network is used to encode the instances and the task network is used to predict the output class distribution, $p(y|x, emb, w) = g_w(f_{emb}(x))$. Unlike the previous memory-based approaches, \citet{sprechmann2018memorybased} stores instances in the form of key and value pairs, i.e., $M_t = \{(h_i, v_i)\}$, where $h_i = f_{emb}(x_i)$ and $v_i = y_i$. During training, the usual maximum likelihood estimation is used to estimate the parameters $\{w, emb\}$. Apart from populating the episodic memory with the observed examples, it is not used during training.

\paragraph{Inference.} Similar to the iCaRL, MbPA uses episodic memory for classification. Particularly, the encoding of the current input $f_{emb}(x)$ is used to retrieve $k$ nearest neighbors from the episodic memory $C = {(h_i, v_i, w_i)}_{i=1}^{k}$. The weight $w_i$ measures the closeness of the example to the $f_{emb}(x)$ and is defined using the kernel function:
\begin{align}
    \text{kern}(h_i, f_{emb}(x)) = \frac{1}{\epsilon + ||h_i-f_{emb}(x)||_2^2}\,. \label{eq:kernelfn}  
\end{align}

The local adaptation component corresponds to adapting the output parameters $w$ to minimize the weighted average negative likelihood over the retrieved $k$ neighbors. Formally, the update rule is defined as:
\begin{align}
    w^x = \argmin_w - \sum_{i=1}^{k}w_i \log p(y_i|x_i;w)\,.
\end{align}

Notice that the episodic memory contains keys from the embedding network at different points during the training. \citet{de2019episodic} argues that this results in the embedding network from drifting over time, and the key of the test examples is closer to that of the recently seen examples. To circumvent this issue, they suggest freezing the embedding network. Given the recent surge of generic pre-trained models, \citet{de2019episodic} initializes their embedding network with pre-trained transformer-based BERT model \citep{devlin2018bert} for lifelong language learning. 

As MbPA based approaches locally adapt the model at test-time, \citet{wang2020efficientML} argue that this results in train and test-time discrepancy as the model is never locally updated during train time. This discrepancy results in negative transfer when locally updated models are evaluated on test examples from the last task. To address this problem, \citet{wang2020efficientML} proposes an efficient meta-lifelong learning framework, Meta-MbPA, by recasting the local adaptation problem as learning to ``quickly" remember using the episodic memory. 

\section{Memory Read \& Write Sampling Strategies}
\label{sec:mem_read_write}

In this section, we discuss several strategies for selecting which examples to store in the episodic memory. Some of these strategies are inspired by recent neuroscience research, while others are based on statistical insights. \citet{gupta2010hippocampal} suggest that humans replay infrequent events more often than frequent ones. Particularly, infrequent events deemed to be surprising \citep{cheng2008new, mcnamara2014dopaminergic} or rewarding \citep{atherton2015memory, olafsdottir2015hippocampal}. On the other hand, statistical strategies promote matching the data distribution of all tasks with that of the episodic memory \citep{bickel2008transfer, Rebuffi_2017}. Some suggest \citep{aljundi2019gradient, wang2020efficientML} to maximize the coverage of the episodic memory by selecting diverse examples. Alternatively, \citet{wang2020efficientML} provide active learning inspired view of sample selection strategies: a diversity-based method that picks the most representative examples and an uncertainty-based method that picks the most unsure examples (surprise \citep{ramalho2018adaptive}, forgettable \citep{toneva2018an}). We will discuss a subset of these strategies in this section.

\paragraph{Herding.} \citet{Rebuffi_2017} propose to select $m$ examples per class by iteratively selecting the examples that best approximate the average feature vector over all $n$ training examples. This iterative example selection is called herding \citep{welling2009herding} and works in an offline manner, i.e., after training the model for the given class c. Due to the offline iterative selection strategy, resulting examples constitute a representative set of samples from a distribution. Thus, this strategy strives to match the distribution of examples with that of the episodic memory at the individual class level. Let $f_{emb}: \mathcal{X} \rightarrow R^d$ be a feature extractor, $D_c = \{x_1, \cdots, x_n\}$ be the $n$ examples corresponding to $c^{th}$ class. 
\begin{align}
\mu_c &= \frac{1}{|D_c|} \sum_{x_i \in D_c} f_{emb}(x_i), \\
M_j^c &= \argmin_{x_i \in D_c} \bigg|\bigg|\mu_c - \frac{1}{k}[f_{emb}(x_i) + \sum_{j=1}^{k-1}f_{emb}(x_j)]\bigg|\bigg|\,.
\end{align}

\paragraph{Surprise.} \citet{cheng2008new, mcnamara2014dopaminergic} discuss that replay in rodents is connected to unexpected events and based upon this inspiration, \citet{isele2018selective} propose a surprise criterion for sampling transitions in incremental reinforcement learning. On the other hand, \citet{ramalho2018adaptive} propose an algorithm to approximate the task distribution based upon the surprising examples encountered during training. Formally, surprise for an example is computed using the model’s prediction as $S = - \log(y_t)$. Intuitively, the higher the probability that the model assigns to the true label $y_t$, the less surprising that example is. Further, \citet{wang2020efficientML} investigate this approach in the context of lifelong language learning by viewing it as one of the uncertainty-based sample selection strategies.

\paragraph{Reward.} Similar to surprise, some other neuroscience studies \citep{atherton2015memory, olafsdottir2015hippocampal} suggest that rewarding events are often associated with the replay. Therefore, in the context of incremental reinforcement learning,
 \citet{isele2018selective} propose a reward-based sample selection strategy. Concretely, the absolute value of the future discounted return is used to select the rewarding experiences, $\mathcal{R}(e_i) = |R_i(e_i)|$.

\paragraph{Diversity / Coverage maximization.} \citet{isele2018selective, wang2020efficientML} argue that when the memory buffer is limited in size, it is helpful to sample diverse examples to maximize coverage of the underlying data distribution. \citet{isele2018selective} propose to sample by ranking experiences based upon the number of neighbors in the episodic memory. The experience with the most number of neighbors is selected for replacements. Similarly, \citet{wang2020efficientML} leverage a pre-trained feature extractor for estimating the diversity of the sampled examples. Intuitively, for a given example, if there are nearest neighbors in the episodic buffer, that particular example is less diverse and sampled rarely (low probability). Concretely, given a feature extractor $f_{emb}$, episodic memory module $\mathcal{M}$, the probability for selecting example $x'$ is defined as follows:
\begin{align}
    \log(p(x')) \propto \displaystyle \min_{x \in \mathcal{M}} \|g_{emb}(x')-f_{emb}(x)\|^2_2\,. \label{eq:diversity_feat}
\end{align}

\paragraph{Reservoir sampling.} \citet{isele2018selective} argue that the best strategy for sampling is the one that matches the distribution of the episodic buffer with that of the global train/test distribution over all tasks. In lifelong learning, we see online streams of data, and the global distribution is not known in advance. Therefore, most of the recent works resort to reservoir sampling \citep{vitter1985random}. Given an input stream with unknown length, $m$ to be the maximum capacity of the buffer, and $n$ to be the number of examples observed so far, reservoir sampling picks examples with the probability $m/n$.  

\citet{wang2020efficientML} compares a representative set of the above-discussed strategies and finds that diversity-based sample selection strategies outperform uncertainty-based selection strategies. Notice that reservoir sampling can be viewed as a diversity-based method since it picks examples representing the true data distribution.

\paragraph{Gradient based sample selection.}
All of the above sample selection methods make an implicit or explicit assumption about the availability of task boundary. However, we might not have access to the information in some scenarios when a particular task changes. Motivated by this scenario, \citet{aljundi2019gradient} develop a gradient-based sample selection strategy to populate the replay buffer without any knowledge about the underlying task identity. Specifically, sample selection is formulated as a constraint reduction problem based on a constrained optimization view of the continual learning (see Section \ref{sec:gem} for original formulation, \cref{eq:gem_org_obj} and \cref{eq:gem_mod_obj}). From the original constraints in the gradient space ( \cref{eq:gem_mod_obj}), \citet{aljundi2019gradient} propose selecting examples so that the feasible region formed by the constraints corresponding to the selected subset of examples is close to that of the original region. Given previous $t$ tasks ($[0, \cdots, t-1]$), the original feasible region ($C$) and the reduced feasible region ($\tilde{C}$) corresponding to the memory $\mathcal{M}$ are defined as follows:
\begin{align}
    C &= \bigcap_{i\in[0,\cdots,t-1]} \{g|\langle g, g_i \rangle \geq 0\} \label{eq:feasible_reg_org} \\
    \tilde{C} &= \bigcap_{g_i \in \mathcal{M}} \{g|\langle g, g_i \rangle \geq 0\}\,. \label{eq:feasible_reg_red}
\end{align}
As $\mathcal{M}$ is a subset of the previous tasks examples, the reduced feasible region $\tilde{C}$ is infact larger than the original region $C$ (the number of examples corresponds to the number of constraints defining the feasible region). Therefore, finding the smallest $\tilde{C}$ suffices the criterion that $\tilde{C}$ is close to $C$. To define the notion of closeness, the size of the feasible region (convex cone) is defined in terms of the solid angle between the cone and the unit sphere.  Moreover, the number of constraints (gradients) is smaller than the dimension of the gradient. Therefore, the feasible region and the solid angle can be defined in $M$-dimensional space $span(\mathcal{M})$. Thus, the sample selection objective is defined as follows:

\begin{align}
    & \min_{\mathcal{M}} \lambda_{M-1} \bigg(S_{M-1}^{span(\mathcal{M})} \cap \bigcap_{g_i \in \mathcal{M}} \{g|\langle g, g_i \rangle \geq 0 \} \bigg)\,, \label{eq:sample_sel_obj}
\end{align}
where $M=|\mathcal{M}|$, $S_{M-1}^{span(\mathcal{M})}$ denotes a unit sphere in $M-1$ dimensional space, and $\lambda_{M-1}$ is Lebesgue measure. As the above optimization problem is hard to minimize, \citet{aljundi2019gradient} propose a surrogate to \cref{eq:sample_sel_obj}. Based on the observation that one can reduce the feasible region by increasing the angle between each pair of gradients, the surrogate objective for sample selection is defined as follows:
\begin{align}
    \min_{\mathcal{M}} \sum_{i,j \in \mathcal{M}} \frac{\langle g_i, g_j \rangle}{||g_i||.||g_j||} \text{ s.t. } \mathcal{M} \subset [\mathcal{D}_1, \cdots, \mathcal{D}_{t-1}]\,. \label{eq:sample_sel_mod_bj}
\end{align}

\paragraph{Gradient-based diversity maximization.} Interestingly, \citet{aljundi2019gradient} show that minimizing the above objective corresponds to maximizing the variance of the gradient direction, $Var[\hat{g}]$, where $\hat{g}$ is a unit vector.
 
\begin{align}
    Var_{\mathcal{M}} \left[\hat{g}\right] &= \frac{1}{M} \sum_{k \in \mathcal{M}} ||\hat{g}_k||_{2}^2 - \left|\left|\frac{1}{M}\sum_{k \in \mathcal{M}} \hat{g}_k\right|\right|_{2}^2 = 1 - \frac{1}{M^2} \sum_{i,j \in \mathcal{M}} \frac{\langle g_i, g_j\rangle}{||g_i||.||g_j||}\,. \label{eq:diversity}
\end{align}

Previously, we looked at a diversity-based sample selection strategy where diversity is defined in terms of the hidden representations as features. Gradient-based sample selection surrogates can be interpreted as selecting diverse examples based on gradients as features.

\paragraph{Forgettable.} Instead of analyzing the model performance under distributional shift, \citet{toneva2018an} investigate the learning dynamics of neural networks on a single task. Specifically, \citet{toneva2018an} define the occurrence of a forgetting event when the model transitions from correct to incorrect classification for individual training examples. They report that different examples are forgotten at different frequencies, and removing a significant fraction of least forgettable examples from training data still results in competitive performance. On the other hand, forgettable examples have uncommon features and are difficult to classify. Inspired by these findings, \citet{wang2020efficientML} studies the effectiveness of forgettable examples for replay by considering it as one of the uncertainty-based sample selection strategies.

\paragraph{Hindsight anchor learning.} 

The forgettable examples are sampled from the actual observations in the above method. On the other hand, \citet{chaudhry2021using} propose to explicitly construct pseudo examples/anchors such that the anchors undergo maximum forgetting after training on future tasks. Formally, given current task $t$, the desirable anchor $a_t$ with label $y_t$ can be obtained by maximizing the following loss:
\begin{align}
    (a_t, y_t) \leftarrow \argmax_{(x,y) \sim P_t} \ell(f_{\theta_T}(x), y_t) - \ell(f_{\theta_t}(x), y_t)\,,
\end{align}
where $\theta_T$ is the model after training on the future task $T (>t)$. However, for the above optimization problem, one requires access to the entire distribution $P_t$ and future tasks. To avoid storing the entire dataset for estimating $P_t$, \citet{chaudhry2021using} maintain the running average of the mean feature embedding $f_{emb}^t$ as follows:
\begin{align}
f_{emb}^t \leftarrow \beta f_{emb}^t + (1-\beta) \frac{1}{|b_t|} \sum_{x \in b_t} f_{emb}(x)\,.
\end{align}
As we do not have access to future tasks, \citet{chaudhry2021using} suggest approximating the future by simulating the past, i.e., evaluate forgetting of the current task after fine-tuning on the past tasks. Hence, the modified objective to learn maximal forgettable anchor is defined as:
\begin{align}
    (a_t, y_t) \leftarrow \argmax_{a_t \in \mathbb{R}^D} \ell(f_{\theta_{\mathcal{M}}}(x), y_t) - \ell(f_{\theta_t}(x), y_t) - \gamma (f_{emb}(a_t) - f_{emb}^t)^2\,. \label{eq:hal_objective}
\end{align}
As the above method evaluates forgetting in hindsight, the method is called hindsight anchor learning.

\section{Generative Replay}
\label{sec:generative_replay}
As mentioned earlier, the CLS theory \citep{mcclelland1995there, o2002hippocampal} proposes that human memory consists of dual complementary systems: one for gradual accumulation of the structured knowledge (neocortex) and another one for rapid encoding of the current inputs (hippocampus). Moreover, the hippocampal system reactivates the memory trace during sleep \citep{stickgold2007sleep} for the long-term memory consolidation in the neocortex with the help of multiple replays of the encoded experiences. In line with this mechanism, the memory-based approaches retain examples from past tasks for replaying them to alleviate forgetting. Further, there are pieces of evidence \citep{stickgold2007sleep, ramirez2013creating} that the hippocampal system also generates false memory experiences while replaying, thus, performing more than a naive replay. Based upon these studies, \citet{shin2017continual} argue that generative models are better conceptualizations of the hippocampal system than the replay buffer.
Further, one of the issues with simply replaying of examples from past tasks is that it requires a large memory, which is often unrealistic in real-world applications where access to the past tasks' data is limited (privacy concerns). By considering the generative models of the data, one can generate pseudo-data for experience replay, thus, relaxing the need to retain the actual examples. In this primer, we discuss two canonical works along this line, (1) \citet{shin2017continual} propose a deep generative replay framework with a generative adversarial network (GAN) to mimic the past data and studies the problem on image classification tasks, and (2) \citet{sun2020lamol} introduce a language model that simultaneously learns to solve the task and generate pseudo-samples of previous NLP tasks.

\paragraph{GAN framework.} Generative models learn to generate realistic samples by maximizing the likelihood of generated samples being in a given data distribution. GAN is one such kind of generative model that defines a zero-sum game between a generator network (G) and a discriminator network (D). The discriminator learns to distinguish between the real and the generated samples, while the generator learns to mimic the given data distribution so that it can fool the discriminator. Formally, given the real data distribution $p_{data}$, the overall objective for both the networks is defined as follows:
\begin{align}
    \min_{G} \max_{D} V(D, G) = \mathbf{E}_{x \sim p_{data}(\mathbf{x})} [ \log D(x)] + \mathbf{E}_{z \sim p_z(\mathbf{z})} [\log (1 - D(G(z)))]\,. \label{eq:minimax}
\end{align} 

\subsection{Continual Learning with Deep Generative Replay}
Based upon the above GAN framework, \citet{shin2017continual} propose a scholar model H consisting of a generator G and a solver S. Given a new task $t$, a scholar model $H_t$ is trained in two stages using the task $t$'s data and a previous scholar model $H_{t-1}$. During the first stage, a generator $G_t$ learns to reconstruct the current data ($x \sim p_{t}$) and the past data (pseudo-samples from $G_{t-1}$). In the next stage, a solver $S_t$ learns to solve the given task $t$ while remembering the previous tasks (pseudo-labels $\hat{y} = S_{t-1}(x)$). The overall objective for the scholar model $H_t$ is as follows:

\begin{align}
     L (H_t) = r \mathbf{E}_{(x,y) \sim p_t} [ L(S_t(x), y)] + (1-r) \mathbf{E}_{x \sim G_{t-1}}[ L(S_t(x), S_{t-1}(x))]\,, \label{eq:scholar_obj}
\end{align}
where $r$ is the mixing coefficient for the two objectives.

Another approach \citep{van2020brain}, motivated by anatomy, modifies standard generative replay by merging the generator into the main model. This allows replaying the hidden representations that are generated by the model's context-modulated feedback connections.

\subsection{LAnguage MOdeling for Lifelong Language Learning (LAMOL)}

\citet{mccann2018natural} show that multiple NLP tasks can be cast to a unified question-answering task, thereby enabling the use of a single language model (LM) to solve multiple tasks, i.e., given the context and question, the language model generates an answer. Based upon this observation, \citet{sun2020lamol} explore language modeling for lifelong language learning (LAMOL). Fundamentally, LM is inherently a text generator and can learn to generate samples from previous tasks. Taking inspiration from the deep generative replay, \citet{sun2020lamol} propose to continuously train a pre-trained language model that simultaneously answers the questions and generates pseudo-samples of the previous tasks. 

Although the generative replay-based approaches learn a single model without retaining old task examples, their performance is strictly upper bounded by the replay-based approaches that retain at least a few examples in the buffer \citep{sun2020lamol}. There are open questions around the scalability of the generator with the number of tasks and potential conflicts between the generator and downstream tasks due to fixed shared model capacity.

\section{Summary}

In this chapter, we presented the unified view of memory-based methods and algorithms in lifelong learning. These methods maintain an episodic memory, containing a few examples from past tasks, and revisit it while learning a new task. We saw in regularization-based methods that different ways to penalize drastic changes in the model parameters were employed and incorporated into the overall objective. Along similar lines, memory-based methods are realizations of three primary strategies combined: 1) how to sample examples from memory, 2) how to update the model with current task loss along with the replay memory loss, and 3)  how to select examples to write to the memory.  

Most memory-based methods involve defining specific model update strategies using episodic memory for training the model on each new task. In addition, we also presented approaches that use episodic memory during test time for evaluation to prevent catastrophic forgetting. 

Next, we presented numerous read/write sampling strategies employed by lifelong learning researchers. Several statistical sample selection strategies are inspired by areas like CLS theory, reinforcement learning, neuroscience and can be classified as diversity-based and uncertainty-based. 

While memory-based methods retain examples from past tasks for replaying, generative replay methods avoid storing the examples. Instead, these methods take inspiration from neuroscience and the anatomy of the human brain and generate pseudo-data for experience replay. These methods learn a single model to generate replay data that mimic the actual examples. 

So far, we have presented lifelong learning methods that assumed a fixed capacity of the ML model. In the next chapter, we will discuss methods based on isolating task-specific parts of the model and even modifying its architecture to avoid interference for training on diverse tasks.

\chapter{Architecture-based Approaches}
\label{sec:architecture}
In this chapter, we will study the different architecture families (and their instantiations) that have been proposed for training lifelong learning systems. Here, \textit{architecture} refers to the overall structure of a training system, and \textit{architecture family} refers to a family of related architectures. For example, ResNet18, ResNet32, ResNet50, ResNet152~\citep{he2016deep}, and WideResNet32~\citep{xie2017aggregated} are architectures that belong to the family of residual networks. We organize this chapter in terms of the different architectural families.

Some of these architectures are general-purpose and can be used with different settings (for example,  class-incremental or task-incremental), modalities (for example, computer vision or natural language), tasks (for example, classification or regression), and datasets. Other architectures are more specialized and useful in specific setups and scenarios. Essentially, the less general approaches make additional assumptions about the setup. If the assumptions hold for the given setup, these approaches are expected to perform better than the general architectures. The choice between general-purpose and specialized architectures is often driven by how much information we have about the setup; the more information we have, the more specialized architecture can be used. We will start the discussion with the general-purpose architectures and introduce the specialized architectures along the way.

Architecture-based approaches can be seen as a mechanism to provide useful inductive biases to the learning system. For example, when the system is trained on a sequence of closely related tasks, it may be helpful to infer the changes across the tasks as new tasks are encountered. For example, the first task could be to ``pick up a ball" and the second task could be to ``pick up a cube". Knowing ``what changes across tasks" enables the use of knowledge (from previous tasks) to train the system for the incoming task. The underlying idea is that since we have some additional information about the problem setup (for example, the tasks are closely related), we can design an architecture that can leverage the extra information. We note that while we are focusing on architectures in this chapter, in practice, these architectures are often used in conjunction with other approaches such as regularization-based methods like elastic weight consolidation (as discussed in~\cref{sec:regularization}) or memory-based methods like experience replay (as discussed in~\cref{sec:memory}). Despite the complementary nature of architecture-based approaches, we study them in a separate chapter to understand the common principles and motivations behind these architectures while abstracting away some details like the different replay mechanisms available. The rest of the chapter is organized as follows: We start with Modular Networks, the motivation behind their use, general architecture design, some common manifestations, and limitations. Next, we discuss the parameter isolation systems, which includes both fixed-capacity approaches like masking and pruning and dynamics-capacity approaches \citep{yoon2018lifelong}. We conclude with a discussion on some recently proposed approaches for lifelong learning on graphs. We include these approaches in this chapter since these approaches rely on the inductive biases specific to learning problems in the context of graphs.

\section{Modular Networks}
\label{subsec:modular_networks}

Much of the recent success of the machine learning models are limited to the \textit{single-task} setups where the data distribution is well-defined and known beforehand. We know that as the system is trained on new tasks, its performance on the previous tasks deteriorates (due to catastrophic forgetting). However, there are several challenges even before a new task is encountered. If the data distribution changes between training and evaluation, the learning system's performance often diminishes drastically~\citep{zhang2018natural, zhang2020learning, cobbe2020leveraging} suggesting that the system is not able to adapt to even small variations in the data distribution. This behavior is in stark contrast to how humans learn and operate. Not only are humans more \textit{robust} learners, they are much more \textit{sample efficient} and can quickly adapt to new tasks/data distributions. 

Several hypotheses have been proposed to explain the discrepancy between the learning behavior of humans vs. machine learning systems. One of the more popular hypotheses is that the world is inherently compositional, i.e., the representation of the \textit{whole} is composed of the representation of the \textit{parts} and the humans exploit this compositionality to understand and operate in the world~\citep{pmlr-v80-parascandolo18a}. This compositionality also implies that a \textit{novel} task can be broken down into \textit{parts} that we have already encountered in the previous tasks (in a different manifestation). For example, when reading a sentence, we break it down into phrases and words and derive meaning from it (even if we have never seen the same sentence before). Essentially, we exploit the compositionality of the world by learning modular, reusable, and general-purpose mechanisms \citep{goyal2021recurrent} (or skills). These mechanisms can be shared across different tasks, thus making learning in humans more efficient than learning in neural networks. Since leveraging compositionality is useful for humans, it may be a useful inductive bias for developing lifelong learning systems that operate in the real world and make decisions over extended periods. 

\begin{figure}[htp]
    \centering
    \includegraphics[width=\textwidth]{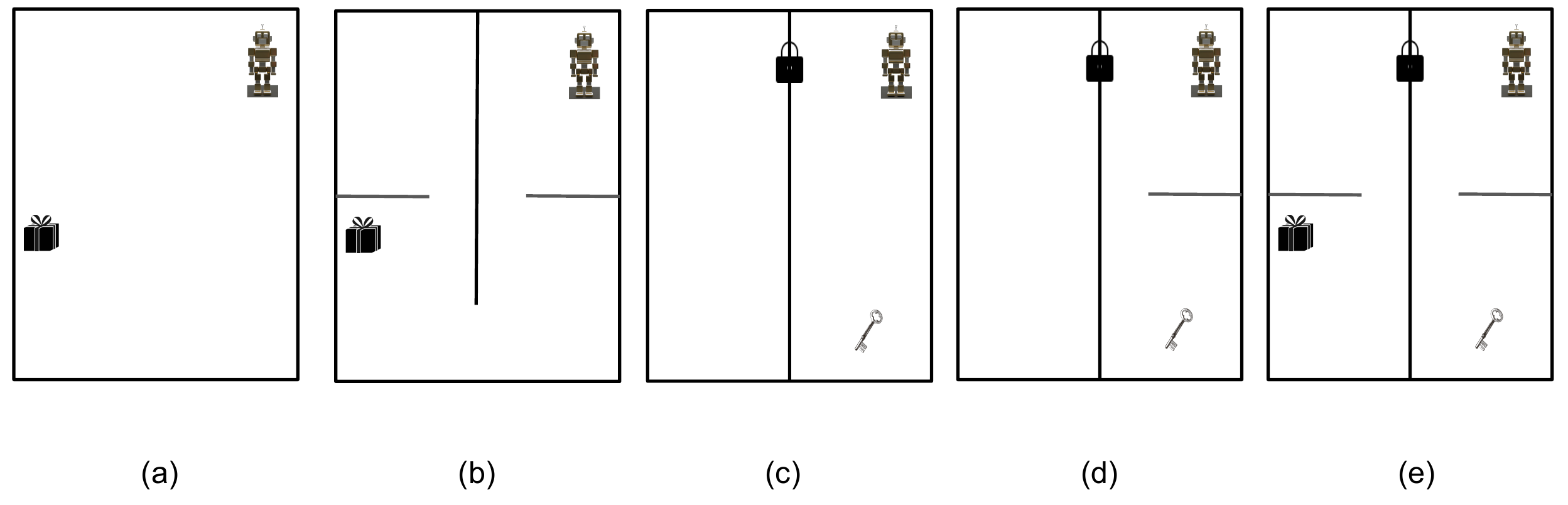}
    \caption{Example of compositionality}
    \label{fig:example_robot}
\end{figure}

Another example for such a case is given in~\cref{fig:example_robot}. To reach the goal (the gift box) in (e), the robot will have to learn to solve the first four tasks that involve walls (b), locked doors (c) and their composition (d). When the robot solves these relatively easier tasks, it can exploit the gained knowledge and reuse it to solve the final much harder task quickly. 

\textit{Modular Networks} are proposed as one of the promising direction to learn systems that can effectively leverage compositionality to learn more efficiently~\citep{happel1994design-and-evolution-of-modular-neural-network, 1996_on_combining_artificial_neural_nets, sharkey1997modularity-combining-and-artificial-neural-networks, 1998_modular_neural_network_classifiers_a_comparative_study, 1999_modularity_in_neural_computing, 1999_modular_neural_networks_a_survey, andreas2016-neural-module-networks, andreas2016learning-to-compose-neural-networks-for-question-answering, johnson2017-clevr-a-diagnostic-dataset-for-compositional-language-and-elementary-visual-reasoning, santoro2017simple-neural-network-module-for-relational-reasoning, yu2018mattnet-modula-attention-network-for-referring-expression-comprehension, alet2018modular-meta-learning-in-abstract-graph-neural-networks-for-combanitorial-reasoning, alet2018modular-meta-learning}. Modular Networks incorporate \textit{modularity} as the primary inductive bias. Modularity is the property of a system that it can be broken down into several relatively independent, replicable, and composable \textit{modules} (or smaller networks)~\citep{2019_a_review_of_modularization_techniques_in_artificial_neural_networks}. Each module can be thought of as learning to solve a \textit{subtask} (or part of a given task). In the context of compositional world hypothesis, a modular network can solve a given task by~(i) breaking it into subtasks,~(ii) using modules to solve the subtasks, and~(iii) using the solutions of the subtasks to solve the given task. Thus modular networks can also be interpreted as factorizing knowledge into different modules. Some of the underlying subtasks may change when the task/data distribution changes, even though the high-level task may remain the same. In such a case, if the knowledge is appropriately factorized,  only some modules will need to adapt/change (to account for the change in some subtasks), and the other modules can be used as-is. In practice, this would result in faster adaptation to the new distribution~\citep{bengio2019meta}. For example, consider an system that is trained to ``pick up a cup from a table". The system could learn two modules - one for \textit{reaching the table} and the other for \textit{picking the cup}. If the task changes such that the height of the table is increased, then the system only needs to update the module corresponding to \textit{picking the cup} as the subtask of reaching the table is not changed.

The benefits of modularity can be easily extended to lifelong learning. As an system trains over a distribution of tasks, it could decompose the tasks into subtasks that are shared across the task distribution. When the system encounters a new task, it could break the task into a combination of novel and previously seen tasks. In that case, the system only needs to learn the novel subtasks, instead of learning the new task from scratch. Building upon the previous example of ``pick up a cup from a table", the next task could be ``place a knife on the shelf". Both these tasks require the ability to ``move around". If the system has learned a module for ``moving around" (as part of the first task), it can use that module in the second task as well, thus enabling the positive forward transfer of knowledge. Moreover, since the module for ``moving around" may also be improved while training on the second task, it can potentially lead to a positive backward transfer of knowledge where training on the second task improves the performance on the first task.

We note that similar to modular networks, several others areas like Out-of-Distribution (OOD) generalization~\citep{wang2021generalizing}, zero-shot generalization~\citep{purushwalkam2019task}, few-shot generalization, etc., also focus on narrowing the gap in the performance of humans and machine learning systems.

\subsection{Motivation}

The use of modular networks for lifelong learning can be motivated from various other perspectives as well.

\begin{enumerate}
    \item \textbf{Cognitive Science Perspective}: ~\citet{spelke1990principles, pinker1994language, pinker2005so, spelke2007core, xu2009induction} hypothesized several theories to explain how humans learn coherent, abstract, and highly structured representations of the world from the fragmented but concrete instances of \textit{experiences}. The ``theory theory''~\citep{carey1985conceptual, gopnik1988conceptual, wellman1992cognitive} states that as humans interact in the world, they construct intuitive theories of the world. These theories have three key aspects: {(i)} They involve coherent, abstract, causal representations of the world. {(ii)} They have distinct cognitive functions. For example, theories enable both prediction of the future as well as counterfactual inferences. {(iii)} They have distinctive dynamic features. For example, the theories can be updated as humans undergo novel experiences and discover new knowledge. These different theories aim to explain the interplay between abstract knowledge and concrete knowledge. In the modular networks, the network topology can be seen as a manifestation of the abstract knowledge, and the modules can be seen as encapsulating the concrete knowledge. 
    \item \textbf{Evolutionary Perspective}: \citet{2005_spontaneous_evolution_of_modularity_and_network_motifs, 2007_varying_environments_can_speed_up_evolution} hypothesized that environments with Modularly Varying Goals (MVGs), i.e., environment consisting of varying goals with common subgoals, leads to modular networks.~\citet{2013_the_evolutionary_origins_of_modularity} hypothesized that modularity evolves as a byproduct from selection to reduce connection costs (like creating, sustaining connections) in a network. This modularity is then sustained by the Modularly Varying Goals (MVGs). A lifelong learning system is expected to learn (and retain the knowledge of) a series of tasks over its lifetime. Modularity can be a useful inductive bias for the neural network if these tasks share some common substructure.
    \item \textbf{Empirical Perspective}: From a practical perspective, a modular neural network can be interpreted as a system of modules where each module is designed to solve one specific task, and the controller learns the mapping between the tasks and the modules. If multiple tasks share a common subtask, the modules corresponding to these tasks can share knowledge with each other. When new tasks are encountered, the network will eventually run out of capacity, a problem referred to as \textit{capacity saturation}~\citep{sodhani2020toward}. Modular neural networks provide a workaround for that by enabling the addition of new modules that can be added to the system, without disrupting the existing modules (and the knowledge encoded by them). Modularity also helps with catastrophic forgetting by localizing the forgetting effect, i.e., forgetting knowledge about a task should only affect the modules related to that task and not the other modules.
\end{enumerate}

\subsection{Architecture}

The high-level architecture of modular neural network can be described in terms of the following two components:

\begin{enumerate}
    \item A system of $n$ modules denoted as $M = \{m_{i} \forall i \in \{1,\cdots n\}\}$. 
    \item A \textit{controller} mechanism that is used to decide the topology of connection between the modules.
\end{enumerate}

We note that different works use different terminology for describing the architecture of neural modular networks. For example, some works denote the modules as \textit{experts}~\citep{2017_encoder_based_lifelong_learning, 2017_expert_gate_lifelong_learning_with_a_network_of_experts} or as \textit{primitives}~\citep{MLSH, Goyal2020Reinforcement} etc. Similarly, some works refer to the \textit{controller} as the \textit{gating mechanism}\citep{2017_encoder_based_lifelong_learning, 2017_expert_gate_lifelong_learning_with_a_network_of_experts} or as \textit{router}\citep{rosenbaum2017routing}. While these different works have subtle differences in terms of how modules/controller are instantiated (or interact), the terminology we use here suffices to study them from the perspective of lifelong learning.

\paragraph{System of Modules.}

\begin{figure}[htp]

    \centering
    \includegraphics[width=0.5\textwidth]{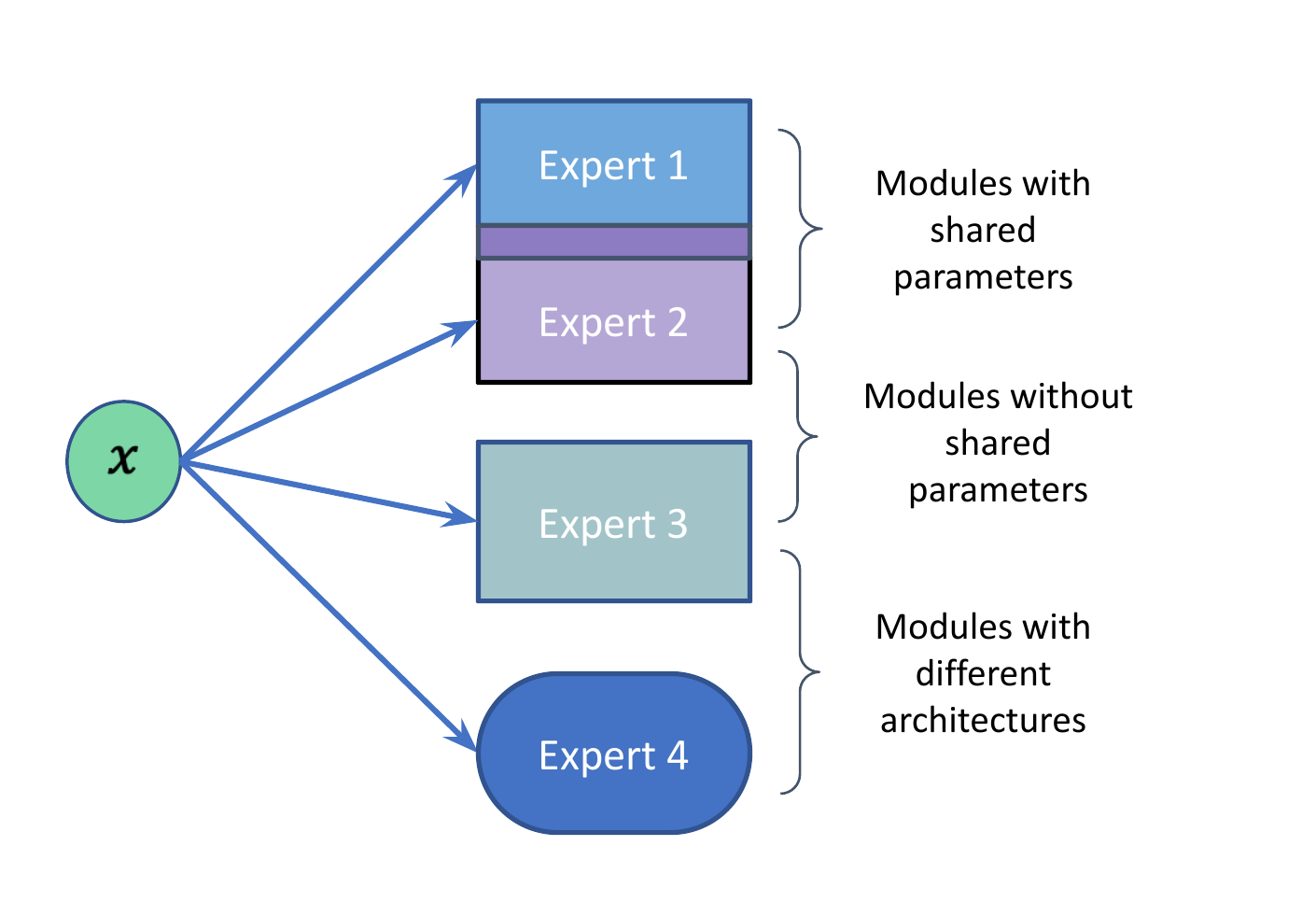}
    \caption{Different kind of modules in Modular Architectures}
    \label{fig:example_modular_net}
\end{figure}

The first key component of modular networks is the system of $n$ modules $M$. Each module $m_i (\forall i \in \{1\cdots n\})$ is a neural network with parameters denoted as $\theta_i$. In terms of the network architecture, the modules may be identical (for example, a collection of ResNet50 models) or similar (for example, a collection of models from the ResNet family) or different (for example, some modules from the ResNet family and some modules from the VGG family). In the case of modules belonging to different architectures, the modules may also operate on different modalities. In terms of functional representation, the modules may be explicitly trained to encode different aspects of the input (maybe by providing direct supervision), or the modules may learn different representations on their own. Some modules may also share parameters with other modules (we denote the shared parameters of the $i^{th}$ module as $\theta_{i}^{shared}$). An example of a system of modules is shown in~\cref{fig:example_modular_net}.

\paragraph{Controller Mechanism.}

The second key component of modular networks is the controller mechanism. A controller is any function $C$ that defines a network topology using the system of modules. The input to the controller mechanism could include: input data points, metadata in the form of the task description, output from the system of modules, or a combination of these. Note that this description of the controller includes both implicit and explicit controllers (parameterized as well as non-parameterized). The output of the controller is a network topology $z$. This topology defines the \textit{arrangement} of (or interaction between) modules, thus instantiating a function $f_z$ as shown in \cref{eq::modular}. Note that all the modules are not required to be part of the network topology. Moreover, the controller may not have to learn the entire topology as some parts of the topology may be pre-determined. The controller may generate the entire network topology at once \citep{2017_expert_gate_lifelong_learning_with_a_network_of_experts} or step by step (as done in \citet{goyal2021recurrent}).

\begin{align}
z & = C(x, metadata, M),\\
y & = f_z(y | x, M). 
 \label{eq::modular}
\end{align}

\paragraph{Training Mechanism.}
Several training mechanisms have been proposed for training Neural Modular Networks. These mechanisms vary in terms of the following dimensions:

\begin{enumerate}
    \item Should the modules and the controller be trained jointly, in an end-to-end manner (as done in for example~\citet{chang2018automatically}) or have separate losses for the two components. Should the controller be meta-trained?
    \item Is the network layout $z$ a discrete variable, like a hard-attention mask (or graph with 0-1 edges), or is the network layout represented as a continuous variable, like a soft-attention mask (or graph with soft-edges).
    \item Is the controller mechanism explicit or implicit?
    \item Is a per-module loss available?
\end{enumerate}

\paragraph{Relationship with standard Deep Learning architectures.}
The standard, monolithic, deep learning architectures (like ResNets) can be seen as a special form of modular networks. The layers (of the monolithic network) can be seen as the modules, and the controller mechanism is the pre-determined and hardcoded topology where the output of one layer feeds into the next layer. In this sense, the process of learning modules is similar to the process of training layers in a neural network. 

\subsection{Lifelong Learning}

Modular Networks represent a very general and flexible family of neural network architectures. They provide several benefits like the ability to use different model architectures as constituent modules and ability to add new modules throughout training. Modular neural networks are also biologically inspired~\citep{2015_the_modular_and_integrative_functional_architecture_of_the_human_brain} and the brain has been shown to be modular at different spatial scales, from the micro level of synapses to the macro level of brain regions~\citep{2002_revealing_modular_organization_in_the_yeast_transcriptional_network, 2006_modularity_and_community_structure_in_networks, 2008revealing_modular_architecture_of_human_brain_structural_networks_by_using_cortical_thickness_from_MRI, 2007_the_road_to_modularity, 2009complex_brain_networks_graph_theoretical_analysis_of_structural_and_functional_systems, 2010efficient_physical_embedding_of_topologically_complex_information_processing_networks_in_brains_and_computer_circuits, 2010modular_and_hierarchically_modular_organization_of_brain_networks}. In the context of deep learning, Modular Network initially focused on visual questions answering~\citep{andreas2016-neural-module-networks}. Since then, they have been extended to several settings like multitask learning~\citep{zhang2020multi, sodhani2021multi}, compositional generalization~\citep{Goyal2020Reinforcement, goyal2021recurrent, chang2018automatically}, etc. Different applications of modular networks, in context of lifelong learning, differs in terms of what the modules learn (or should learn).

\paragraph{Modules can represent composable skills.}

Several works~\citep{MLSH, chang2018automatically, feudal_networks_for_hierarchical_reinforcement_learning, Goyal2020Reinforcement} use the modules to learn composable \textit{skills}. The general idea is the following: When a model is trained on a given task, the model \textit{learns} skills that it composes to solve the given task. The benefit of learning skills is that skills can be selectively transferred across tasks. In this case, the controller mechanism should help to ensure that modules learn a diverse set of skills and the model learns to solve a task by choosing and composing a subset of modules (skills).

Using modules for learning skills has an additional benefit: A new task may require learning new skills. In that case, new modules can be instantiated and trained on the given task (and used for the subsequent tasks). Some common challenges in this setup are:~(i) learning a diverse set of skills,~{(ii)} not forgetting the previously acquired skilled (in principle, this can be easily achieved by not finetuning a module on subsequent tasks, but this may force the model to learn very similar skills),~{(iii)} controlling the growth of the number of modules (if the tasks share the skills, the number of modules should grow sub-linearly with the number of tasks).

We explain the general design and implementation of works in this category using ~\citet{veniat2021efficient} that proposed a neural modular network architecture for lifelong learning where each module $m_i$ represents a composable, atomic skill. The model solves a task by transforming the given task to a new task: searching through an exponentially large search space (of the composition of modules) and inferring a composition that enables the model to solve the given task. While modules are shared across tasks (to enable knowledge transfer), modules corresponding to the older tasks are not updated for the newer tasks, thus avoiding catastrophic forgetting. New modules are added only when new skills are required. If tasks share skills, the module grows sub-linearly with the number of tasks.

At the start of training, the model is initialized as a collection of $l$ modules, arranged in $l$ layers with one module per layer. The modules can be different across the layers. During training, new modules could be added across the layers, with the constraint that all the modules in a layer are of the same architecture (but can learn different weights). The setup can be explained with an example where the model has been trained on $t-1$ tasks. Now, when the $t^{th}$ task arrives, a new randomly initialized module is added to each layer, and a search space is defined over all the possible ways to combine the modules (both old and new). 

The resulting search space is exponentially large, and several constraints are imposed to keep the search tractable:

\begin{enumerate}
    \item Modules within the same layer do not connect with each other, so only one module can be selected from each layer.
    \item A newly added module, say at the $i^{th}$ layer, can only connect to another newly added module at the $i+1^{th}$.
\end{enumerate}

The second restriction is motivated by the need to reduce the size of the otherwise exponentially large search space. In practice, this restriction can be justified as follows: as new tasks are added, changes are expected in the output distribution and not the input distribution. If the tasks are related, initial layers can be shared across the tasks. 

The training objective is to minimize the loss corresponding to the $t^{th}$ task, as a function of the parameters (of all the modules) and connection between the modules. This connection is essentially a path in the grid of modules. The connection of the $t^{th}$ task is denoted as $\pi_{t}$ and the parameters of the modules, that are part of this path, are denoted as $\theta(\pi_{t})$. The model optimizes the loss function:
\begin{equation}
\Gamma^*, \theta^* =  \arg \min_{\theta, \Gamma} \mathbb{E}_{j \sim
  \Gamma, (x,y) \sim \mathcal{D}^t} \mathcal{L}( f(x, t| \mathcal{S}, \theta(\pi_j)), y). \label{eq:loss_general}
\end{equation}

Here $\mathcal{S}$ is the set of tasks model has seen so far, $x, y$ denote the input and the target label for the $t^{th}$ task, $f(x, t| \mathcal{S}, \theta(\pi_j))$ is an instance of modular network using the path $\pi_j$, $\mathcal{L}$ is the loss function and $\Gamma$ is a distribution over the set of possible paths. Other works have considered different approaches for selecting a path. For example~\citet{2019_an_adaptive_random_path_selection_approach_for_incremental_learning} randomly select a path connecting the modules and keeps using that path until it saturates (i.e., the parameters have been fully used for learning on the previous tasks), while ~\citet{2017_pathnet_evolution_channels_gradient_descent_in_super_neural_networks} use genetic algorithms to select a path connecting the modules.

Note that only the newly added modules can be trained, and the parameters of the previously added modules remain unchanged. At the end of training on the $t^{th}$ task, any new module, that does not appear on the optimal path, is not retained for the subsequent tasks, thus keeping the model's growth sublinear in the number of tasks. The resulting path is saved for the given task and can be retrieved during testing. 

There are two approaches for optimizing the loss in~\cref{eq:loss_general}. In the first approach, called the stochastic variant, the model alternates between optimizing the path and optimizing the parameters of a given path. In terms of general modular network architecture, this corresponds to alternatively optimizing the controller mechanism and the modules. Specifically, the distribution $\Gamma$ is modeled by a product of multinomial distributions, one for each layer of the model. The modules are selected \textit{one-layer-at-a-time}. An entropy regularizer is used to encourage the model to explore different paths. The second approach, called the deterministic variant, is more straightforward - an exhaustive search is performed over the set of paths.

~\citet{veniat2021efficient} uses a \textit{data-driven prior} which works as follows: First, $k$ tasks, which are most similar to the given task, are selected from the set of previous tasks. The search space for the current task is limited to the perturbations of the paths corresponding to these previous tasks. The most similar previous tasks are chosen by computing the predictions on the current task data, using paths from all the previous tasks (specifically, using the feature from the penultimate layer). The paths that yield the best nearest neighbor classification accuracy are selected. Several other mechanisms are also used to restrict the search space over paths further. For example~\citet{andreas2016-neural-module-networks} selects only one module at a time (instead of selecting an arbitrary number of modules at any time), while~\citet{goyal2021recurrent} attends to all the modules.

\subsection{Limitations}

There are two key limitations to existing approaches for training modular neural networks which limit their usability in lifelong learning applications:

\begin{enumerate}
    \item Selecting the right level of \textit{modularity}: One big open problem is how to enable the models to learn modularity at the right level of abstraction. If the modules learn very high-level skills, then the modules will become task-specific, while if the modules learn very low-level skills, a large number of modules need to be trained.
    \item Learning a diverse set of modules: Another big challenge is how to train the modules so that they learn \textit{diverse} skills and do not \textit{collapse} to the same skill.
\end{enumerate}

Due to these limitations, several works use hand-crafted modules, i.e., they train the modules to learn specific skills. While this approach may work reasonably well for the single-task setup, scaling it to lifelong learning (or even multi-task) setups has been difficult.

\section{Parameter Isolation Systems}
\label{sec:isolation}

\textit{Parameter Isolation} based approaches work on the idea that different tasks should have their own set of ``isolated" parameters. If no two tasks share any parameters, training on any task can not cause catastrophic forgetting on the other task. The idea of ``isolating parameters" is in direct contrast to the idea of ``learning composable modules" as the parameter isolation approaches focus more on avoiding catastrophic forgetting. In contrast, the modular network-based approaches focus more on knowledge transfer (both forward and backward). Even though these approaches are designed to avoid catastrophic forgetting by not sharing parameters, catastrophic forgetting happens because parameters are updated using loss from multiple tasks (and not because parameters are used to make predictions on several tasks). In practice, this means that parameters of different tasks can be used together in the forward pass but not in the backward pass, thus enabling forward transfer of knowledge, but not a backward transfer of knowledge.

However, in practice, many parameter isolation-based approaches can be described as modular neural networks with slightly different inductive biases. For example, a~\textit{controller-like} mechanism can be used to select which parameters belong to which task. Despite their similarity (and close relationship with) modular architectures, we consider them as a different category because: 
\begin{enumerate}
    \item Unlike the modular systems, which explicitly share modules across many tasks (thereby enabling both forward and backward knowledge transfer), parameter isolation-based approaches use different parameters for each task.
    \item Parameter isolation approaches focus on avoiding catastrophic forgetting (at the expense of backward transfer).
    \item Often, these approaches add more parameters as new tasks are encountered. Hence the number of parameters grows as $\mathcal{O}(n)$ where $n$ is the number of tasks. Thus one frequently encountered challenge is to reduce the memory footprint when adding new tasks. On the other hand, the common challenge for modular networks is to ensure that the knowledge is decomposed into different skills.
\end{enumerate}

Parameter Isolation Systems are generally studied across two dimensions: ~{(i)} fixed capacity networks and~{(ii)} increasing capacity networks.

\subsection{Fixed Capacity Networks}

The first sub-category of parameter isolation networks consists of models with a fixed capacity (i.e., the model's capacity does not change as it trains over subsequent tasks). The isolated set of parameters is instantiated by learning task-specific~\textit{masks} which are used to determine which parameters will be used in the forward/backward pass for which task. These approaches trade-off catastrophic forgetting with model capacity by controlling the gradient flow through the network. Different mask-based approaches differ in terms of~{(i)} what is masked and~{(ii)} how are the mask values computed.

\paragraph{Masking-based Approaches.}

In parameter isolation systems, it is not required that the parameters for a given task will always form a contiguous block of parameters (like a feedforward layer). The masking function can operate at the fine level of parameters, i.e., the masking function can generate a mask per parameter. The general idea of masking is based on previous works like~\citet{2015_binaryconnect_training_deep_neural_networks_with_binary_weights_during_propagations,2016_binarized_neural_networks} that train neural networks with binary-valued weights. The basic idea is to use a masking function to binarize real-valued weights during the forward pass and update the real-valued weights using the gradients corresponding to the binarized weights. Other works~\citep{2016_dynamic_network_surgery_for_efficient_DNNs} have used implicit masking functions that use the magnitude of the weight as a criterion for masking. 

~\citet{2018_piggyback_adapting_a_single_network_to_multiple_tasks_by_learning_to_mask_weights} proposed to learn bit-wise binary masks for each task. These masks are used to activate/deactivate the weights of a fixed and shared backbone network (on which the masks are applied) by element-wise multiplication with the binary masks. During training, the first step is to pre-train the backbone network.~\citet{2018_piggyback_adapting_a_single_network_to_multiple_tasks_by_learning_to_mask_weights} used the ImageNet dataset~\citep{deng2009imagenet} and reported that the pretrained backbone network works well across multiple tasks. When training on a given task, a mask network is used to generate real-valued weights. These weights are mapped to binary masks using a deterministic thresholding function. These masks are multiplied (elementwise) with the weights of the backbone network, to generate the task-specific weights. The entire setup is trained end to end based on ideas from network binarization~\citep{2015_binaryconnect_training_deep_neural_networks_with_binary_weights_during_propagations, 2016_binarized_neural_networks_training_deep_neural_networks_with_weights_and_activations_constrained_to_plus_1_or_minus_1} and pruning~\citep{2016_dynamic_network_surgery_for_efficient_DNNs}. Once the model is trained on the given task, the masking network is discarded, and only the task-specific bitwise mask is retained. The learned masks ``piggyback" on the backbone network to solve a given task. One limitation of this approach is the dependence on pretraining the backbone network as the randomly initialized backbone networks perform quite poorly in practice.

Learning binary bit-masks may appear too restrictive in practice, and some works have explored the possibility of learning real-valued masks as well, although learning real-valued masks can be difficult in practice due to the possibility of forgetting the knowledge useful for the previous tasks. As such, learning real-valued masks requires designing the system carefully. 

\citet{2018_overcoming_catastrophic_forgetting_with_hard_attention_to_the_task} extend the idea of learning task-specific binary masks over weights to task-specific \textit{almost-binary} attention masks over features (or activations). The proposed method, HAT (Hard Attention to Tasks), uses attention vectors of the previous tasks to define a mask for the current task and constrain the updates to the model's weights. The paper motivates this approach as follows: \textit{When training on a sequence of tasks, different tasks could reuse the same intermediate features, but in different ways. For example, given a dataset of birds and dogs, the first task may require the model to differentiate between birds and dogs, and the second task may require the model to differentiate between black and brown animals. In such cases, the task identifier could be a useful feature for the model to perform well on both tasks, by conditioning the network layers on the task identifier.} The per-layer attention weight is computed as follows:
$$\bm{a_l^t} = \sigma(s\bm{e_l^t})\,,$$
where $\bm{a_l^t}$ is the attention vector for the $l^{th}$ layer of the model and $\bm{e_l^t}$ is the single-layer task embedding for the $l^{th}$ layer of the model, when training on the $t^{th}$ task. $\sigma$ is a gating function which squashes any input to the range $[0, 1]$ and acts as a \textit{pseudo-step} function. A hard step function is not used to enable flow of gradients to the layers of the model. $s$ denotes a positive scaling parameter that controls the \textit{hardness} of the \textit{pseudo-step} function. In the case of the final layer, the attention vector is binary-hardcoded (since the final layer is a task-conditioned multi-headed output layer). After training the model on $t^{th}$ task, a cumulative attention mask is computed by taking an element-wise max on all the previous attention vectors. i.e., $\bm{a_l^{\leq t}} = max(\bm{a_l^t}, \bm{a_l^{\leq t-1}})$. Since max operation is used, any feature (that was important for any of the previous tasks) gets a high attention score. When computing the gradient on the ${t+1}^{th}$ task, the gradient $g_{l,ij}$, corresponding to the $i^{th}$ output and $j^{th}$ input units in the $l^{th}$ layer is masked using $[1-min(a_{l, i}^{\leq t}, a_{l-1, j}^{\leq t})]$. This mask prevents large updates to weights that are important for the previous tasks. When computing the attention weights, $s$ is annealed in an epoch as follows:
$$s=\frac{1}{s_{max}} + (s_{max} - \frac{1}{s_{max}})(\frac{b-1}{B-1})\,,$$
where $s_{max}$ is a large positive constant ($>>1$), $B$ is the total number of batches in an epoch and $b$ is the number of batches seen so far in the epoch. So at the start of the epoch, all features are approximately equally likely to be activated, and as training progresses, the selection becomes more binarized. However, the gradient annealing scheme introduces some optimization challenges. Specifically, the embeddings $\bm{e_l^t}$ do not change much during training, and gradient magnitude is weak. Hence an additional gradient compensation term is introduced to compensate the effects of the annealed sigmoid, and some intermediate values (like $|s\bm{e_l^t}|$) are clamped to obtain well-behaved gradients. For the $t^{th}$ task, activations with a hard attention value are dedicated to that task. A sparsity constraint is introduced to reserve some model capacity for subsequent tasks by constraining the capacity spent on each task. During inference, $s$ is set to $s_{max}$ so that the gating function behaves like a step function.

Generally, \textit{masking approaches} uses the same set of masks for both forward and backward transfer. While this makes sense in practice (weights that were not used during the forward pass can not have gradients during the backward pass), it also reduces the model's flexibility, in terms of transferring knowledge across the tasks. 

\citet{2020_ternary_feature_masks_continual_learning_without_any_forgetting} proposed using ternary, instead of binary, masks for each task. Similar to~\citet{2018_overcoming_catastrophic_forgetting_with_hard_attention_to_the_task}, these masks are applied to the features (or activations) of each layer and not to the weights. Since the number of activations tends to be smaller than the number of weight parameters, the memory overhead of ~\citet{2020_ternary_feature_masks_continual_learning_without_any_forgetting} is lower in practice, compensating for the need to store 2x bits per feature instead of $1$ binary bit per weight. However, unlike~\citet{2018_overcoming_catastrophic_forgetting_with_hard_attention_to_the_task}, the masks are binary (and not real-valued) as only binary masks can guarantee to avoid \textit{forgetting} of previous tasks. Two sets of masks are learned - the masks for features that will be \textit{used} (forward pass) and another set for the features that will be \textit{learned} (backward pass). This is related to how freezing layers work, features used in the forward pass do not necessarily have to be used in the backward pass. Specifically, the features can be in one of the three states: \textit{used} (forward pass only), \textit{learned} (forward and backward pass), and \textit{unused} (neither forward nor backward pass).

Generally,~\textit{masking approaches} do not allow any change to the features corresponding to the previous tasks, so~\citet{2020_ternary_feature_masks_continual_learning_without_any_forgetting} add task-specific feature normalization that makes the (previously learned) features more optimal use for the later tasks. However, feature normalization comes at the price of storing two extra floating point numbers per activation per task.

\paragraph{Pruning-Based Approaches.}

An alternate design choice would be to iteratively free-up parameters (via, say, network pruning) so that the subsequent tasks can use the parameters freed by the previous tasks, without having to add new parameters. PackNet~\citep{2018_packnet_adding_multiple_tasks_to_a_single_network_by_iterative_pruning} uses network pruning approach from previous works like~\citet{2015_learning_both_weights_and_connections_for_efficient_neural_networks, 2016_dsd_dense_sparse_dense_training_for_deep_neural_networks} and proposes to sequentially \textit{pack} multiple tasks into a single network by performing iterative pruning and network re-training. One major difference between PackNet and previous masking-based works is, all the unmasked parameters (that is, parameters that have not already been masked in the previous tasks) are used during both forward and backward passes. The mask for the $i^{th}$ task is computed after training is finished on the $i^{th}$ task. The training setup is quite straightforward. The model starts by training on the first task. After convergence, a certain percentage of weights are \textit{pruned} i.e., set to $0$. The network is retrained on the current task (without using the pruned weights) to account for the effect of pruning. When the model is trained on the second task, the unpruned weights (used in the first task) are kept fixed, and the pruned weights (not used in the second task) are trained. After convergence, some of the weights (trained in the second task) are pruned, and the model is trained on the unpruned weights (from the second task). This process is repeated every time a new task is added. In each round of pruning, the weights (in all the convolutional and fully connected layers) are sorted by their absolute magnitude, and the lowest $50\%$ (or
$75\%$) weights are pruned. The paper reports that subsequent retraining uses half as many epochs as original training.

One important thing to note is that PackNet is evaluated in setups where up to three new tasks are added. While some of the previous works have used fewer tasks~\citep{ewc, 2017_learning_without_forgetting}, scaling PackNet to a large number of tasks may require some changes. For example, it is expected that at some point, the network's capacity would have to be increased, probably using approaches like Net2Net~\citep{2015_net2net_accelerating_learning_via_knowledge_transfer, sodhani2020toward}.

\paragraph{Learning paths in the network.}

\citet{2017_pathnet_evolution_channels_gradient_descent_in_super_neural_networks} proposed PathNet that uses evolutionary algorithms to discover \textit{paths} (subparts of the network) to re-use for new tasks. A PathNet is a deep neural network having $L$ layers, with each layer consisting of $M$ modules. A module is said to be \textit{active} if it is on the currently selected path. At most, $N$ ($3/4$ in practice) modules can be active in any layer at any time. A pathway is represented by a matrix of atmost $N \times L$ integers. The integers in the $i^{th}$ column refer to the active modules of the $i^{th}$ layer. The output of each layer is summed up before passing to the subsequent layers. Pathways can evolve in two ways - serially or parallelly. In the \textbf{Serial Pathway Evolution} $P$, pathways are initialized randomly and are represented by a matrix of atmost $N \times L$ integers. A binary tournament selection algorithm is used where two pathways are selected randomly and trained for $T$ epochs. The \textit{fitness} of a pathway is measured in terms of classification error during training. The winning path is mutated by randomly selecting some modules (on the path) and swapping them with nearby modules in the layer. In the \textbf{Parallel Pathway Evolution}, multiple paths are trained in parallel, and as some paths are trained, they are compared with the other trained paths. The winning path overrides the losing paths, followed by a round of mutations. 

After the model has learned a task, the best pathway is fixed, and its parameters are no longer updated. Modules that are not part of the best path are reinitialized. As the model trains on the next task, the same procedure (sampling random paths, comparing paths, and mutating the winner) is repeated. The paper reports that the previous best paths are active (during forward pass) in the RL experiments but not in the supervised learning experiments. The previous best paths are never used in the backward pass.

One limitation of the work is that it is evaluated on a sequence of only two tasks, which could limit its usefulness in practice.

Related to previous works on selecting a task-specific \textit{path} through the network,~\citet{2019_random_path_selection_for_continual_learning} proposed RPS-Net (Random Path Selection Network) that starts with some random candidate paths and discovers the optimal path for a given task. The network consists of $L$ distinct layers, where each layer has a set of $M$ modules, stacked in parallel, along with a skip connection. During training, path selection is performed for every $J$ task. During path selection, $N$ paths are randomly chosen and followed by the training process. The best path is then used for the next $J$ tasks. Since the previously selected paths are fixed, the computation remains bounded, as at most one module for each of the $L$ layers is being trained. In practice, the work also leverages techniques from regularization (\cref{sec:regularization}) in the form of knowledge distillation and experience replay (\cref{sec:memory}), thus making it a hybrid approach. Additionally, a controller is used to balance between the current task loss and the knowledge distillation loss. The controller increases the weight of the knowledge distillation loss as the training progresses.

One important strength of RPS-Net is that, during inference, it does not need to know the task to which the given data point belongs as a common inference path is used. 

\subsection{Expert-Based Systems}

An extreme version of Parameter Isolation Systems is the idea of adding a new network/model per task (which may or may not use predictions from the previously trained models). A very popular instantiation of this approach is Progressive Neural Networks~\citep{2016_progressive_neural_networks} where a new \textit{column} (network) is added every time a new task is encountered. The newly added column has lateral connections to the previous tasks that enable the forward transfer of knowledge from the previously trained models to the newly added model. During training, only the newly added model is trained, and the old weights are kept fixed, thus protecting from catastrophic forgetting. The obvious downside of this approach is that the number of parameters increases linearly with the number of tasks. Progressive Neural Networks also notes that the newly added models are not used to their full capacity, thus leaving scope for improvement.

~\citet{schwarz2018progress} propose a related, but much more memory efficient idea where two networks are maintained. One network (referred to as the \textit{active column}) is used for training on the current task, and the second network (referred to as the \textit{knowledge base}) stores the knowledge for solving the previous tasks. Training happens in two phases. In the first phase (called \textit{progress} phase), the \textit{active column} is trained on the current task. Once the \textit{active column} converges, the second phase starts where the \textit{active column} is distilled into the \textit{knowledge base}. This phase is referred to as the \textit{compress} phase. Thus training over multiple tasks proceeds as a sequence of \textit{progress} and \textit{compress} steps, and the approach is known as Progress and Compress. During the \textit{progress} phase, lateral connections between the knowledge base and the active column as used to transfer knowledge from the previous tasks to the current task. Only the active column is trained during the progress phase. This is similar to how~\citet{2016_progressive_neural_networks} works. However, unlike \citet{2016_progressive_neural_networks}, extra care needs to be taken to avoid catastrophic forgetting during \textit{compress} stage. To that end, the paper uses a modified version of Elastic Weight Consolidation~\citep{2017_overcoming_catastrophic_forgetting_by_incremental_moment_matching}.
Specifically, when \textit{compressing} the knowledge of the $k^{th}$ task into the knowledge base, the following loss is optimized, with respect to the parameters $\theta^{KB}$ of the knowledge base (while keeping the parameters of the active column fixed):
\begin{align}
\mathbb{E}\Big[KL(\pi_k(\cdot|x)\|\pi^{KB}(\cdot|x))\Big] + \frac{1}{2} \|\theta^{KB}-\theta^{KB}_{k-1}\|^2_{ \gamma F^*_{k-1}}\,,
\end{align}
where $\pi_k(\cdot|x)$ and $\pi^{KB}(\cdot|x)$ are the outputs of the active column (after learning on $k^{th}$ task) and knowledge base respectively. $x$ represents the input, $\mathbb{E}$ denotes the expectation over either the data under the active column, $\theta^{KB}_{k-1}$ and $F^*_{k-1}$ represent the mean and the diagonal Fisher of the online EWC Gaussian approximation resulting from previous tasks, and $\gamma$ is a hyperparameter. 

\paragraph{Relation to Modular Neural Networks.}

The general architecture of adding task-specific experts is similar to the work on modular neural networks. In fact, the high-level motivation is also very similar: when doing lifelong learning, different tasks share a common substructure or common sub-problems. We hope to capture some of these shared structures/knowledge via the experts, i.e., different experts will learn to solve different tasks, and this knowledge can be shared across tasks. In practice, mixture-of-expert systems generally do not \textit{compose} modules for a given data point or use simplistic aggregation operations like \textit{average} (similar to how ensembles work). The controller mechanism either selects $k$ experts (or assigns soft-attention scores to all the experts). The application setup is closer to \textit{system identification or task identification}~\citep{zadeh1956identification, e0da8f25-d850-4a80-af21-e151cc28c4f4, swevers1997optimal, bhat2002computing, gevers2006system, LJUNG20101, van2012subspace, chiuso2019system, ajay2019combining, yu2017osi, zhu2017fast} 
setup, and the goal is to identify the correct expert to use for a given task. In the mixture-of-experts-based setup, the controller mechanism is often non-parameterized, as we explain in the examples below.

\citet{2017_expert_gate_lifelong_learning_with_a_network_of_experts} proposed using a mixture of experts for the lifelong learning setup by training task-specific experts as follows: The model is initialized with one expert which is trained on the first task. As the model trains on subsequent tasks, new experts are added to the model and trained on the newly added tasks. This training setup does not require access to the previously seen data. Moreover, as new experts are added with new tasks, the model does not have the challenge of \textit{capacity saturation}. There are two challenges that need to be addressed:~{(i)} how to select the \textit{correct} expert during inference and~{(ii)} adding one new expert per task leads to linear growth in the number of parameters as new tasks are encountered.

The first challenge is addressed as follows: a set of gating auto-encoders whose job is to decide which expert should be used for a given sample. Specifically, each expert uses a shallow auto-encoder (which is trained with the expert). During inference, all the auto-encoders encode the input sample, and the expert, corresponding to the auto-encoder with the smallest reconstruction error, is selected. The use of auto-encoders also provides a mechanism to measure the \textit{relatedness} between the different tasks, by comparing the reconstruction error between the different encoders. Specifically, given the $i^{th}$ and the $j^{th}$ tasks, the \textit{relatedness} is given as:
\begin{equation}
    Rel(T_i, T_j) = 1 - \frac{Er_j - Er_i}{Er_i}\,,
\end{equation}
where $Er_i$ is the reconstruction error for the $i^{th}$ task. This task relatedness is used to select the most related task (for a new task) to be used as a prior model for learning the new task. Note that in the general neural modular networks, this problem is solved by the controller module and the mechanism used by~\citet{2017_expert_gate_lifelong_learning_with_a_network_of_experts} (or in general mixture-of-expert based approaches) can be seen as a non-parametric controller. 

The second challenge, of the linear growth of the number of model parameters with new tasks, violates one of the desiderata of lifelong learning systems~\ref{subsec:desidrata}: the number of parameters should increase sub-linearly with the number of tasks.~\citet{2017_expert_gate_lifelong_learning_with_a_network_of_experts} leaves this limitation for the future work.

\citet{2017_encoder_based_lifelong_learning} use mixture-of-experts based architecture by training task-specific auto-encoders for mitigating catastrophic forgetting. Their proposed solution works as follows: Let us say that the model has a shared feature extractor, a shared model trunk, and some task-specific layers (that use the activations from the trunk as the input). After training the model on the first task, an auto-encoder is trained on the representations from the first task, in order to capture the most important features from the first task. Specifically, the auto-encoder is trained with two losses:~{(i)} the reconstruction loss and~{(ii)} the supervised learning loss using the data from the first task (i.e., the features learned by the auto-encoder should be informative enough to solve the first task). When the model is trained on the second task, two constraints are added, along with the supervised learning loss on the second task. The first constraint is in the form of distillation loss used in~\citet{2017_learning_without_forgetting}. As described in~\cref{sec:distill}, the distillation loss aims to mitigate the influence of the use of different data distributions. The second constraint ensures that the features learned by the auto-encoder (for the first task) are still good for performance on the first task. This procedure can be applied to a sequence of tasks. Like~\citet{2017_expert_gate_lifelong_learning_with_a_network_of_experts}, this method also leads to linear growth in the number of parameters.~\citet{2017_encoder_based_lifelong_learning} justify the trade-off by arguing that the memory footprint of the encoders is much smaller than the memory footprint of the overall model.

\subsection{Expanding Networks}\label{sec:expansion}

An important aspect of lifelong learning that does not get as much focus is Capacity Saturation. Only a few selective works have focused on that problem~\citep{yoon2018lifelong, sodhani2020toward}. Specifically,~\citet{yoon2018lifelong} proposed Dynamically Expandable Network (DEN), that can increase the network capacity dynamically as it trains on a sequence of tasks. There are three key steps involved in training DEN: Selective retraining, Dynamic network expansion, and Network split/duplication.

\textbf{Selective retraining.} At the start, the network is trained with L1 regularization to induce sparsity. As new tasks are encountered, a sparse linear model is trained to solve the task, using the topmost hidden units of the network. The nodes/weights, that changed in the top layer, provide the starting positions for breadth-first search, to find the nodes (in the layers below) that have paths to the nodes in the topmost layer. Only the weights of the selected sub-network are trained.

\textbf{Dynamic Network Expansion.} If selective retraining is not sufficient to train the model on a given task, its capacity is increased by expansion. This step is made efficient by using group sparse regularization. Specifically, the capacity of each layer is increased by $k$ neurons. Group sparsity
regularization removes the hidden units that are not necessary for training, thus preventing the wasteful addition of neurons as new tasks are encountered. This is one significant improvement over previous approaches like Progressive Neural Networks~\citep{2016_progressive_neural_networks}. 

The last key step is \textbf{Network Split}. As the model trains through multiple tasks, the semantic drift of the neurons is tracked, and if the semantic drift becomes too high, the neuron is split into two copies. After the split/duplication step, the network is trained again (to ensure task knowledge is not lost).

Some recent works have looked into hand-designing experts based on the task/domain setup.~\citet{2020_compositional_language_continual_learning} proposed to leverage compositionality to develop a new approach for lifelong learning in sequence-to-sequence tasks. They build on the idea of decomposing syntactic and semantic representations with compositionality by learning separate representations for the semantic and syntactic knowledge~\citep{2019_compositional_generalization_for_primitive_substitutions} and the networks encoding these two representations are analogous to experts in the mixture-of-expert based systems. There is no explicit gating mechanism, and the representation from the modules is used for making the final prediction. The network encoding the syntax is trained only on the first task and not updated afterward (with the assumption that the syntax does not change across the tasks).  

~\citet{2015_net2net_accelerating_learning_via_knowledge_transfer} proposed network expansion techniques that can \textit{grow} a smaller network into a larger network using function-preserving transformations. The paper presents two variants: {(i)} Net2WiderNet (for expanding the width of a given network) and {(ii)} Net2DeeperNet (for expanding the depth of a given network). 

Consider a neural network where the $i^{th}$ and the ${i+1}^{th}$ layers are fully connected layers and layer $i$ uses a point-wise non-linearity. If the $i^{th}$ layer has $m$ inputs and $n$ outputs and the ${i+1}^{th}$ layer has $p$ outputs, then the Net2WiderNet operation can be used to widen $i^{th}$ layer by replacing it with a layer that has $N$ outputs, where $N > n$. First, a random mapping function $r$, from $\{1,2, \cdots, N\} \rightarrow \{1,2, \cdots, n\}$, is defined as follows:
\[
r(j) = \left\{
\begin{array}{ll}
j & j \leq n\\
\mbox{ random sample from } \{1,2,\cdots n\} & j > n\\
\end{array}
\right.
\]
The weight matrices $W^i$ and $W^{i+1}$ are replaced by $U^i$ and $U^{i+1}$ such that 
\[
	U^{(i)}_{k,j} =  W^{(i)}_{k, g(j)},
\]
\[
	U^{(i+1)}_{j,h} = \frac{1}{|\{x| g(x) = g(j)\}|} W^{(i+1)}_{g(j), h}
    .
\]
The first $n$ columns of $W^{i}$ are copied as it is into $U^{i}$. Columns $n+1$ through $N$ of $U^{{(i)}}$ are created by randomly selecting columns of $W$ (with replacement), followed by normalization with a replication factor (computed using the frequency of sampled columns).

Similarly, the Net2DeeperNet operation can be used to replace the $i^{th}$ layer (representing $\phi(h^{(i-1)\top}W^{(i)})$) by a deeper layer (representing $\phi(U^{(i)\top}\phi(W^{(i)\top}h^{(i-1)}$). The newly added $U$ matrix is initialized as indentity matrix and updated during finetuning. One limitation of this approach is that $\phi$ (the activation function) must satisfy $ \phi(I \phi(v)) = \phi(v) $
for all vectors $v$ where $I$ is an identity matrix. So while this operation holds for activations like ReLU, using this operation with maxout units requires some modifications to the $U$ matrix and the operation does not hold for the sigmoid activation. 

The technique is proposed in the context of accelerating the training of a large neural network by first training a smaller network and then expanding it into a larger network. However, the paper mentions lifelong learning as an application area for their approach. Indeed, subsequent works have applied the technique for alleviating capacity saturation. For example,~\citet{2018_on_training_recurrent_neural_networks_for_lifelong_learning} combined Net2Net (specifically Net2WiderNet) with Gradient Episodic Memory to create a hybrid model. When training on a given task, the model uses Gradient Episodic Memory updates to retain the knowledge of the previous task. When the model's capacity is saturated (and the model is unable to learn new tasks), it is expanded using the Net2Net approach, and the enlarged model continues to train on the subsequent tasks.~\citet{2018_on_training_recurrent_neural_networks_for_lifelong_learning} shows that the combined approach works better in practice than the individual components.

\section{Summary}
In this chapter, we introduced the architecture-based methods proposed for lifelong learning that assign a model copy to every new task that arrives. We started by discussing the idea of using Modular networks, their motivation from multiple perspectives, their architectures, and finally, their benefits to lifelong learning. Understanding a more general biologically-inspired type of method, called modular networks, is necessary because they provide the ability to use different model architectures as constituent modules, add new modules throughout training, and also help in representing composable skills.

Parameter isolation methods are the types of architecture-based methods which are similar to modular networks. The main difference is that parameter isolation methods have a different inductive bias such that the tasks have their own set of isolated parameters. They are further divided into fixed capacity networks and increasing capacity networks. Fixed capacity networks can be masking-based, pruning-based, and involve discovering paths in the network as different modules. There are expert-based methods that require adding a new network for each task. This is similar to modular networks in the sense that the task-specific experts are continuously added to the main network.

Finally, we presented the capacity saturation aspect of lifelong learning by describing the DEN method. This method avoids capacity saturation and ensures stable learning for future tasks by following three important steps: selective retraining, dynamic network expansion, and network split/duplication. 

\chapter{Benchmarks}
\label{sec_metrics_benchmarks}

Lifelong learning methods today are designed and implemented with different assumptions and setups. In earlier chapters, we described how the lifelong learning methods can be categorized from an algorithmic point of view. But benchmarking the proposed methods in different datasets with different settings makes comparing these approaches very complicated. Such untrustworthy benchmarks and comparisons raise the importance of introducing some systematic benchmarks in lifelong learning. With the availability of huge datasets in the machine learning domain, there is an abundance of benchmarks to evaluate lifelong learning methods. In this chapter, we go over some of the most commonly used benchmarks by dividing them based on their application: vision-based and NLP-based. 

\section{Vision Benchmarks}
Many methods proposed in lifelong learning are evaluated on image-based benchmarks. These benchmarks are typically adapted from fields such as image classification, reinforcement learning (Atari games, robot manipulation, imitation), and generative models \citep{lesort2019continual}. To evaluate a lifelong learning method, these benchmarks are modified, augmented, and concatenated together to create sequences of tasks. In this section, we describe such benchmarks in detail and also describe the benchmarks that are designed specifically for lifelong learning settings.

\subsection{Variants of existing datasets}

\begin{figure}[htbp!]
    \centering
    \includegraphics[width=.7\textwidth]{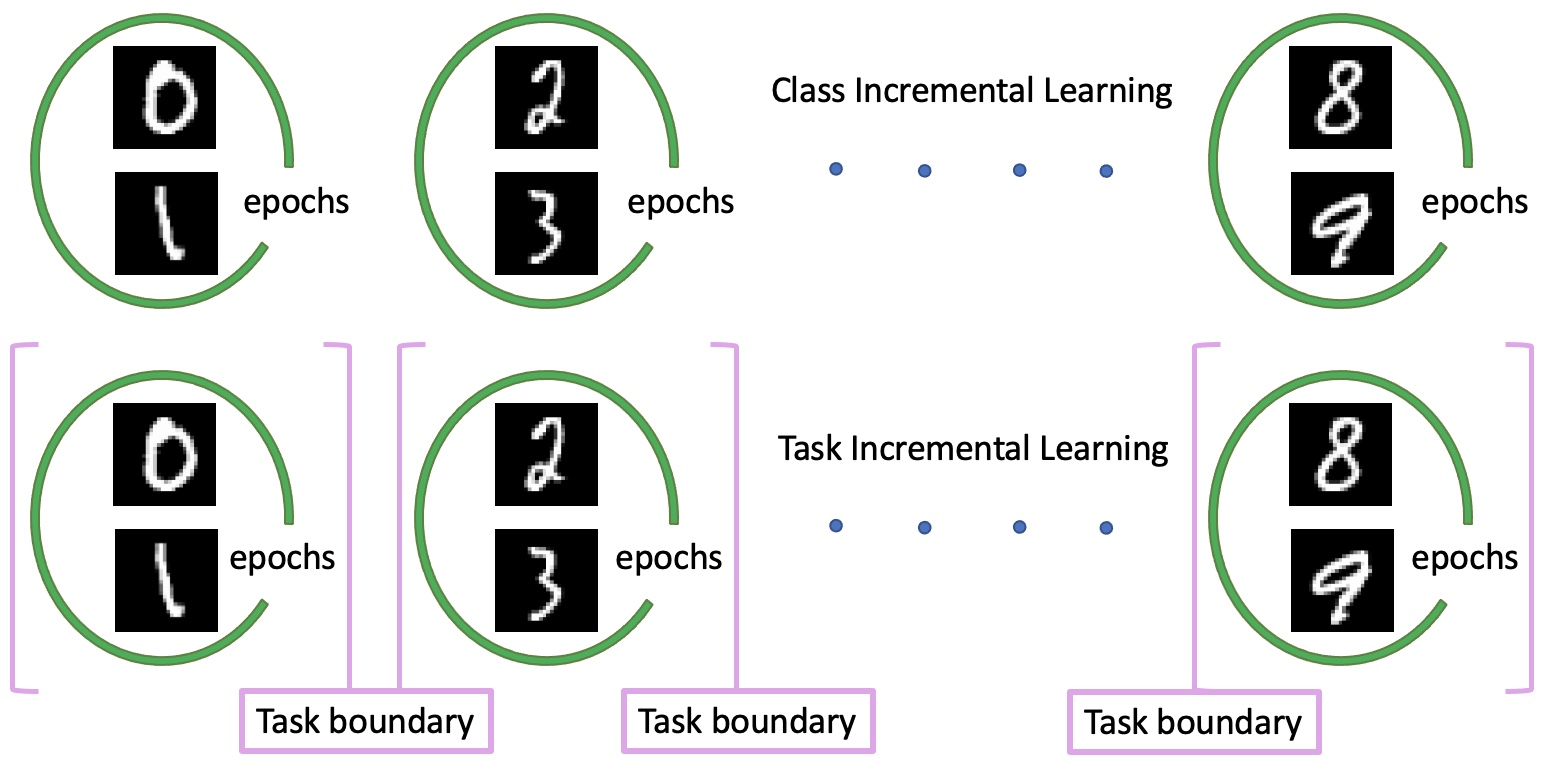}
    \caption{Class and task incremental learning on the Split MNIST benchmark.}
    \label{fig:benchmarks_overview}
\end{figure}

\begin{figure}[htbp!]
    \centering
    \includegraphics[width=.9\textwidth]{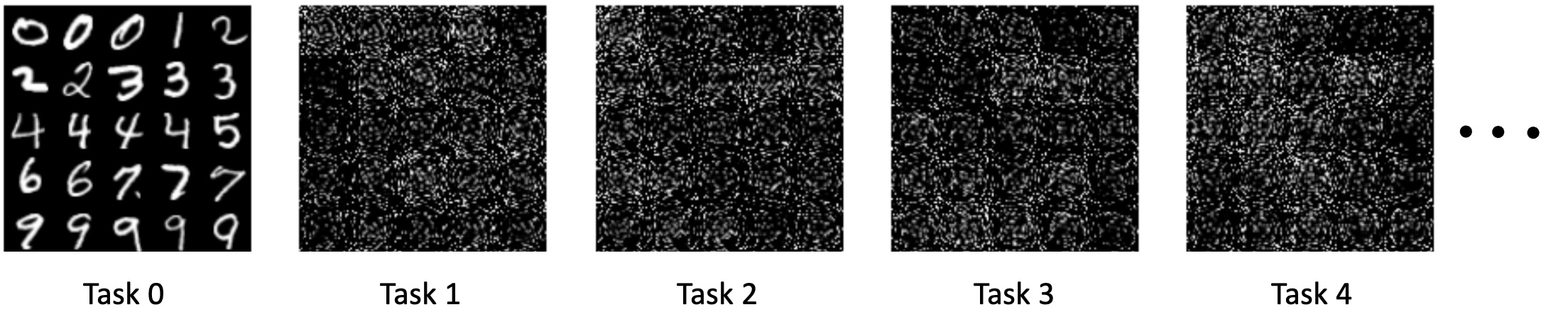}
    \caption{Permuted MNIST that have been used as a benchmark for assessing the lifelong learning methods.}
    \label{fig:benchmmarks_permuted_mnist}
\end{figure}

\paragraph{MNIST:} 
Early lifelong learning methods, with the focus on vision tasks, started with benchmarking methods on some variation of the MNIST dataset~\citep{lecun2010mnist}. Permuted MNIST \citep{goodfellow2015empirical} and Split MNIST \citep{lopez2017gradient} were some of the early benchmarks for lifelong learning.
Figure~\ref{fig:benchmarks_overview} illustrates a simple training protocol on the Split MNIST benchmark. It shows both class and task incremental learning cycle. MNIST dataset contains training samples for supervised learning of classification of handwritten digits zero to nine. 
In Split MNIST benchmark, the tasks are basically the disjoint sets of classes from the MNIST dataset. The model has to learn from the training samples that arrive sequentially from these tasks at each training step. Following the supervised learning, the model iterates over given training samples for several epochs and is evaluated on all classes seen so far. In the Split MNIST benchmark~\citep{pmlr-v70-zenke17a,farquhar2019robust,aljundi2019online,vandeven2019generative,swaroop2019improving,vandeven2019scenarios}, the difference between task and class incremental learning is the awareness of the model to the task shift. To this end, in the task incremental learning, trained on Split MNIST, the model knows the tasks' boundaries.
Permuted MNIST is another benchmark that was introduced after the Split MNIST benchmark. The approach of creating the Permuted MNIST benchmark is straightforward. In this case, the model receives all training samples of ten digits at each training time. The model learns from a regular MNIST dataset as the first task. Then, the model receives the permuted version of regular MNIST as the second task. So the model should learn from the permuted image sample and also should not forget what it learned at the previous task. Similarly, in the next steps, the model will receive samples that have different permutations and will have to adapt without forgetting catastrophically.  Figure~\ref{fig:benchmmarks_permuted_mnist} shows the flow of training samples that the model should learn overtime in the Permuted MNIST benchmark. 

\begin{figure}[htbp!]
    \centering
    \includegraphics[width=.9\textwidth]{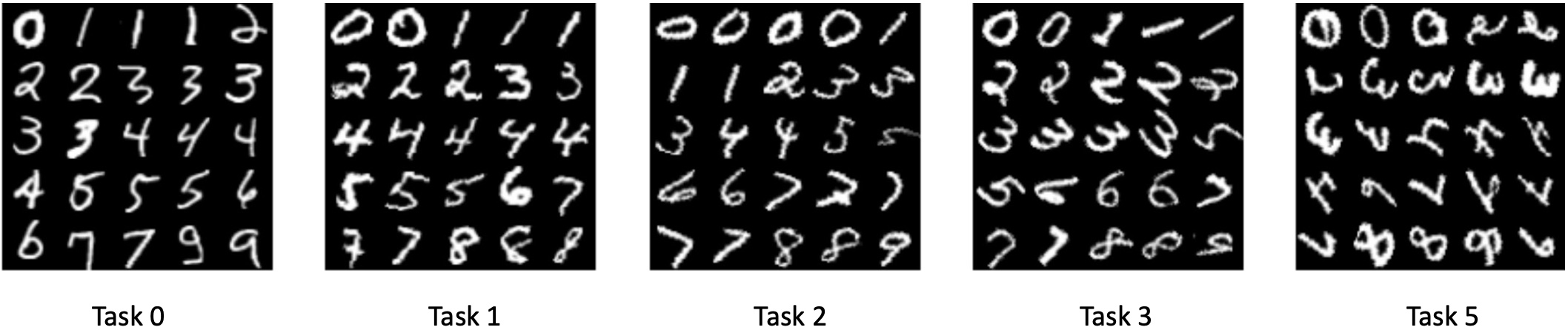}
    \caption{Rotated MNIST benchmark. Each block shows few samples that the model should learn at each training step.}
    \label{fig:benchmark_rotated_mnist}
\end{figure}

\citet{lopez2017gradient} introduced and used the Rotated MNIST benchmark to evaluate their Gradient of Episodic Memory (GEM) method. Rotated MNIST is a more practical and meaningful benchmark than the Permuted MNIST. In this benchmark, the goal of the model is to have a robust behavior through time by learning from a sequence of tasks that differ by rotation transformation. In other words, the images in each task are MNIST images with some fixed degree rotation transformation. Figure~\ref{fig:benchmark_rotated_mnist} gives an example of using Rotated MNIST in a lifelong learning setup. 

KMNIST is a dataset, adapted from Kuzushiji Dataset that consists of Kuzushiji-MNIST, Kuzushiji-49, and Kuzushiji-Kanji datasets~\citep{clanuwat2018deep}. 
KMNIST Dataset is a drop-in replacement for the MNIST dataset.
KMNIST chooses one character to represent each of the 10 rows of Hiragana. KMNIST adds some complexity to the Split MNIST benchmark since the shape of characters is more complex than the shape of simple digits.

\begin{figure}[htbp!]
    \centering
    \includegraphics[width=.9\textwidth]{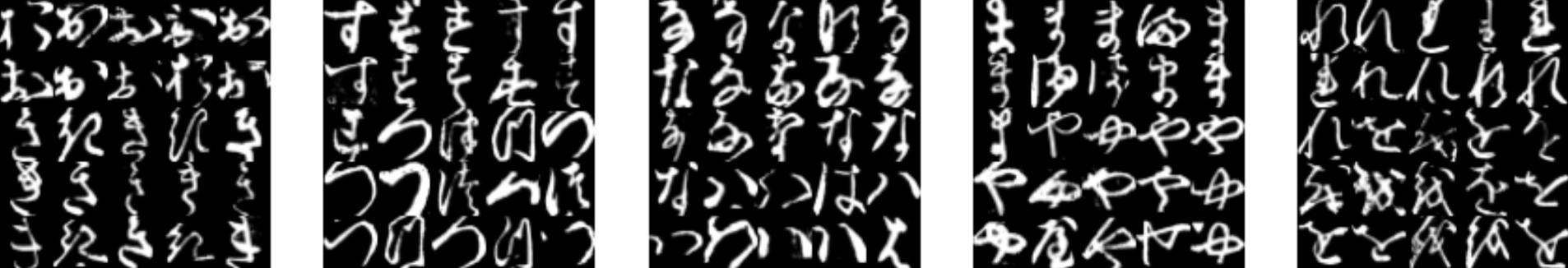}
    \caption{Split KMNIST sample that can be used in a lifelong learning task.}
    \label{fig:benchmarks_kmnist}
\end{figure}

\begin{figure}[htbp!]
    \centering
    \includegraphics[width=.9\textwidth]{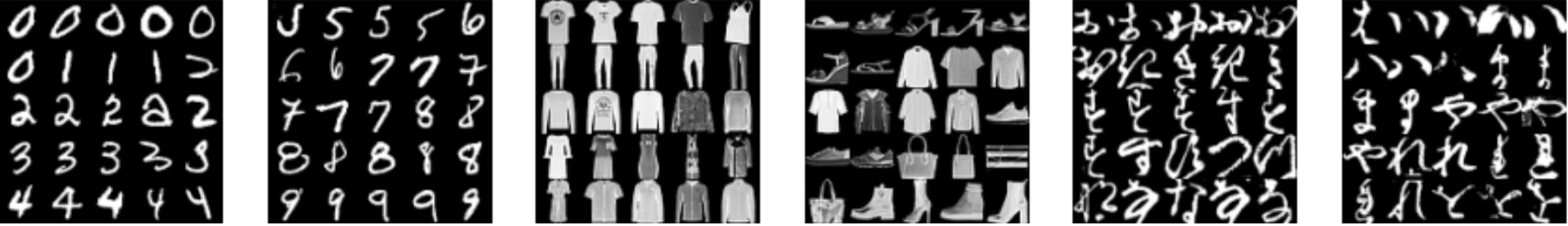}
    \caption{MNIST Fellowship benchmark. Each task could be created using a subset of either MNIST, Fashion MNIST, or KMNIST datasets.}
    \label{fig:benchmarks_MNISTFellowship}
\end{figure}

Figure~\ref{fig:benchmarks_MNISTFellowship} shows the combination of three datasets: MNIST, FashionMNIST (dataset of the articles of clothing at low resolution~\citep{xiao2017_online}), and KMNIST datasets in form of a sequence of three tasks also known as the MNIST Fellowship benchmark. In this benchmark, each task is a variation of the MNIST dataset that arrives in a sequence. MNIST Fellowship has 30 classes in total and each instance is in the size of 28x28 images.

\paragraph{ILSVRC2012 and CIFAR:} 
As explained above Split MNIST, Permuted MNIST, KMNIST, and MNIST Fellowship benchmarks are the simple benchmarks that have been used in early research in lifelong learning. As the  field grew, having a more complex benchmark became crucial for evaluating lifelong learning methods. To create more complex and difficult benchmarks split versions of other popular datasets like CIFAR and ILSVRC2012 have been used. 
The performance of the models trained on lifelong learning settings are mostly reported for MNIST~\citep{lecun2010mnist}, Permuted MNIST, rotated MNIST, CIFAR-10, CIFAR-100~\citep{krizhevsky2009learning}, ImageNet~\citep{deng2009imagenet} where data is split into sequences of classes or tasks. MNIST, Permuted MNIST, and rotated MNIST are usually split into two and five consecutive sets of classes\citep{pmlr-v70-zenke17a, ewc, lopez2017gradient, pmlr-v70-zenke17a, nguyen2017variational, lesort2018generative}. CIFAR-100 is split into two, five, ten, or twenty sets of classes such that the model should learn each set of classes at each time consecutively~\citep{lopez2017gradient, pmlr-v70-zenke17a, Rebuffi_2017,Hou_2019_CVPR, castro2018endtoend}. Recently, most of the approaches report the performance of models on split ImageNet~\citep{Rebuffi_2017,Hou_2019_CVPR,castro2018endtoend,wu2019large} and Celeb-10000~\citep{wu2019large}. Some approaches benchmark on datasets such as ImageNet, CIFAR-100, SVHN, UCF101, Omniglot, GTSR, DPed, Flower, Aircraft, and DTD~\citep{li2019learn}.

\paragraph{5-datasets}: 
Recently proposed methods benchmark using several existing datasets as a sequence of individual tasks. For instance, \textbf{5-datasets} is a sequence of five different datasets as five 10-way classification tasks \citep{2018_overcoming_catastrophic_forgetting_with_hard_attention_to_the_task,saha2021gradient,ebrahimi2020adversarial}. These datasets are: \textbf{CIFAR-10}, \textbf{MNIST}, \textbf{SVHN} \citep{netzer2011reading}, \textbf{notMNIST} \citep{notmnist}, and \textbf{Fashion-MNIST} \citep{xiao2017_online}.

\subsection{CORe50}
\citet{lomonaco2017core50} introduced CORe50 that is constructed specifically to evaluate lifelong learning methods. Continual object Recognition benchmark (known as CORe50) consists of 50 domestic objects belonging to 10 categories of simple objects including plug adapters, mobile phones, scissors, light bulbs, cans, glasses, balls, markers, cups, and remote controls. Since the lifelong learning setup pose significant challenges for deep learning models, having clean object-centered instances may reduce learning complexity at each training step. With this idea, \citet{lomonaco2017core50} attempted to collect clean images centered by the main object and avoid having other objects in the same image like in ImageNet. 

\begin{figure}[htbp!]
    \centering
    \includegraphics[width=.8\textwidth]{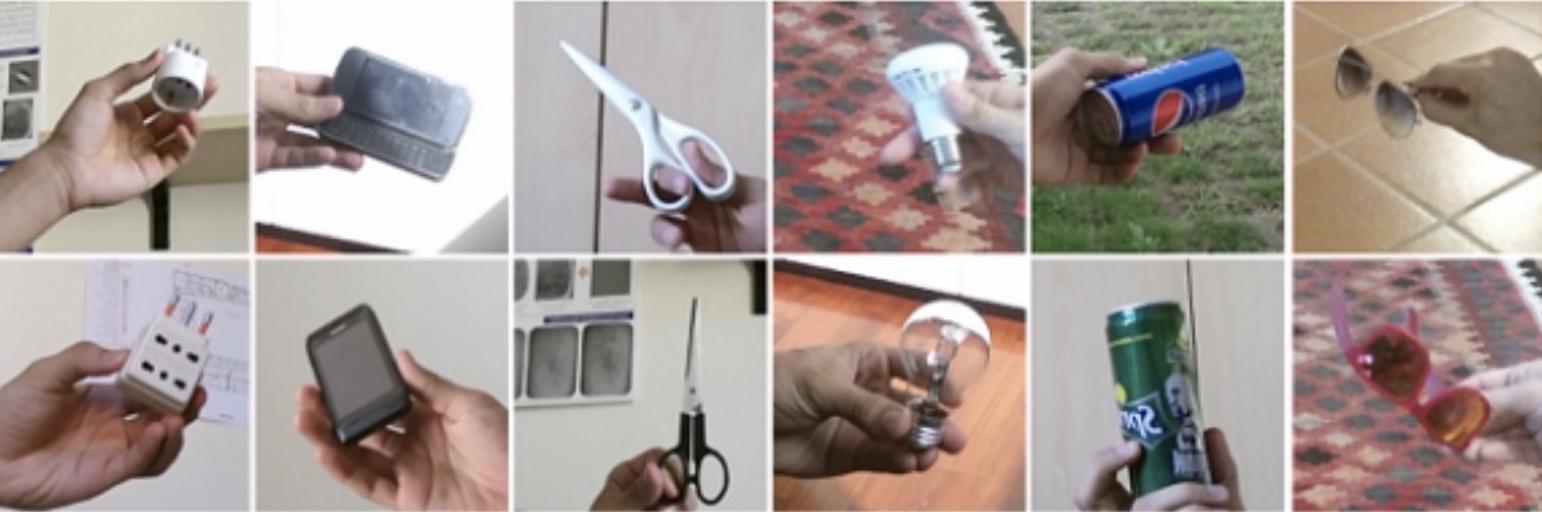}
    \caption{Some examples of CORe50 training samples. The grabbing hand (left or right) changes while collecting images of different objects.}
    \label{fig:benchmarks_core50}
\end{figure}

Figure~\ref{fig:benchmarks_core50} shows some examples of CORe50 training samples. As shown in Figure~\ref{fig:benchmarks_core50} objects are presented in each instance such that the camera point-of-view mimics the operator's eyes point of view. In this way, models can learn a training sample that is simple and provided to the model clearly with minimum complexity.
The original benchmark was designed for studying lifelong learning in the robotic domain and it contained short videos instead of images. In order to collect samples, the operator smoothly moves his/her arm to present objects from different angles. It is worth mentioning that the operator changes hands throughout the sessions.

This provides a chance to produce more samples for relevant objects.  
\citet{lomonaco2017core50} collected data in 11 distinct sessions that included samples from eight indoor and three outdoor views designated by different backgrounds and lighting. Each object in the corresponding session presented by a 15 seconds video (at 20 fps) has been recorded with a Kinect 2.0 sensor delivering 300 RGB-D frames.

For low-data streams on lifelong learning setup, \citet{antoniou2020defining} introduced a benchmark that defines a systematic approach for the setting of continual few-shot learning on various datasets such that the performance of the model is reported on each dataset individually. The learning agent has access to a very limited set of training samples in most real-world scenarios in each task as described by~\citet{antoniou2020defining}. However, most proposed approaches in lifelong learning need to revisit training samples for several epochs at each training step. That could be considered as the first limitation of the above benchmarks. To this end, benchmarking the lifelong learning methods by using the introduced benchmarks in this chapter such that visiting the training samples are allowed once or for a few numbers of epochs~\citep{hayes2020remind,laleh2020chaotic} are interesting and challenging approaches to evaluating lifelong learning methods. 

The second limitation of the current approaches is that they alleviate catastrophic forgetting through time for incrementally learned classes or tasks that arise from the same distribution and same dataset. 
To this end, current approaches have been focused on a homogeneous lifelong learning problem. To make the benchmark more realistic and closer to real-life scenarios some other benchmarks have been proposed recently to include more challenges for assessing lifelong learning methods. In the next section, we introduce some of the recently proposed benchmarks that try to overcome one or both limitations explained above. 

\subsection{CRIB}
\citet{Stojanov_2019_CVPR} introduced CRIB benchmark that is motivated by infant learning concepts in psychology. Infant learning is characterized based on five pillars: Incremental learning, Repetitive exposure to objects, Temporally contiguous visual experience, self-supervised learning (since labeling events are sparse and noisy), and Object Instance learning that precedes categorization. The infant object learning process consists of repetitive exposure to the object, starting with toys in their environment, which is an inherently incremental procedure. The visual experience that infants can see is temporary continuous and poorly smooth, particularly for the first two years. Their supervision is fairly sparse and they do not have constant object naming supervision by their parent.
Infants tend to pick up different toys while playing, explore them for a few minutes, and then put them down. This is a continuous pattern observed in infants.

Therefore, they may pick up a bunch of different objects that only get naming supervision from their parents for a small subset of them. \citet{Stojanov_2019_CVPR} tried to connect incremental learning with infant learning by having a simple model of object interaction by picking up and putting down objects during episodes. They called picking an object up and examining it for a while, then putting it down and picking another as one learning exposure.

\begin{figure}[htbp!]
    \centering
    \includegraphics[width=.8\textwidth]{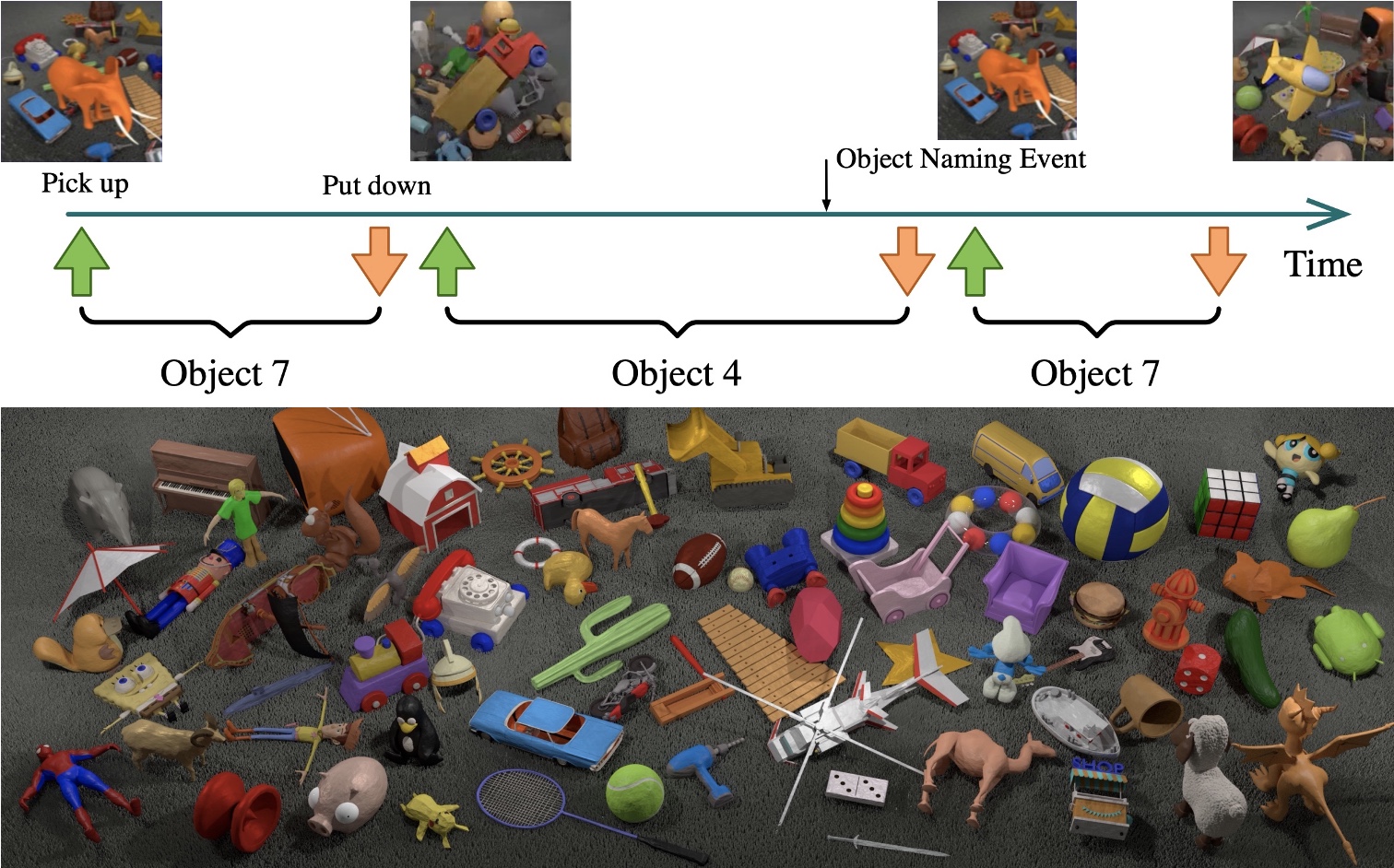}
    \caption{CRIB benchmark. Toys-200 dataset of 200 unique toy objects that model infants learning in a lifelong learning approach.}
    \label{fig:benchmakrs_crib}
\end{figure}

The object can be presented to the learner after a while again. But if it has been seen before, the label will not be provided to the learner. That can mimic the sparse supervision situation. To model such a procedure, they used Toys-200 3D that contains object samples in the dataset with a toy-like appearance. They generate a small smooth video of the picked-up object and rotate it in front of a background in various lighting conditions and then put it down and select another toy object. The selected object in each training sample will not appear in the background. Figure~\ref{fig:benchmakrs_crib} illustrates the CRIB benchmark idea and the exposure learning process in the lifelong learning approach.   

\subsection{OpenLORIS-Object}
\label{sec:OpenLORIS-Object}
OpenLORIS is another well-known benchmark that simulates closely a real-life scenario for a lifelong learning agent~\citep{she2019openlorisobject}. (L)ifel(O)ng (R)obotic V(IS)ion (OpenLORIS) is an Object Recognition benchmark that is designed for facilitating lifelong learning research primarily for the robotic domain and extended to other application domains as well. This benchmark examines the capability of learning the common objects in the home scenario in some limited conditions. Since fully retraining robotic agents at each time for a new task or same task but with different lighting in the environment or different point of view is infeasible, lifelong learning methods can help to overcome such challenges but alleviate catastrophic forgetting and evaluating proposed methods in such conditions need benchmarks that can assess the agents' capabilities in these conditions. \citet{she2019openlorisobject} include the common challenges that the learning agents might need to deal with in the environment such as changes in illumination, occlusion, object size, camera-object distance, camera-object angle, and clutter. Including these factors from real-life environments into the benchmark creates more realistic scenarios for assessing a lifelong learning agent's capabilities.
\begin{figure}[htbp!]
    \centering
    \includegraphics[width=1.\textwidth]{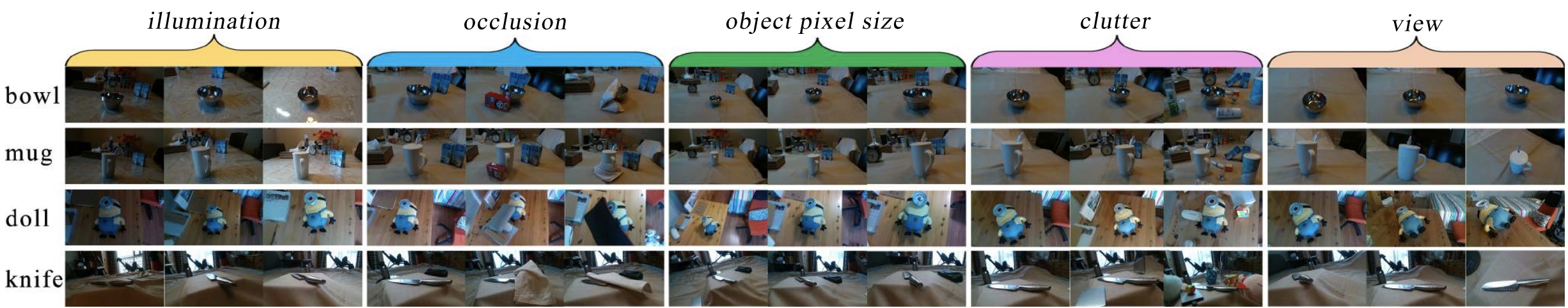}
    \caption{OpenLORIS-Object benchmark.}
    \label{fig:OpenLORIS}
\end{figure}

\subsection{Stream-51}
\label{sec:benchmarks_tream-51}

Conventional neural networks require running a loop over a batch of i.i.d data multiple times to improve performance. Although several lifelong learning training methods are proposed to alleviate forgetting under non-i.i.d. conditions, many of these methods are inflexible and inefficient in real-world scenarios where data comes from a dynamic data distribution and constantly changes over time. 

Stream-51 is used to evaluate whether an agent can robustly handle shifts and novel inputs in training data in a lifelong learning scenario. Online streaming learning is a more pragmatic approach where a model needs to learn one sample at a time that it receives from a stream of data. Stream-51 provides sufficient data classes with high-quality training instances for such an online lifelong learning setup.

\begin{figure}[htbp!]
    \centering
    \includegraphics[width=.8\textwidth]{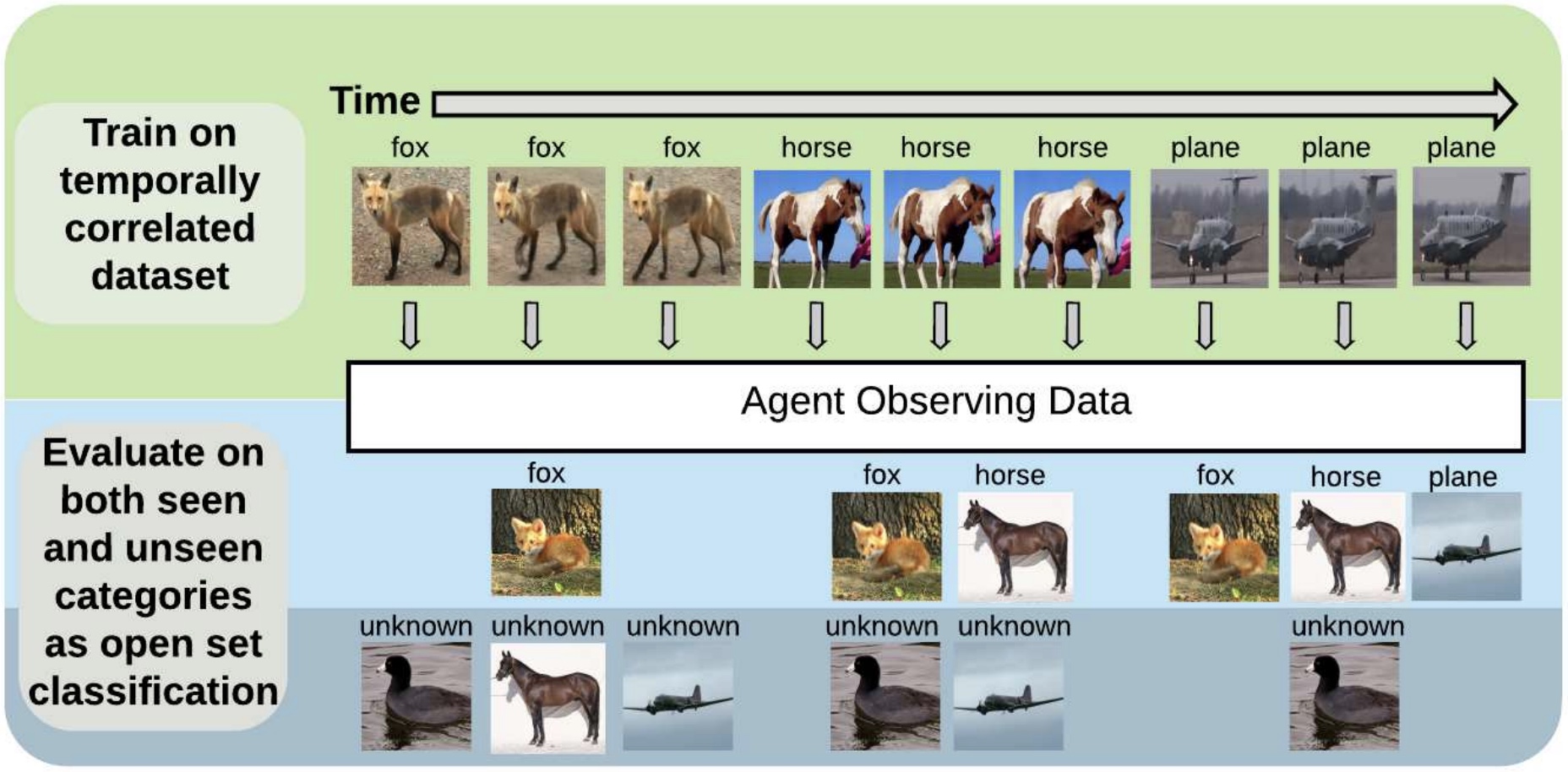}
    \caption{Stream-51}
    \label{fig:benchmarks_stream51}
\end{figure}

This benchmark consists of temporally correlated images from 51 unique object categories and additional evaluation classes for testing novelty recognition. The evaluation samples are not provided to the model at training time. As Figure~\ref{fig:benchmarks_stream51} shows, the model learns from a temporally correlated stream of samples. Similar to the other benchmarks, the model is evaluated on the previously seen classes and the ability of the model to detect unlearned concepts. 

\textbf{Stream-51 Specific Metrics}: 
In the Stream-51 benchmark~\ref{sec:benchmarks_tream-51}, the model receives samples in a stream of data. A few of these samples are considered unseen data or novel samples. Therefore the goal is to compute the overall classification performance and its ability to detect novel inputs \citep{Roady_2020_Stream51}. Since Stream-51 works in the stream of data where the model is not allowed to iterate over the training samples for several epochs, the overall classification performance is computed as follows:
\begin{align}
    \Omega_{\mathrm{Classif.}}=\min \left(1, \frac{1}{T} \sum_{t=1}^{T} \frac{\alpha_{t}}{\alpha_{\mathrm{offline}, \mathrm{t}}}\right),
\end{align}
where $T$ is the total number of testing events, $\alpha_{t}$ is the accuracy of the streaming learner at time $t$, and $\alpha_{\text {offline }, t}$ is the accuracy of an optimized offline model at time $t$. This metric normalizes a streaming learner's performance using an optimized offline learner. Normalizing the streaming learner's performance to an offline learner makes the metric easier to interpret across various orderings~\citep{Roady_2020_Stream51}.
For novelty detection,~\citet{Roady_2020_Stream51} propose an incremental variant of the area under the OSC curve
(AUOSC) which normalizes an incremental learner’s performance to an optimized offline baseline as follows:
\begin{align}
    \Omega_{\mathrm{AUOSC}}=\min \left(1, \frac{1}{T} \sum_{t=1}^{T} \frac{\gamma_{t}}{\gamma_{\mathrm{offline}, \mathrm{t}}}\right),
\end{align}
where $T$ is the total number of testing events, $\gamma_{t}$ is the AUOSC score of the incremental learner at time $t$, and $\gamma_{\text {offline }, \mathrm{t}}$ is the AUOSC score of the optimized offline learner at time $t$~\citep{Roady_2020_Stream51}. 

\subsection{IIRC}
\citet{Abdelsalam_2021_CVPR} introduced Incremental Implicitly-Refined Classification (IIRC) as an extension to the class incremental learning setup. Unlike other lifelong learning benchmarks, IIRC breaks the assumption of having the same level of hierarchy for the samples the model should learn from through time. IIRC provides a benchmark and scenario which is more challenging and more aligned with real-life learning scenarios. 

In this setup, the incoming batches of classes might have two levels of granularity: a high level (coarse) label like “bear” and a low-level (fine) label like “polar bear”. Only one label is provided at a time, and the model has to figure out the other label if it has already learned it. Therefore, the model should be able to expand its knowledge about a concept while not forgetting high-level information that is previously learned knowledge (at a different granularity) about the same concept. Another challenge included in the IIRC benchmark is that the classes have imbalanced sample distribution similar to real-life where not all classes are observed at the same frequency.

IIRC benchmark provides IIRC-ImageNet and IIRC-CIFAR built based on ImageNet and CIFAR datasets respectively. These two datasets are most popular in lifelong learning literature. IIRC-ImageNet simulates data diversity challenges. They provided a shorter version of IIRC-ImageNet called IIRC-ImageNet-li. \citet{Abdelsalam_2021_CVPR} clearly states that the IIRC-ImageNet-lite version is not for benchmarking the model performance but only for performing and debugging experiments.    

\begin{figure}[htbp!]
    \centering
    \includegraphics[width=.9\textwidth]{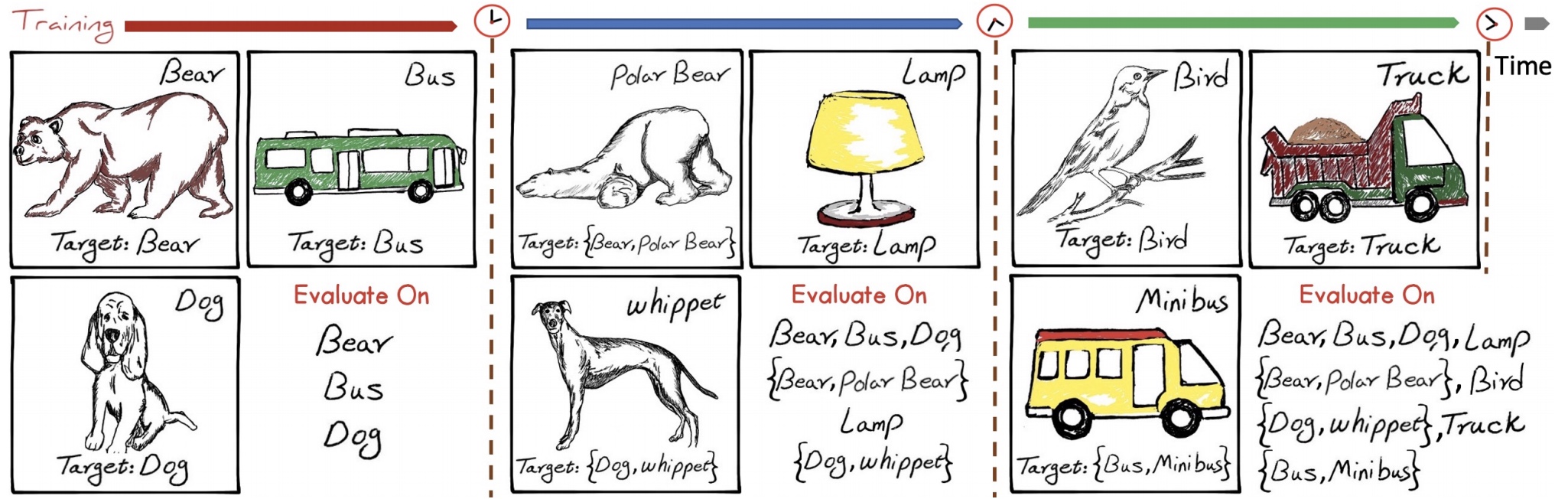}
    \caption{The learning and evaluation procedure in the IIRC benchmark.}
    \label{fig:benchmarks_iirc}
\end{figure}

IIRC benchmark reveals the model's ability to expand its knowledge and associate and re-associate labels over time. Figure~\ref{fig:benchmarks_iirc} illustrates the training and evaluation paradigm in the IIRC benchmark. In the figure, the top right label is available to the label model during training, whereas the bottom label, defined as a target, is predicted by the model during evaluation. The right bottom panel also shows the set of classes used for model evaluation, and the dashed line represents the task boundary in the task incremental learning setup.

\textbf{IIRC Specific Metrics}: 

As discussed previously, the Average Accuracy metric is one of the most popular metrics used in lifelong learning. But multi-label classification setup that involves hierarchy requires a different evaluation metric. IIRC, a benchmark closer to real-life scenarios, enabled the multi-labels classification for super-classes and sub-classes that the model might see over its lifetime. In the IIRC benchmark, limiting the model to not predicting a large number of possible classes is essential, otherwise the model will tend to predict more labels to receive more positive feedback during its supervision periods.

\citet{Sorower10aliterature} proposes to use Exact-Match Ratio (MR) as a metric for the multi-label classification. The Exact-Match Ratio (MR) is defined and computed as:
\begin{align}
    MR = \frac{1}{n} \sum_{i=1}^{n} I(Y_{i}==\hat{Y}_i),
\end{align}
where $I$ is the indicator function, $Y_{i}$ are the ground truth labels for sample $i$, $\hat{Y}_i$ is the set of predictions for the corresponding sample, and $n$ is the total number of samples. The Exact-Match Ratio penalizes partial correct predictions in the same way as completely incorrect ones. In other words,  a partially corrected prediction will have the same score as the completely incorrect one. That is an important weakness of the Exact-Match Ratio metric.

In partial multi-labels or complete multi-label classification literature using Jaccard Similarity (JS)~\citep{Sorower10aliterature} is very common. The Jaccard Similarity (JS) computes the score for evaluating a model's performance for the multi-labels prediction as intersection over the union of true labels. It is mathematically formalized as follows: 
\begin{align}
    JS = \frac{1}{n} \sum_{i=1}^{n} \frac{|Y_i \cap \hat{Y}_i|}{|Y_i \cup \hat{Y}_i|}.
\end{align}
The precision weighted JS (pw-JS) is further weights JS by sampling precision. Similar to the other lifelong learning benchmarks, after training the model on the task $j$, pw-JS is computed to evaluate the model performance on all tasks from the beginning to task $j$ as follow:
\begin{align}
    R_j = \frac{1}{n} \sum^n_{i=1}\frac{|Y_i\cap \hat{Y_i}|}{|Y_i \cup \hat{Y_i}|}\times \frac{|Y_i\cap \hat{Y_i}|}{ | \hat{Y_i}|},
\end{align}
where $n$ is the total number of evaluation samples for all the tasks seen so far, $Y_i$ is the set of ground truth labels of sample $i$ and $\hat{Y_i}$ is the set of predictions from the model for sample $i$~\citep{Abdelsalam_2021_CVPR}. 
Compared to JS, the weights additionally differentiate completely correct predictions from partially correct predictions. \citet{Abdelsalam_2021_CVPR} show that algorithms that tend to generate more labels have lower pw-JS scores than JS scores. 

\section{NLP Benchmarks}
Natural language processing is used in many day-to-day applications that take the advantage of ML to solve tasks. With the surge of attention on using lifelong learning methods in practical scenarios, having a good benchmark to evaluate proposed methods is a key element in applying the lifelong learning methods in the NLP domain. Several benchmarks have been proposed recently in this direction and we discuss some of them in this section.

\subsection{Personalized Online Language Learning and FIREHOSE datasets}
\citet{hu2020drinking} collected data from the popular social media platform Twitter to introduce a benchmark for evaluating a personalized language model for each user. Each user is considered as a different task in this setup. The goal here is to propose a Personalized Online Language Learning (POLL) method that helps in finding a personalized language model for each user over time. The key characteristic of this setup is that users are added and dropped over time. Since users tweet with different frequencies, the setup has a highly non-stationary distribution. This setup provides both multi-task and continual settings of tasks. Learning the language model for each user is an online multi-task setting since each user is considered a task in this setup. Since it requires learning a shared language model with the non-stationary data distribution, it is a lifelong learning problem ~\citep{hu2020drinking}.    

\citet{hu2020drinking} collected more than $100$ million tweets with more than $1.5$ billion tokens, posted by one million users over six calendar years called the FIREHOSE datasets. Since language learning is an unsupervised or semi-supervised problem in real life, this dataset does not require human or automatic labeling process to create the benchmark for evaluating lifelong learning algorithms. Therefore, FIREHOSE is a massive web-scale dataset that can also support research on POLL. \citet{hu2020drinking} provided a smaller version of the dataset called Firehose10M and the bigger one as Firehose100M. Table~\ref{tab:benchmarks_NLP_Firehose} provides the statistic for each version.

\begin{table}[hbt]
    \centering
    \caption{The statistic for both Firehose10M and Firehose100M~\citep{hu2020drinking}.}
    \resizebox{.7\textwidth}{!}{\begin{tabular}{@{}lrrr@{}}
        \toprule
         &  \# Users &  \# Tweets & \# Tokens \\ \midrule
        {Firehose10M}  & $94.0$K & $10.4$M & $173.3$M \\
        {Firehose100M} & $917.4$K & $100.4$M & $1672.7$M \\
        \bottomrule
    \end{tabular}}
    \label{tab:benchmarks_NLP_Firehose}
\end{table}

\subsection{Text Classification and Question Answering Datasets}
Text classification tasks have a lot of practical and industrial usage. Therefore, having a lifelong learning benchmark to evaluate proposed method's performance in continual text classification tasks seems quite important. \citet{de2019episodic} used the publicly available text classification datasets that are collected from various domains such as news classification, sentiment analysis, Wikipedia article classification, and questions and answers categorization. They constructed a benchmark using AGNews (4 classes), Yelp (5 classes), DBPedia (14 classes), Amazon (5 classes), and Yahoo (10 classes) datasets.
Table~\ref{tab:benchmarks_NLP_text} shows the benchmark specification in detail. 

\begin{table}[htbp!]
    \centering
    \caption{The datasets that are used in a lifelong learning Benchmark proposed to evaluate methods on text classification and question answering for consecutive tasks~\citep{de2019episodic}.}
    \label{tab:benchmarks_NLP_text}
    \resizebox{.8\textwidth}{!}{\begin{tabular}{@{}llc@{}}
    \toprule
    Dataset & Task type                            & \multicolumn{1}{l}{\# Classes} \\ \midrule
    AGNews  & news classification                  & 4                              \\
    Yelp    & sentiment analysis                   & 5                              \\
    Amazon  & sentiment analysis                   & 5                              \\
    DBPedia & Wikipedia article classification     & 14                             \\
    Yahoo   & questions and answers categorization & 10                             \\ \bottomrule
    \end{tabular}}
\end{table}

\begin{table}[htbp!]
\centering
\caption{The datasets that are used in a challenging lifelong learning benchmark proposed to evaluate methods on text classification \citep{mehta2021empirical}. All tasks are either single sentence or sentence pair classification. \# Train, \# Dev, \# Test denotes the number of examples in train, valid, test splits respectively. \# L denotes the number of classes for each tasks.}
\label{tab:benchmarks_15NLP_text}
\begin{small}
\resizebox{\columnwidth}{!}{
\begin{tabular}{cccrrrrc}
\hline
Dataset & Task          & \begin{tabular}[c]{@{}c@{}}Source \\  Domains(s)\end{tabular}           & \# Train & \# Valid & \# Test & \# L & Metrics  \\ \hline
CoLA \citep{warstadt2019neural}   & \begin{tabular}[c]{@{}c@{}}Linguistic \\  Acceptability\end{tabular} & \begin{tabular}[c]{@{}c@{}}Journal \\ articles \\ and books\end{tabular} & 7,695    & 856      & 1,043   & 2    & \begin{tabular}[c]{@{}c@{}}Matthews \\ correlation \end{tabular} \\ \hline 
BoolQ \citep{clark2019boolq}    & \begin{tabular}[c]{@{}c@{}}Boolean \\ Question \\ Answering \end{tabular} & \begin{tabular}[c]{@{}c@{}}Google \\ queries, \\ Wikipedia \\ passages \end{tabular} & 8,483   & 944   & 3,270  & 2   & Acc. \\ \hline
SST-2 \citep{socher2013recursive}    & \begin{tabular}[c]{@{}c@{}}Sentiment \\ Analysis\end{tabular} & \begin{tabular}[c]{@{}c@{}}Movie\\ reviews\end{tabular}  & 9,971   & 873   & 872    & 2   & Acc.        \\ \hline
QQP \citep{wang2018glue} & \begin{tabular}[c]{@{}c@{}}Paraphrase \\ Detection\end{tabular}  & \begin{tabular}[c]{@{}c@{}}Quora \\ questions \end{tabular}    & 10,794  & 4,044 & 4,043  & 2   & Acc. \& F1  \\ \hline
\begin{tabular}[c]{@{}c@{}} YahooQA \\ \citep{zhang2015character}  \end{tabular} & \begin{tabular}[c]{@{}c@{}}Q \& A \\ Categories\end{tabular}  & \begin{tabular}[c]{@{}c@{}}Yahoo! \\ Answers \end{tabular}     & 13,950  & 4,998 & 4,998  & 10  & Acc.    \\ \hline
Yelp \citep{zhang2015character} & \begin{tabular}[c]{@{}c@{}}Review \\ Rating \\ Prediction \end{tabular}            &  \begin{tabular}[c]{@{}c@{}}Business \\ reviews \end{tabular}    & 12,920  & 3,999 & 3,998  & 5   & Acc.        \\  \hline
\begin{tabular}[c]{@{}c@{}} Decomp \\ \citep{poliak2018collecting}    \end{tabular} &  \begin{tabular}[c]{@{}c@{}}Event \\ Factuality \end{tabular}        & FactBank                  & 10,176  & 4,034 & 3,934  & 2   & Acc.        \\ \hline
AAC \citep{stab2018cross}     & \begin{tabular}[c]{@{}c@{}}Argument \\ Aspect \\Detection \end{tabular} & \begin{tabular}[c]{@{}c@{}}Web \\documents \end{tabular} & 10,893  & 2,025 & 4,980  & 3   & Acc. \& F1  \\ \hline
\begin{tabular}[c]{@{}c@{}}DISCONN8 \\ \citep{prasad2019penn, kim2020implicit} \end{tabular}
 & \begin{tabular}[c]{@{}c@{}}Explicit \\ Discourse \\Marker \\Prediction \end{tabular} & \begin{tabular}[c]{@{}c@{}}Penn \\ Discourse \\ TreeBank \end{tabular}     & 9,647   & 1,020 & 868    & 8   & Acc. \& F1  \\ \hline
QNLI \citep{wang2018glue} & \begin{tabular}[c]{@{}c@{}}Question \\ Answering \\ NLI \end{tabular}    & Wikipedia          & 9,927   & 5,464 & 5,463  & 2   & Acc.        \\ \hline
\begin{tabular}[c]{@{}c@{}} RocBSO \\ \citep{mostafazadeh-etal-2016-corpus}  \end{tabular} & \begin{tabular}[c]{@{}c@{}}Binary \\ Sentence \\ Order \\ Prediction \end{tabular}  &  \begin{tabular}[c]{@{}c@{}}Roc story \\ corpus \end{tabular},         & 10,000  & 2,400 & 2,400  & 2   & Acc.  \\ \hline
MNLI \citep{williams2018broad} & \begin{tabular}[c]{@{}c@{}}Natural \\ Language \\ Inference \end{tabular} & \begin{tabular}[c]{@{}c@{}} speech, \\ fiction, \\ govt. reports \end{tabular}    & 11,636  & 4,816 & 4,815  & 3   & Acc.  \\ \hline 
\begin{tabular}[c]{@{}c@{}} SciTAIL \\ \citep{khot2018scitail}    \end{tabular}  & \begin{tabular}[c]{@{}c@{}}Multi-choice \\ Science QA \end{tabular}  & \begin{tabular}[c]{@{}c@{}}Science \\ exams \end{tabular} & 11,145  & 1,305 & 1,304  & 2   & Acc.        \\ \hline
\begin{tabular}[c]{@{}c@{}}PDTB2L1 \\ \citep{prasad2019penn, kim2020implicit} \end{tabular} & \begin{tabular}[c]{@{}c@{}} Implicit \\ Discourse \\ Relation \\ Classification \end{tabular} &   \begin{tabular}[c]{@{}c@{}} Penn \\ Discourse \\ TreeBank \end{tabular} & 13,046  & 1,183 & 1,046  & 4   & Acc. \& F1  \\ \hline
\begin{tabular}[c]{@{}c@{}}Emotion \\ \citep{saravia-etal-2018-carer} \end{tabular} & \begin{tabular}[c]{@{}c@{}} Emotion \\ Detection\end{tabular}  & Twitter            & 9,600   & 2,000 & 2,000  & 6   & Acc. \& F1  \\ \hline
\end{tabular}
}
\end{small}
\end{table}

In Table~\ref{tab:benchmarks_NLP_text}, both Yelp and Amazon dataset are used for the same purpose (sentiment classification). Therefore the selected classes are merged into the same semantic analysis task. \citet{de2019episodic} provided two versions for benchmarking lifelong learning methods: a balanced dataset that contains $115,000$ training and $7,600$ test examples and an imbalanced version that contains $575,000$ training and $38,000$ test examples. The second version is imbalanced due to the different number of examples selected from each source. To study lifelong learning under the realistic scenario of a large number of tasks, \citet{mehta2021empirical} introduces a novel suite of $15$ diverse text classification tasks. They showed that the introduced suite proves more challenging than the previously discussed benchmark with $5$ datasets. Table~\ref{tab:benchmarks_15NLP_text} shows the benchmark datasets in detail. Future works should consider repurposing ExMix \citep{aribandi2022ext}, a massive collection of $107$ supervised tasks across diverse domains to be an even more challenging and realistic lifelong language learning benchmark. To construct a lifelong learning benchmark for the question and answering tasks, \citet{de2019episodic} create their benchmark using the dataset described in Table~\ref{tab:benchmarks_NLP_QA}. 

\small{
\begin{table}[htbp!]
    \centering
    \caption{The datasets that are used in a lifelong learning Benchmark for consecutive question-answering tasks~\citep{de2019episodic}.}
    \label{tab:benchmarks_NLP_QA}
    \resizebox{\textwidth}{!}{\begin{tabular}{@{}lcccc@{}}
\toprule
Dataset   & \multicolumn{1}{l}{Characteristics}                                                               & Source                                                                                                                         & \# Training & \# Test                                                        \\ \midrule
SQuAD 1.1~\citep{rajpurkar2016squad} & Reading comprehension                                                                             & Wikipedia articles                                                                                                             & 90,000      & 10,000                                                         \\ \midrule
TriviaQA~\citep{joshi2017triviaqa}  & \begin{tabular}[c]{@{}c@{}}QA pairs written \\ by trivia enthusiasts \\ and evidence\end{tabular} & Wikipedia                                                                                                                      & 60,000      & 8,000                                                          \\ \midrule
TriviaQA~\citep{joshi2017triviaqa}  & \begin{tabular}[c]{@{}c@{}}QA pairs written \\ by trivia enthusiasts \\ and evidence\end{tabular} & Web                                                                                                                            & 76,000      & \begin{tabular}[c]{@{}c@{}}10,000 \\ (unverified)\end{tabular} \\ \midrule
QuAC~\citep{choi2018quac}      & \begin{tabular}[c]{@{}c@{}}Information-seeking \\ dialog-style\end{tabular}                       & \begin{tabular}[c]{@{}c@{}}Wikipedia article \\ and a teacher answers \\ with a short excerpt \\ from the article\end{tabular} & 80,000      & 7,000                                                          \\ \bottomrule
\end{tabular}}
\end{table}}

\subsection{Word and Sentence Representations}
Word embedding plays a huge role in improving NLP task performance. Some NLP applications may need access to distributed word vectors that can be built sequentially. In a large general-purpose corpus, finding the embedding may not be useful for domain-specific downstream tasks because languages and the meaning of vocabulary can slightly change over time. Emerging social media cause a faster change in the meaning of some vocabularies. For example, the meanings of a word used in specific situations and contexts may differ in some viral videos or posts. As a result, ~\citet{Biesialska_2020} argue that learning word embedding poses an important lifelong learning paradigm. A lifelong learning setup can fulfill the requirements of vocabulary changes over time. According to our best knowledge, there is still a lack of a good benchmark for evaluating lifelong learning methods on learning embedding in a lifelong learning setup.

\subsection{Dialogue Systems}
Lifelong learning in task-oriented dialogue systems has been previously studied by \citet{Lee2017TowardCL}, who perform lifelong learning over three tasks in an end-to-end (E2E) dialogue modeling setting. \citet{wu-etal-2019-transferable} propose a dialogue state tracking (DST) model that trains on a set of domains and can transfer knowledge to a new domain without forgetting. They evaluate DST on different domains in the MultiWOZ dataset \citep{budzianowski2020multiwoz}. \citet{mi2020continual} propose a method for lifelong learning in the natural language generation (NLG) setting, where different domains from the MultiWOZ dataset are presented in a sequence. \citet{geng-etal-2021-continual} also propose a lifelong learning method in the NLG setting that uses network pruning and expansion to adapt to new tasks. They evaluate on a benchmark of 7 tasks, composed of domains from the In-Car Assistant \citep{eric2017keyvalue}, MultiWOZ, and CamRest \citep{wen2017networkbased} datasets. However, these methods look at lifelong learning in task-oriented dialogue with a relatively small number of tasks (3-7), with a smaller amount of data, and in a single setting.

\citet{madotto2020continual} propose a benchmark that presents 37 tasks in a sequence in 4 different settings. They consider single-domain dialogues from 4 different datasets: Taskmaster-1 and Taskmaster-2 \citep{byrne2019taskmaster1}, Schema-Guided Dialogue (SGD) dataset \citep{rastogi2020scalable}, and MultiWOZ \citep{budzianowski2020multiwoz}. Each domain in the datasets forms a single task in the lifelong learning formulation. The four settings they consider are as follows:
\begin{itemize}
    \item Intent prediction (INTENT), where an intent label has to be predicted from the given conversation history.
    \item Dialogue state tracking (DST), where the intent and a sequence of slot-value pairs that are passed to an API call to be tracked have to be predicted from the conversation history.
    \item Natural language generation (NLG), where a natural language response has to be predicted given the result of an API call.
    \item End-to-end (E2E), which combines the three settings mentioned above.
\end{itemize}

\citet{madotto2020continual} unify the settings described above into a sequence-to-sequence learning problem by presenting the dialogue history as a sequence of turns and serializing the API calls.

\begin{table}[htbp!]
\centering
    \caption{The statistics of four tasks that are collected to create a benchmark for Dialog systems in lifelong learning setup~\citep{madotto2020continual}.}
    \label{tab:benchmarks_NLP_dialog}
\resizebox{\columnwidth}{!}{
\begin{tabular}{rcccccc}
\toprule
 & \multicolumn{3}{c}{\textbf{\#Samples}} &  &  &  \\\cmidrule{2-4}
\textbf{Name}  & \textbf{Train} & \textbf{Valid} & \textbf{Test} & \textbf{\#Domains} & \textbf{\#Intents} & \textbf{Average \#turns} \\ \midrule
TM19 & 4,403           & 551            & 553           & 6             & 112           & 19.97          \\
TM20 & 13,839          & 1,731           & 1,734          & 7             & 128           & 16.92          \\
MWoZ & 7,906           & 1,000           & 1,000          & 5             & 15            & 13.93          \\
SGD & 5,278           & 761            & 1,531          & 19            & 43            & 14.71          \\ \midrule
Total                         & 31,426          & 4,043           & 4,818          & 37            & 280           & 16.23          \\ \bottomrule
\end{tabular}
}
\end{table}

\textbf{Metrics on Lifelong Dialogue Systems}: 
To evaluate the dialogue systems, \citet{madotto2020continual} adopt a modular approach for evaluating the models in each setting with a suitable metric as follows: 
\begin{itemize}
    \item \textit{INTENT}: Intent accuracy, which compares the predicted intent label against the gold intent label. 
    \item \textit{DST}: Joint goal accuracy (JGA) metric, which measure the proportion of samples for which all slot-value pairs are predicted correctly. 
    \item \textit{NLG}: Entity error rate (EER) which measures the number of slots in the input that do not appear in the generated utterance, and the BLEU score between the generated and gold utterance. 
    \item \textit{E2E}: All the above metrics.
\end{itemize}

To measure performance in the lifelong learning setting, the average value of the metric over the tasks is used. Forward or backward transfer are measured by aggregating the metrics defined above.

\section{Summary}
To achieve state-of-the-art performance in real-world scenarios, lifelong learning methods must be capable of generalization on a broad range of complex datasets comprising a large number of tasks. The goal of generalization requires lifelong learning benchmarks to be challenging and closer to real-world applications to ensure robust behavior over time. 

We first described numerous vision-based benchmarks involving multiple tasks of image classification with varying levels of difficulty by inducing shifts in the data distribution. We also described NLP domain-based benchmarks. These benchmarks are designed specifically for the type of task at hand such as text classification, question answering, text representation, dialogue systems, etc. 

Some of these benchmarks also required introducing new metrics for evaluations for a more intuitive evaluation of the baselines. It is worth noting that several benchmarks for lifelong learning presented in this chapter are inspired by multi-task learning benchmarks and are often used without any modification.

\chapter{Future Challenges}
\label{sec:future_challenges}

So far, we have looked at both a high-level overview of lifelong learning systems (definition and desiderata of lifelong learning systems and the relationship between lifelong learning and other paradigms) as well as specific instances of lifelong learning algorithms and benchmarks and metrics for evaluating lifelong learning systems. In this last section, we conclude the primer with a discussion on future challenges and important research directions in lifelong learning systems.

The first important research direction is~\textbf{developing methods for organizing (or compartmentalizing) the system's knowledge}. This would enable the system to manipulate specific parts of its knowledge without over-writing all the previous knowledge. Such methods would be useful in the context of catastrophic forgetting and faster adaptation to new tasks as the system can choose which knowledge it wants to forget/update and retain all the other knowledge. These methods are especially relevant in the context of training large-scale models on constantly evolving datasets (like the web data where the ground truth data is continuously changing). Ability to selectively update parts of the knowledge system is a crucial requirement for deploying such systems in the real world~\citep{bommasani2021opportunities}. 

One promising line of work in this direction is the work on~\textit{adapters}~\citep{houlsby2019parameter,pfeiffer2020adapterfusion,pfeiffer2020mad} - small modules that are added to the intermediate layers of a large, pretrained transformer model. When transferring to a downstream task, only the newly added adapter modules are finetuned, and the original, pretrained transformer model is kept fixed. Previous works like~\citet{houlsby2019parameter, pfeiffer2020mad, pfeiffer2020adapterfusion} have demonstrated that finetuning just the adapter modules can also provide excellent performance on the downstream tasks. The use of adapters enables learning downstream models that are \textit{compact} (uses fewer trainable parameters for each downstream task) and \textit{extensible} (uses new adapters that can be incrementally trained on downstream tasks). Adapters can be seen as~\textit{knowledge modules} that can be swapped in and out, depending on the downstream task. However, adapters are designed for the forward transfer of knowledge (start with a pre-trained model and adapt to the new tasks) and do not provide a straightforward mechanism to enable the backward transfer of knowledge (updating the knowledge in the pre-trained model, in a non-catastrophic way, while training on the new tasks). Combining insights from existing work in modular architectures(\cref{sec:architecture}) and adapters could provide a useful inductive bias for compartmentalizing the systems knowledge in terms of its parameters.

Another important research direction is to~\textbf{enable the lifelong learning systems to interface with external knowledge sources}. This capability is especially relevant in the context of interactive systems (like dialogue systems or embodied agents) operating in the real world. These interactive systems are often likely to encounter new topics/objects/items and have to account for the continually changing state of the world. For example, it is not possible for a static language model, trained at one point in time, to answer ``what is the current temperature in Montreal'' without access to an external knowledge base that can provide it the current temperature in Montreal. One way to work around this challenge is to develop lifelong learning systems that learns~\textit{skills} (e.g., querying the external knowledge-bases, aggregating the results from different queries, etc.) instead of knowledge. One popular approach for teaching such~\textit{skills} is to curate a list of tasks, with or without a curriculum, such that each task requires the system to either learn a new skill or learn to combine previously learned skills~\citep{bengio2009curriculum, babyai, weston2015towards}. For example, a question-answering system may start by learning to choose relevant facts from a curated list of facts. The system may learn to create a curated list of facts using a single knowledge base as the next step. In the third task, the system may have to answer the question using the knowledge base (requiring it first to create a list of facts and then answer the question). In the next task, the system could learn to choose the relevant knowledge base(s) to query. In this way, the system learns a sequence of skills and combines them for solving new tasks. The key challenge with this approach is that the list of tasks is often hand-designed, thus limiting the scope, both in terms of the number of tasks and the spectrum of skills they cover.

We note that the system doesn't need to rely exclusively on external knowledge bases for all the information. The system could employ memory-augmented neural networks (like Neural Turning Machine~\citep{graves2014neural, luders2016continual}, Differentiable Neural Computer~\citep{graves2016hybrid} or memory network~\citep{sukhbaatar2015end, sukhbaatar2015end, kumar2016ask, miller2016key,lample2019large}). The key requirement is that the system should be able to query external knowledge stores for accessing the dynamically changing information. This research direction can be seen as an extension to the first direction where the compartmentalized knowledge lives in external, well-curated knowledge bases that the system can access on-demand.

\backmatter

\printbibliography

@inproceedings{li2021parallel,
  title={Parallel Curriculum Experience Replay in Distributed Reinforcement Learning},
  author={Li, Yuyu and Ji, Jianmin},
  booktitle={Proceedings of the 20th International Conference on Autonomous Agents and MultiAgent Systems},
  pages={782--789},
  year={2021}
}

@article{andrychowicz2017hindsight,
  title={Hindsight experience replay},
  author={Andrychowicz, Marcin and Wolski, Filip and Ray, Alex and Schneider, Jonas and Fong, Rachel and Welinder, Peter and McGrew, Bob and Tobin, Josh and Abbeel, Pieter and Zaremba, Wojciech},
  journal={arXiv preprint arXiv:1707.01495},
  year={2017}
}

@inproceedings{rao2019continual,
  title={Continual unsupervised representation learning},
  author={Rao, Dushyant and Visin, Francesco and Rusu, Andrei and Pascanu, Razvan and Teh, Yee Whye and Hadsell, Raia},
  booktitle={Advances in Neural Information Processing Systems},
  pages={7647--7657},
  year={2019}
}

@article{mermillod2013stability,
  title={The stability-plasticity dilemma: Investigating the continuum from catastrophic forgetting to age-limited learning effects},
  author={Mermillod, Martial and Bugaiska, Aur{\'e}lia and Bonin, Patrick},
  journal={Frontiers in psychology},
  volume={4},
  pages={504},
  year={2013},
  publisher={Frontiers}
}

@article{netzer2011reading,
  title={Reading digits in natural images with unsupervised feature learning},
  author={Netzer, Yuval and Wang, Tao and Coates, Adam and Bissacco, Alessandro and Wu, Bo and Ng, Andrew Y},
  year={2011}
}

@techreport{notmnist,
  title={Notmnist dataset},
  author={Bulatov, Yaroslav},
  institution = {Google (Books/OCR)},
  url = {http://yaroslavvb.blogspot.it/2011/09/notmnist-dataset.html},
  year={2011}
}

@article{saha2021gradient,
  title={Gradient projection memory for continual learning},
  author={Saha, Gobinda and Garg, Isha and Roy, Kaushik},
  journal={arXiv preprint arXiv:2103.09762},
  year={2021}
}

@article{van2020brain,
  title={Brain-inspired replay for continual learning with artificial neural networks},
  author={van de Ven, Gido M and Siegelmann, Hava T and Tolias, Andreas S},
  journal={Nature communications},
  volume={11},
  number={1},
  pages={1--14},
  year={2020},
  publisher={Nature Publishing Group}
}

@inproceedings{buzzega2021rethinking,
  title={Rethinking Experience Replay: a Bag of Tricks for Continual Learning},
  author={Buzzega, Pietro and Boschini, Matteo and Porrello, Angelo and Calderara, Simone},
  booktitle={2020 25th International Conference on Pattern Recognition (ICPR)},
  pages={2180--2187},
  year={2021},
  organization={IEEE}
}

@article{mai2020batch,
  title={Batch-level Experience Replay with Review for Continual Learning},
  author={Mai, Zheda and Kim, Hyunwoo and Jeong, Jihwan and Sanner, Scott},
  journal={arXiv preprint arXiv:2007.05683},
  year={2020}
}

@article{gupta2020maml,
  title={La-maml: Look-ahead meta learning for continual learning},
  author={Gupta, Gunshi and Yadav, Karmesh and Paull, Liam},
  journal={arXiv preprint arXiv:2007.13904},
  year={2020}
}

@inproceedings{douillard2021plop,
  title={Plop: Learning without forgetting for continual semantic segmentation},
  author={Douillard, Arthur and Chen, Yifu and Dapogny, Arnaud and Cord, Matthieu},
  booktitle={Proceedings of the IEEE/CVF Conference on Computer Vision and Pattern Recognition},
  pages={4040--4050},
  year={2021}
}

@article{zhong2021discriminative,
  title={Discriminative Distillation to Reduce Class Confusion in Continual Learning},
  author={Zhong, Changhong and Cui, Zhiying and Wang, Ruixuan and Zheng, Wei-Shi},
  journal={arXiv preprint arXiv:2108.05187},
  year={2021}
}

@inproceedings{fini2020online,
  title={Online continual learning under extreme memory constraints},
  author={Fini, Enrico and Lathuiliere, St{\'e}phane and Sangineto, Enver and Nabi, Moin and Ricci, Elisa},
  booktitle={European Conference on Computer Vision},
  pages={720--735},
  year={2020},
  organization={Springer}
}

@inproceedings{kumar2021bayesian,
  title={Bayesian structural adaptation for continual learning},
  author={Kumar, Abhishek and Chatterjee, Sunabha and Rai, Piyush},
  booktitle={International Conference on Machine Learning},
  pages={5850--5860},
  year={2021},
  organization={PMLR}
}

@article{ebrahimi2019uncertainty,
  title={Uncertainty-guided continual learning with bayesian neural networks},
  author={Ebrahimi, Sayna and Elhoseiny, Mohamed and Darrell, Trevor and Rohrbach, Marcus},
  journal={arXiv preprint arXiv:1906.02425},
  year={2019}
}

@article{ahn2019uncertainty,
  title={Uncertainty-based continual learning with adaptive regularization},
  author={Ahn, Hongjoon and Cha, Sungmin and Lee, Donggyu and Moon, Taesup},
  journal={arXiv preprint arXiv:1905.11614},
  year={2019}
}

@article{jung2020continual,
  title={Continual learning with node-importance based adaptive group sparse regularization},
  author={Jung, Sangwon and Ahn, Hongjoon and Cha, Sungmin and Moon, Taesup},
  journal={arXiv preprint arXiv:2003.13726},
  year={2020}
}

@inproceedings{ozgun2020importance,
  title={Importance driven continual learning for segmentation across domains},
  author={{\"O}zg{\"u}n, Sinan and Rickmann, Anne-Marie and Roy, Abhijit Guha and Wachinger, Christian},
  booktitle={International Workshop on Machine Learning in Medical Imaging},
  pages={423--433},
  year={2020},
  organization={Springer}
}

@article{gou2021knowledge,
  title={Knowledge distillation: A survey},
  author={Gou, Jianping and Yu, Baosheng and Maybank, Stephen J and Tao, Dacheng},
  journal={International Journal of Computer Vision},
  volume={129},
  number={6},
  pages={1789--1819},
  year={2021},
  publisher={Springer}
}

@article{pless2009survey,
  title={A survey of manifold learning for images},
  author={Pless, Robert and Souvenir, Richard},
  journal={IPSJ Transactions on Computer Vision and Applications},
  volume={1},
  pages={83--94},
  year={2009},
  publisher={Information Processing Society of Japan}
}

@article{friston2007variational,
  title={Variational free energy and the Laplace approximation},
  author={Friston, Karl and Mattout, J{\'e}r{\'e}mie and Trujillo-Barreto, Nelson and Ashburner, John and Penny, Will},
  journal={Neuroimage},
  volume={34},
  number={1},
  pages={220--234},
  year={2007},
  publisher={Elsevier}
}

@article{tokdar2010importance,
  title={Importance sampling: a review},
  author={Tokdar, Surya T and Kass, Robert E},
  journal={Wiley Interdisciplinary Reviews: Computational Statistics},
  volume={2},
  number={1},
  pages={54--60},
  year={2010},
  publisher={Wiley Online Library}
}

@inproceedings{hershey2007variational,
  title={Variational Kullback-Leibler divergence for hidden Markov models},
  author={Hershey, John R and Olsen, Peder A and Rennie, Steven J},
  booktitle={2007 IEEE Workshop on Automatic Speech Recognition \& Understanding (ASRU)},
  pages={323--328},
  year={2007},
  organization={IEEE}
}

@inproceedings{li2015generative,
  title={Generative moment matching networks},
  author={Li, Yujia and Swersky, Kevin and Zemel, Rich},
  booktitle={International Conference on Machine Learning},
  pages={1718--1727},
  year={2015},
  organization={PMLR}
}

@inproceedings{selvaraju2017grad,
  title={Grad-cam: Visual explanations from deep networks via gradient-based localization},
  author={Selvaraju, Ramprasaath R and Cogswell, Michael and Das, Abhishek and Vedantam, Ramakrishna and Parikh, Devi and Batra, Dhruv},
  booktitle={Proceedings of the IEEE international conference on computer vision},
  pages={618--626},
  year={2017}
}

@article{hinton2015distilling,
  title={Distilling the knowledge in a neural network},
  author={Hinton, Geoffrey and Vinyals, Oriol and Dean, Jeff},
  journal={arXiv preprint arXiv:1503.02531},
  year={2015}
}

@article{khetarpal2020towards,
  title={Towards continual reinforcement learning: A review and perspectives},
  author={Khetarpal, Khimya and Riemer, Matthew and Rish, Irina and Precup, Doina},
  journal={arXiv preprint arXiv:2012.13490},
  year={2020}
}

@inproceedings{mirzadeh2020dropout,
  title={Dropout as an implicit gating mechanism for continual learning},
  author={Mirzadeh, Seyed Iman and Farajtabar, Mehrdad and Ghasemzadeh, Hassan},
  booktitle={Proceedings of the IEEE/CVF Conference on Computer Vision and Pattern Recognition Workshops},
  pages={232--233},
  year={2020}
}

@article{french1999catastrophic,
  title={Catastrophic forgetting in connectionist networks},
  author={French, Robert M},
  journal={Trends in cognitive sciences},
  volume={3},
  number={4},
  pages={128--135},
  year={1999},
  publisher={Elsevier}
}

@inproceedings{purushwalkam2019task,
  title={Task-driven modular networks for zero-shot compositional learning},
  author={Purushwalkam, Senthil and Nickel, Maximilian and Gupta, Abhinav and Ranzato, Marc'Aurelio},
  booktitle={Proceedings of the IEEE/CVF International Conference on Computer Vision},
  pages={3593--3602},
  year={2019}
}

@inproceedings{ritter2018online,
 author = {Ritter, Hippolyt and Botev, Aleksandar and Barber, David},
 booktitle = {Advances in Neural Information Processing Systems},
 editor = {S. Bengio and H. Wallach and H. Larochelle and K. Grauman and N. Cesa-Bianchi and R. Garnett},
 pages = {},
 publisher = {Curran Associates, Inc.},
 title = {Online Structured Laplace Approximations for Overcoming Catastrophic Forgetting},
 url = {https://proceedings.neurips.cc/paper/2018/file/f31b20466ae89669f9741e047487eb37-Paper.pdf},
 volume = {31},
 year = {2018}
}

@article{mackay1992practical, author = {MacKay, David J. C.}, title = {A Practical Bayesian Framework for Backpropagation Networks}, year = {1992}, issue_date = {May 1992}, publisher = {MIT Press}, address = {Cambridge, MA, USA}, volume = {4}, number = {3}, issn = {0899-7667}, url = {https://doi.org/10.1162/neco.1992.4.3.448}, doi = {10.1162/neco.1992.4.3.448}, journal = {Neural Comput.}, month = may, pages = {448–472}, numpages = {25} }

@misc{huszar2017quadratic,
      title={On Quadratic Penalties in Elastic Weight Consolidation}, 
      author={Ferenc Huszár},
      year={2017},
      eprint={1712.03847},
      archivePrefix={arXiv},
      primaryClass={stat.ML}
}

@inproceedings{dhar2019learning,
  title={Learning without memorizing},
  author={Dhar, Prithviraj and Singh, Rajat Vikram and Peng, Kuan-Chuan and Wu, Ziyan and Chellappa, Rama},
  booktitle={Proceedings of the IEEE Conference on Computer Vision and Pattern Recognition},
  pages={5138--5146},
  year={2019}
}

@misc{benzing2021unifying,
      title={Unifying Regularisation Methods for Continual Learning}, 
      author={Frederik Benzing},
      year={2021},
      eprint={2006.06357},
      archivePrefix={arXiv},
      primaryClass={cs.LG}
}

@inproceedings{lopez2017gradient,
  title={Gradient episodic memory for continual learning},
  author={Lopez-Paz, David and Ranzato, Marc'Aurelio},
  booktitle={Advances in neural information processing systems},
  pages={6467--6476},
  year={2017}
}

@inproceedings{chaudhry2019efficient,
  title={Efficient Lifelong Learning with A-GEM},
  author={Chaudhry, Arslan and Ranzato, Marc’Aurelio and Rohrbach, Marcus and Elhoseiny, Mohamed},
  booktitle={International Conference on Learning Representations},
  year={2019}
}

@article{chaudhry2019tiny,
  title={On tiny episodic memories in continual learning},
  author={Chaudhry, Arslan and Rohrbach, Marcus and Elhoseiny, Mohamed and Ajanthan, Thalaiyasingam and Dokania, Puneet K and Torr, Philip HS and Ranzato, Marc'Aurelio},
  journal={arXiv preprint arXiv:1902.10486},
  year={2019}
}

@article{guo2020improved,
  title={Improved Schemes for Episodic Memory-based Lifelong Learning},
  author={Guo, Yunhui and Liu, Mingrui and Yang, Tianbao and Rosing, Tajana},
  journal={Advances in Neural Information Processing Systems},
  volume={33},
  year={2020}
}

@inproceedings{farajtabar2020orthogonal,
  title={Orthogonal gradient descent for continual learning},
  author={Farajtabar, Mehrdad and Azizan, Navid and Mott, Alex and Li, Ang},
  booktitle={International Conference on Artificial Intelligence and Statistics},
  pages={3762--3773},
  year={2020},
  organization={PMLR}
}

@inproceedings{shin2017continual,
  title={Continual learning with deep generative replay},
  author={Shin, Hanul and Lee, Jung Kwon and Kim, Jaehong and Kim, Jiwon},
  booktitle={Advances in Neural Information Processing Systems},
  pages={2990--2999},
  year={2017}
}

@inproceedings{
sprechmann2018memorybased,
title={Memory-based Parameter Adaptation},
author={Pablo Sprechmann and Siddhant Jayakumar and Jack Rae and Alexander Pritzel and Adria Puigdomenech Badia and Benigno Uria and Oriol Vinyals and Demis Hassabis and Razvan Pascanu and Charles Blundell},
booktitle={International Conference on Learning Representations},
year={2018},
url={https://openreview.net/forum?id=rkfOvGbCW},
}

@inproceedings{de2019episodic,
  title={Episodic Memory in Lifelong Language Learning},
  author={de Masson d'Autume, Cyprien and Ruder, Sebastian and Kong, Lingpeng and Yogatama, Dani},
  booktitle={Advances in Neural Information Processing Systems},
  pages={13143--13152},
  year={2019}
}

@inproceedings{wang2020efficientML,
  title={Efficient Meta Lifelong-Learning with Limited Memory},
  author={Wang, Zirui and Mehta, Sanket Vaibhav and Poczos, Barnabas and Carbonell, Jaime G},
  booktitle={Proceedings of the 2020 Conference on Empirical Methods in Natural Language Processing (EMNLP)},
  pages={535--548},
  year={2020}
}

@inproceedings{sun2020lamol,
  title={LAMOL: LAnguage MOdeling for Lifelong Language Learning},
  author={Sun, Fan-Keng and Ho, Cheng-Hao and Lee, Hung-Yi},
  booktitle={International Conference on Learning Representations},
  year={2020}
}

@inproceedings{aljundi2019gradient,
  title={Gradient based sample selection for online continual learning},
  author={Aljundi, Rahaf and Lin, Min and Goujaud, Baptiste and Bengio, Yoshua},
  booktitle={Advances in Neural Information Processing Systems},
  pages={11816--11825},
  year={2019}
}

@inproceedings{riemer2019learning,
  title={Learning to Learn without Forgetting by Maximizing Transfer and Minimizing Interference},
  author={Riemer, Matthew and Cases, Ignacio and Ajemian, Robert and Liu, Miao and Rish, Irina and Tu, Yuhai and Tesauro, Gerald},
  booktitle={International Conference on Learning Representations},
  year={2019}
}

@inproceedings{isele2018selective,
  title={Selective experience replay for lifelong learning},
  author={Isele, David and Cosgun, Akansel},
  booktitle={AAAI Conference on Artificial Intelligence 2018},
  pages={3302--3309},
  year={2018},
  organization={Association for the Advancement of Artificial Intelligence (AAAI)}
}

@inproceedings{chaudhry2021using,
  title={Using Hindsight to Anchor Past Knowledge in Continual Learning},
  author={Chaudhry, Arslan and Gordo, Albert and Dokania, Puneet and Torr, Philip and Lopez-Paz, David},
  booktitle={Proceedings of the AAAI Conference on Artificial Intelligence},
  volume={35},
  pages={6993--7001},
  year={2021}
}

@article{ramirez2013creating,
  title={Creating a false memory in the hippocampus},
  author={Ramirez, Steve and Liu, Xu and Lin, Pei-Ann and Suh, Junghyup and Pignatelli, Michele and Redondo, Roger L and Ryan, Tom{\'a}s J and Tonegawa, Susumu},
  journal={Science},
  volume={341},
  number={6144},
  pages={387--391},
  year={2013},
  publisher={American Association for the Advancement of Science}
}

@article{o2002hippocampal,
  title={Hippocampal and neocortical contributions to memory: Advances in the complementary learning systems framework},
  author={O'Reilly, Randall C and Norman, Kenneth A},
  journal={Trends in cognitive sciences},
  volume={6},
  number={12},
  pages={505--510},
  year={2002},
  publisher={Elsevier}
}

@article{stickgold2007sleep,
  title={Sleep-dependent memory consolidation and reconsolidation},
  author={Stickgold, Robert and Walker, Matthew P},
  journal={Sleep medicine},
  volume={8},
  number={4},
  pages={331--343},
  year={2007},
  publisher={Elsevier}
}

@article{mccann2018natural,
  title={The natural language decathlon: Multitask learning as question answering},
  author={McCann, Bryan and Keskar, Nitish Shirish and Xiong, Caiming and Socher, Richard},
  journal={arXiv preprint arXiv:1806.08730},
  year={2018}
}

@inproceedings{
ramalho2018adaptive,
title={Adaptive Posterior Learning: few-shot learning with a surprise-based memory module},
author={Tiago Ramalho and Marta Garnelo},
booktitle={International Conference on Learning Representations},
year={2019},
url={https://openreview.net/forum?id=ByeSdsC9Km},
}

@inproceedings{
toneva2018an,
title={An Empirical Study of Example Forgetting during Deep Neural Network Learning},
author={Mariya Toneva and Alessandro Sordoni and Remi Tachet des Combes and Adam Trischler and Yoshua Bengio and Geoffrey J. Gordon},
booktitle={International Conference on Learning Representations},
year={2019},
url={https://openreview.net/forum?id=BJlxm30cKm},
}

@article{gupta2010hippocampal,
  title={Hippocampal Replay Is Not a Simple Function of Experience},
  author={Gupta, Anoopum S and van der Meer, Matthijs AA and Touretzky, David S and Redish, A David},
  journal={Neuron},
  volume={65},
  number={5},
  pages={695--705},
  year={2010},
  publisher={Elsevier}
}

@article{cheng2008new,
  title={New Experiences Enhance Coordinated Neural Activity in the Hippocampus},
  author={Cheng, Sen and Frank, Loren M},
  journal={Neuron},
  volume={57},
  number={2},
  pages={303--313},
  year={2008},
  publisher={Elsevier}
}

@article{mcnamara2014dopaminergic,
  title={Dopaminergic neurons promote hippocampal reactivation and spatial memory persistence},
  author={McNamara, Colin G and Tejero-Cantero, {\'A}lvaro and Trouche, St{\'e}phanie and Campo-Urriza, Natalia and Dupret, David},
  journal={Nature neuroscience},
  volume={17},
  number={12},
  pages={1658--1660},
  year={2014},
  publisher={Nature Publishing Group}
}

@article{atherton2015memory,
  title={Memory trace replay: the shaping of memory consolidation by neuromodulation.},
  author={Atherton, LA and Dupret, D and Mellor, JR},
  journal={Trends in Neurosciences},
  volume={38},
  number={9},
  pages={560--570},
  year={2015}
}

@article{olafsdottir2015hippocampal,
  title={Hippocampal place cells construct reward related sequences through unexplored space},
  author={{\'O}lafsd{\'o}ttir, H Freyja and Barry, Caswell and Saleem, Aman B and Hassabis, Demis and Spiers, Hugo J},
  journal={Elife},
  volume={4},
  pages={e06063},
  year={2015},
  publisher={eLife Sciences Publications Limited}
}

@article{vitter1985random,
  title={Random sampling with a reservoir},
  author={Vitter, Jeffrey S},
  journal={ACM Transactions on Mathematical Software (TOMS)},
  volume={11},
  number={1},
  pages={37--57},
  year={1985},
  publisher={ACM New York, NY, USA}
}

@inproceedings{bickel2008transfer,
  title={Transfer learning by distribution matching for targeted advertising},
  author={Bickel, Steffen and Sawade, Christoph and Scheffer, Tobias},
  booktitle={Proceedings of the 21st International Conference on Neural Information Processing Systems},
  pages={145--152},
  year={2008}
}

@inproceedings{welling2009herding,
  title={Herding dynamical weights to learn},
  author={Welling, Max},
  booktitle={Proceedings of the 26th Annual International Conference on Machine Learning},
  pages={1121--1128},
  year={2009}
}

@article{graves2014neural,
  title={Neural turing machines},
  author={Graves, Alex and Wayne, Greg and Danihelka, Ivo},
  journal={arXiv preprint arXiv:1410.5401},
  year={2014}
}

@article{tulving1985many,
  title={How many memory systems are there?},
  author={Tulving, Endel},
  journal={American psychologist},
  volume={40},
  number={4},
  pages={385},
  year={1985},
  publisher={American Psychological Association}
}

@article{mcclelland1995there,
  title={Why there are complementary learning systems in the hippocampus and neocortex: insights from the successes and failures of connectionist models of learning and memory.},
  author={McClelland, James L and McNaughton, Bruce L and O'Reilly, Randall C},
  journal={Psychological review},
  volume={102},
  number={3},
  pages={419},
  year={1995},
  publisher={American Psychological Association}
}

@article{mitchell2018never,
  title={Never-ending learning},
  author={Mitchell, Tom and Cohen, William and Hruschka, Estevam and Talukdar, Partha and Yang, Bishan and Betteridge, Justin and Carlson, Andrew and Dalvi, Bhanava and Gardner, Matt and Kisiel, Bryan and others},
  journal={Communications of the ACM},
  volume={61},
  number={5},
  pages={103--115},
  year={2018},
  publisher={ACM New York, NY, USA}
}

@inproceedings{chen2015lifelong,
  title={Lifelong Learning for Sentiment Classification},
  author={Chen, Zhiyuan and Ma, Nianzu and Liu, Bing},
  booktitle={Proceedings of the 53rd Annual Meeting of the Association for Computational Linguistics and the 7th International Joint Conference on Natural Language Processing (Volume 2: Short Papers)},
  pages={750--756},
  year={2015}
}

@inproceedings{schwarz2018progress,
  title={Progress and compress: A scalable framework for continual learning},
  author={Schwarz, Jonathan and Czarnecki, Wojciech and Luketina, Jelena and Grabska-Barwinska, Agnieszka and Teh, Yee Whye and Pascanu, Razvan and Hadsell, Raia},
  booktitle={International Conference on Machine Learning},
  pages={4528--4537},
  year={2018},
  organization={PMLR}
}

@article{cl_dnn_hadsell,
author = "Hadsell, Raia and Rao, Dushyant and Rusu, Andrei A. and Pascanu, Razvan",
title =  "Embracing Change: Continual Learning in Deep Neural Networks",
journal = "Trends in Cognitive Sciences",
year = 2020}

@InProceedings{pmlr-v70-zenke17a,
  title = 	 {Continual Learning Through Synaptic Intelligence},
  author =       {Friedemann Zenke and Ben Poole and Surya Ganguli},
  booktitle = 	 {Proceedings of the 34th International Conference on Machine Learning},
  pages = 	 {3987--3995},
  year = 	 {2017},
  editor = 	 {Doina Precup and Yee Whye Teh},
  volume = 	 {70},
  series = 	 {Proceedings of Machine Learning Research},
  address = 	 {International Convention Centre, Sydney, Australia},
  month = 	 {06--11 Aug},
  publisher =    {PMLR},
  url = 	 {http://proceedings.mlr.press/v70/zenke17a.html},
}

@inproceedings{
yoon2018lifelong,
title={Lifelong Learning with Dynamically Expandable Networks},
author={Jaehong Yoon and Eunho Yang and Jeongtae Lee and Sung Ju Hwang},
booktitle={International Conference on Learning Representations},
year={2018},
url={https://openreview.net/forum?id=Sk7KsfW0-},
}

@article{zhang2020class,
  title={Class-incremental learning via deep model consolidation},
  author={Junting Zhang and Jie Zhang and Shalini Ghosh and Dawei Li and Serafettin Tasci and Larry Heck and Heming Zhang and C.-C. Jay Kuo},
  journal={},
  year={}
}

@article{ewc,
  title={Overcoming catastrophic forgetting in neural networks},
  author={Kirkpatrick, James and Pascanu, Razvan and Rabinowitz, Neil and Veness, Joel and Desjardins, Guillaume and Rusu, Andrei A and Milan, Kieran and Quan, John and Ramalho, Tiago and Grabska-Barwinska, Agnieszka and others},
  journal={Proceedings of the national academy of sciences},
  volume={114},
  number={13},
  pages={3521--3526},
  year={2017},
  publisher={National Acad Sciences}
}

@inproceedings{mas,
  title={Memory aware synapses: Learning what (not) to forget},
  author={Aljundi, Rahaf and Babiloni, Francesca and Elhoseiny, Mohamed and Rohrbach, Marcus and Tuytelaars, Tinne},
  booktitle={Proceedings of the European Conference on Computer Vision (ECCV)},
  pages={139--154},
  year={2018}
}

@inproceedings{chaudhry2018riemannian,
  title={Riemannian walk for incremental learning: Understanding forgetting and intransigence},
  author={Chaudhry, Arslan and Dokania, Puneet K and Ajanthan, Thalaiyasingam and Torr, Philip HS},
  booktitle={Proceedings of the European Conference on Computer Vision (ECCV)},
  pages={532--547},
  year={2018}
}

@article{graves2016hybrid,
  author = {Graves, Alex and Wayne, Greg and Reynolds, Malcolm and Harley, Tim and Danihelka, Ivo and Grabska-Barwińska, Agnieszka and Colmenarejo, Sergio Gómez and Grefenstette, Edward and Ramalho, Tiago and Agapiou, John and Badia, Adrià Puigdomènech and Hermann, Karl Moritz and Zwols, Yori and Ostrovski, Georg and Cain, Adam and King, Helen and Summerfield, Christopher and Blunsom, Phil and Kavukcuoglu, Koray and Hassabis, Demis},
  description = {Hybrid computing using a neural network with dynamic external memory : Nature : Nature Research},
  issn = {00280836},
  journal = {Nature},
  month = oct,
  number = 7626,
  pages = {471--476},
  publisher = {Macmillan Publishers Limited, part of Springer Nature. All rights reserved.},
  title = {Hybrid computing using a neural network with dynamic external memory},
  url = {http://dx.doi.org/10.1038/nature20101},
  volume = 538,
  year = 2016
}

@inproceedings{
tang2021graphbased,
title={Graph-Based Continual Learning},
author={Binh Tang and David S. Matteson},
booktitle={International Conference on Learning Representations},
year={2021},
url={https://openreview.net/forum?id=HHSEKOnPvaO}
}

@misc{mirzadeh2020understanding,
      title={Understanding the Role of Training Regimes in Continual Learning}, 
      author={Seyed Iman Mirzadeh and Mehrdad Farajtabar and Razvan Pascanu and Hassan Ghasemzadeh},
      year={2020},
      eprint={2006.06958},
      archivePrefix={arXiv},
      primaryClass={cs.LG}
}

@article{mehta2021empirical,
  title={An empirical investigation of the role of pre-training in lifelong learning},
  author={Mehta, Sanket Vaibhav and Patil, Darshan and Chandar, Sarath and Strubell, Emma},
  journal={arXiv preprint arXiv:2112.09153},
  year={2021}
}

@inproceedings{foret2020sharpness,
  title={Sharpness-aware Minimization for Efficiently Improving Generalization},
  author={Foret, Pierre and Kleiner, Ariel and Mobahi, Hossein and Neyshabur, Behnam},
  booktitle={International Conference on Learning Representations},
  year={2021}
}

@article{hochreiter1997flat,
  title={Flat minima},
  author={Hochreiter, Sepp and Schmidhuber, J{\"u}rgen},
  journal={Neural computation},
  volume={9},
  number={1},
  pages={1--42},
  year={1997},
  publisher={MIT Press One Rogers Street, Cambridge, MA 02142-1209, USA journals-info~…}
}

@inproceedings{xie2020diffusion,
  title={A Diffusion Theory For Deep Learning Dynamics: Stochastic Gradient Descent Exponentially Favors Flat Minima},
  author={Xie, Zeke and Sato, Issei and Sugiyama, Masashi},
  booktitle={International Conference on Learning Representations},
  year={2021}
}

@inproceedings{jastrzebski2019break,
  title={The Break-Even Point on Optimization Trajectories of Deep Neural Networks},
  author={Jastrzebski, Stanislaw and Szymczak, Maciej and Fort, Stanislav and Arpit, Devansh and Tabor, Jacek and Cho, Kyunghyun and Geras, Krzysztof},
  booktitle={International Conference on Learning Representations},
  year={2020}
}

@article{warstadt2019neural,
  title={Neural network acceptability judgments},
  author={Warstadt, Alex and Singh, Amanpreet and Bowman, Samuel R},
  journal={Transactions of the Association for Computational Linguistics},
  volume={7},
  pages={625--641},
  year={2019},
  publisher={MIT Press}
}

@inproceedings{clark2019boolq,
  title={BoolQ: Exploring the Surprising Difficulty of Natural Yes/No Questions},
  author={Clark, Christopher and Lee, Kenton and Chang, Ming-Wei and Kwiatkowski, Tom and Collins, Michael and Toutanova, Kristina},
  booktitle={Proceedings of the 2019 Conference of the North American Chapter of the Association for Computational Linguistics: Human Language Technologies, Volume 1 (Long and Short Papers)},
  pages={2924--2936},
  year={2019}
}

@inproceedings{socher2013recursive,
  title={Recursive deep models for semantic compositionality over a sentiment treebank},
  author={Socher, Richard and Perelygin, Alex and Wu, Jean and Chuang, Jason and Manning, Christopher D and Ng, Andrew Y and Potts, Christopher},
  booktitle={Proceedings of the 2013 conference on empirical methods in natural language processing},
  pages={1631--1642},
  year={2013}
}

@inproceedings{williams2018broad,
  title={A Broad-Coverage Challenge Corpus for Sentence Understanding through Inference},
  author={Williams, Adina and Nangia, Nikita and Bowman, Samuel},
  booktitle={Proceedings of the 2018 Conference of the North American Chapter of the Association for Computational Linguistics: Human Language Technologies, Volume 1 (Long Papers)},
  pages={1112--1122},
  year={2018}
}

@article{wang2018glue,
  title={GLUE: A multi-task benchmark and analysis platform for natural language understanding},
  author={Wang, Alex and Singh, Amanpreet and Michael, Julian and Hill, Felix and Levy, Omer and Bowman, Samuel R},
  journal={arXiv preprint arXiv:1804.07461},
  year={2018}
}

@inproceedings{poliak2018collecting,
  title={Collecting Diverse Natural Language Inference Problems for Sentence Representation Evaluation},
  author={Poliak, Adam and Haldar, Aparajita and Rudinger, Rachel and Hu, J Edward and Pavlick, Ellie and White, Aaron Steven and Van Durme, Benjamin},
  booktitle={Proceedings of the 2018 Conference on Empirical Methods in Natural Language Processing},
  pages={67--81},
  year={2018}
}

@inproceedings{stab2018cross,
  title={Cross-topic Argument Mining from Heterogeneous Sources},
  author={Stab, Christian and Miller, Tristan and Schiller, Benjamin and Rai, Pranav and Gurevych, Iryna},
  booktitle={Proceedings of the 2018 Conference on Empirical Methods in Natural Language Processing},
  pages={3664--3674},
  year={2018}
}

@inproceedings{khot2018scitail,
  title={SciTaiL: A Textual Entailment Dataset from Science Question Answering},
  author={Khot, Tushar and Sabharwal, Ashish and Clark, Peter},
  booktitle={Thirty-Second AAAI Conference on Artificial Intelligence},
  year={2018}
}

@inproceedings{mostafazadeh-etal-2016-corpus,
    title = "A Corpus and Cloze Evaluation for Deeper Understanding of Commonsense Stories",
    author = "Mostafazadeh, Nasrin  and
      Chambers, Nathanael  and
      He, Xiaodong  and
      Parikh, Devi  and
      Batra, Dhruv  and
      Vanderwende, Lucy  and
      Kohli, Pushmeet  and
      Allen, James",
    booktitle = "Proceedings of the 2016 Conference of the North {A}merican Chapter of the Association for Computational Linguistics: Human Language Technologies",
    month = jun,
    year = "2016",
    address = "San Diego, California",
    publisher = "Association for Computational Linguistics",
    url = "https://www.aclweb.org/anthology/N16-1098",
    doi = "10.18653/v1/N16-1098",
    pages = "839--849",
}

@inproceedings{saravia-etal-2018-carer,
    title = "{CARER}: Contextualized Affect Representations for Emotion Recognition",
    author = "Saravia, Elvis  and
      Liu, Hsien-Chi Toby  and
      Huang, Yen-Hao  and
      Wu, Junlin  and
      Chen, Yi-Shin",
    booktitle = "Proceedings of the 2018 Conference on Empirical Methods in Natural Language Processing",
    month = oct # "-" # nov,
    year = "2018",
    address = "Brussels, Belgium",
    publisher = "Association for Computational Linguistics",
    url = "https://www.aclweb.org/anthology/D18-1404",
    doi = "10.18653/v1/D18-1404",
    pages = "3687--3697",
}

@inproceedings{zhang2015character,
  title={Character-level convolutional networks for text classification},
  author={Zhang, Xiang and Zhao, Junbo and LeCun, Yann},
  booktitle={Proceedings of the 28th International Conference on Neural Information Processing Systems-Volume 1},
  pages={649--657},
  year={2015}
}

@inproceedings{prasad2019penn,
  title={Penn Discourse Treebank Version 3.0},
  author={Prasad, Rashmi and Webber, Bonnie and Lee, Alan and Joshi, Aravind},
  booktitle={LDC2019T05},
  publisher={Philadelphia: Linguistic Data Consortium.},
  year={2019}
}

@inproceedings{kim2020implicit,
  title={Implicit discourse relation classification: We need to talk about evaluation},
  author={Kim, Najoung and Feng, Song and Gunasekara, Chulaka and Lastras, Luis},
  booktitle={Proceedings of the 58th Annual Meeting of the Association for Computational Linguistics},
  pages={5404--5414},
  year={2020}
}

@inproceedings{
aribandi2022ext,
title={ExT5: Towards Extreme Multi-Task Scaling for Transfer Learning},
author={Vamsi Aribandi and Yi Tay and Tal Schuster and Jinfeng Rao and Huaixiu Steven Zheng and Sanket Vaibhav Mehta and Honglei Zhuang and Vinh Q. Tran and Dara Bahri and Jianmo Ni and Jai Gupta and Kai Hui and Sebastian Ruder and Donald Metzler},
booktitle={International Conference on Learning Representations},
year={2022},
url={https://openreview.net/forum?id=Vzh1BFUCiIX}
}

@inproceedings{
Goyal2020Reinforcement,
title={Reinforcement Learning with Competitive  Ensembles of Information-Constrained Primitives},
author={Anirudh Goyal and Shagun Sodhani and Jonathan Binas and Xue Bin Peng and Sergey Levine and Yoshua Bengio},
booktitle={International Conference on Learning Representations},
year={2020},
url={https://openreview.net/forum?id=ryxgJTEYDr}
}

@InProceedings{pmlr-v80-parascandolo18a,
  title = 	 {Learning Independent Causal Mechanisms},
  author =       {Parascandolo, Giambattista and Kilbertus, Niki and Rojas-Carulla, Mateo and Sch{\"o}lkopf, Bernhard},
  booktitle = 	 {Proceedings of the 35th International Conference on Machine Learning},
  pages = 	 {4036--4044},
  year = 	 {2018},
  editor = 	 {Jennifer Dy and Andreas Krause},
  volume = 	 {80},
  series = 	 {Proceedings of Machine Learning Research},
  address = 	 {Stockholmsmässan, Stockholm Sweden},
  month = 	 {10--15 Jul},
  publisher =    {PMLR},
  url = 	 {http://proceedings.mlr.press/v80/parascandolo18a.html}
}

@preamble{"\newcommand{\SortNoop}[1]{}"}

@inproceedings{andreas2016-neural-module-networks,
  title={Neural module networks},
  author={Andreas, Jacob and Rohrbach, Marcus and Darrell, Trevor and Klein, Dan},
  booktitle={Proceedings of the IEEE Conference on Computer Vision and Pattern Recognition},
  pages={39--48},
  year={2016}
}

@inproceedings{johnson2017-clevr-a-diagnostic-dataset-for-compositional-language-and-elementary-visual-reasoning,
  title={Clevr: A diagnostic dataset for compositional language and elementary visual reasoning},
  author={Johnson, Justin and Hariharan, Bharath and van der Maaten, Laurens and Fei-Fei, Li and Lawrence Zitnick, C and Girshick, Ross},
  booktitle={Proceedings of the IEEE Conference on Computer Vision and Pattern Recognition},
  pages={2901--2910},
  year={2017}
}

@inproceedings{Krizhevsky2012-imagenet-classification-with-deep-convolutional-neural-networks,
  title={Imagenet classification with deep convolutional neural networks},
  author={Krizhevsky, Alex and Sutskever, Ilya and Hinton, Geoff},
  booktitle={Advances in Neural Information Processing Systems 25},
  pages={1106--1114},
  year={2012}
}

@article{f,
  title={Universal grammar},
  author={Montague, Richard},
  journal={Theoria},
  volume={36},
  number={3},
  pages={373--398},
  year={1970},
  publisher={Wiley Online Library}
}

@inproceedings{santoro2017simple-neural-network-module-for-relational-reasoning,
  title={A simple neural network module for relational reasoning},
  author={Santoro, Adam and Raposo, David and Barrett, David G and Malinowski, Mateusz and Pascanu, Razvan and Battaglia, Peter and Lillicrap, Timothy},
  booktitle={Advances in neural information processing systems},
  pages={4967--4976},
  year={2017}
}

@article{happel1994design-and-evolution-of-modular-neural-network,
  title={Design and evolution of modular neural network architectures},
  author={Happel, Bart LM and Murre, Jacob MJ},
  journal={Neural networks},
  volume={7},
  number={6-7},
  pages={985--1004},
  year={1994},
  publisher={Elsevier}
}

@inproceedings{yu2018mattnet-modula-attention-network-for-referring-expression-comprehension,
  title={Mattnet: Modular attention network for referring expression comprehension},
  author={Yu, Licheng and Lin, Zhe and Shen, Xiaohui and Yang, Jimei and Lu, Xin and Bansal, Mohit and Berg, Tamara L},
  booktitle={Proceedings of the IEEE Conference on Computer Vision and Pattern Recognition},
  pages={1307--1315},
  year={2018}
}

@article{andreas2016learning-to-compose-neural-networks-for-question-answering,
  title={Learning to compose neural networks for question answering},
  author={Andreas, Jacob and Rohrbach, Marcus and Darrell, Trevor and Klein, Dan},
  journal={arXiv preprint arXiv:1601.01705},
  year={2016}
}

@article{sharkey1997modularity-combining-and-artificial-neural-networks,
  title={Modularity, combining and artificial neural nets},
  author={SHARKEY, AMANDA J C},
  journal={Connection Science},
  volume={9},
  number={1},
  pages={3--10},
  year={1997},
  publisher={Taylor \& Francis}
}

@article{alet2018modular-meta-learning-in-abstract-graph-neural-networks-for-combanitorial-reasoning,
  title={Modular meta-learning in abstract graph networks for combinatorial generalization},
  author={Alet, Ferran and Bauza, Maria and Rodriguez, Alberto and Lozano-Perez, Tomas and Kaelbling, Leslie P},
  journal={arXiv preprint arXiv:1812.07768},
  year={2018}
}

@article{alet2018modular-meta-learning,
  title={Modular meta-learning},
  author={Alet, Ferran and Lozano-P{\'e}rez, Tom{\'a}s and Kaelbling, Leslie P},
  journal={arXiv preprint arXiv:1806.10166},
  year={2018}
}

@inproceedings{solomonoff1989system,
  title={A system for incremental learning based on algorithmic probability},
  author={Solomonoff, Ray J},
  booktitle={Proceedings of the Sixth Israeli Conference on Artificial Intelligence, Computer Vision and Pattern Recognition},
  pages={515--527},
  year={1989}
}

@book{thrun2012explanation,
  title={Explanation-based neural network learning: A lifelong learning approach},
  author={Thrun, Sebastian},
  volume={357},
  year={2012},
  publisher={Springer Science \& Business Media}
}

@inproceedings{carlson2010toward,
  title={Toward an architecture for never-ending language learning.},
  author={Carlson, Andrew and Betteridge, Justin and Kisiel, Bryan and Settles, Burr and Hruschka Jr, Estevam R and Mitchell, Tom M},
  booktitle={AAAI},
  volume={5},
  pages={3},
  year={2010},
  organization={Atlanta}
}

@article{2018_on_training_recurrent_neural_networks_for_lifelong_learning,
  title={On Training Recurrent Neural Networks for Lifelong Learning},
  author={Sodhani, Shagun and Chandar, Sarath and Bengio, Yoshua},
  journal={arXiv preprint arXiv:1811.07017},
  year={2018}
}

@incollection{mccloskey1989catastrophic,
  title={Catastrophic interference in connectionist networks: The sequential learning problem},
  author={McCloskey, Michael and Cohen, Neal J},
  booktitle={Psychology of learning and motivation},
  volume={24},
  pages={109--165},
  year={1989},
  publisher={Elsevier}
}

@inproceedings{sukhbaatar2015end,
  title={End-to-end memory networks},
  author={Sukhbaatar, Sainbayar and Weston, Jason and Fergus, Rob and others},
  booktitle={Advances in neural information processing systems},
  pages={2440--2448},
  year={2015}
}

@ARTICLE{MLSH,
   author = {{Frans}, K. and {Ho}, J. and {Chen}, X. and {Abbeel}, P. and 
	{Schulman}, J.},
    title = "{Meta Learning Shared Hierarchies}",
  journal = {arXiv e-prints},
archivePrefix = "arXiv",
   eprint = {1710.09767},
     year = 2017,
    month = oct,
   adsurl = {http://adsabs.harvard.edu/abs/2017arXiv171009767F}
}

@inproceedings{
chang2018automatically,
title={Automatically Composing Representation Transformations as a Means for Generalization},
author={Michael Chang and Abhishek Gupta and Sergey Levine and Thomas L. Griffiths},
booktitle={International Conference on Learning Representations},
year={2019},
url={https://openreview.net/forum?id=B1ffQnRcKX},
}

@article{cho2014learning,
  title={Learning phrase representations using RNN encoder-decoder for statistical machine translation},
  author={Cho, Kyunghyun and Van Merri{\"e}nboer, Bart and Gulcehre, Caglar and Bahdanau, Dzmitry and Bougares, Fethi and Schwenk, Holger and Bengio, Yoshua},
  journal={arXiv preprint arXiv:1406.1078},
  year={2014}
}

@article{hochreiter1997long,
  title={Long short-term memory},
  author={Hochreiter, Sepp and Schmidhuber, J{\"u}rgen},
  journal={Neural computation},
  volume={9},
  number={8},
  pages={1735--1780},
  year={1997},
  publisher={MIT Press}
}

@article{silver2016mastering,
  title={Mastering the game of Go with deep neural networks and tree search},
  author={Silver, David and Huang, Aja and Maddison, Chris J and Guez, Arthur and Sifre, Laurent and Van Den Driessche, George and Schrittwieser, Julian and Antonoglou, Ioannis and Panneershelvam, Veda and Lanctot, Marc and others},
  journal={nature},
  volume={529},
  number={7587},
  pages={484},
  year={2016},
  publisher={Nature Publishing Group}
}

@article{nichol2018first,
  title={On first-order meta-learning algorithms},
  author={Nichol, Alex and Achiam, Joshua and Schulman, John},
  journal={arXiv preprint arXiv:1803.02999},
  year={2018}
}

@article{zhang2020multi,
  title={Multi-Task Reinforcement Learning as a Hidden-Parameter Block MDP},
  author={Zhang, Amy and Sodhani, Shagun and Khetarpal, Khimya and Pineau, Joelle},
  journal={arXiv preprint arXiv:2007.07206},
  year={2020}
}

@inproceedings{
veniat2021efficient,
title={Efficient Continual Learning with Modular Networks and Task-Driven Priors},
author={Tom Veniat and Ludovic Denoyer and MarcAurelio Ranzato},
booktitle={International Conference on Learning Representations},
year={2021},
url={https://openreview.net/forum?id=EKV158tSfwv}
}

@article{sodhani2021multi,
  title={Multi-Task Reinforcement Learning with Context-based Representations},
  author={Sodhani, Shagun and Zhang, Amy and Pineau, Joelle},
  journal={arXiv preprint arXiv:2102.06177},
  year={2021}
}

@inproceedings{
goyal2021recurrent,
title={Recurrent Independent Mechanisms},
author={Anirudh Goyal and Alex Lamb and Jordan Hoffmann and Shagun Sodhani and Sergey Levine and Yoshua Bengio and Bernhard Sch{\"o}lkopf},
booktitle={International Conference on Learning Representations},
year={2021},
url={https://openreview.net/forum?id=mLcmdlEUxy-}
}

@phdthesis{schmidhuber1987evolutionary,
  title={Evolutionary principles in self-referential learning, or on learning how to learn: the meta-meta-... hook},
  author={Schmidhuber, J{\"u}rgen},
  year={1987},
  school={Technische Universit{\"a}t M{\"u}nchen}
}

@article{spelke1990principles,
  title={Principles of object perception},
  author={Spelke, Elizabeth S},
  journal={Cognitive science},
  volume={14},
  number={1},
  pages={29--56},
  year={1990},
  publisher={Elsevier}
}

@article{pinker1994language,
  title={The Language Instinct. How the Mind Creates Language},
  author={Pinker, Steven},
  year={1994}
}

@article{gopnik1988conceptual,
  title={Conceptual and semantic development as theory change: The case of object permanence},
  author={Gopnik, Alison},
  journal={Mind \& Language},
  volume={3},
  number={3},
  pages={197--216},
  year={1988},
  publisher={Wiley Online Library}
}

@book{carey1985conceptual,
  title={Conceptual change in childhood},
  author={Carey, Susan},
  year={1985},
  publisher={MIT press}
}

@article{wellman1992cognitive,
  title={Cognitive development: Foundational theories of core domains},
  author={Wellman, Henry M and Gelman, Susan A},
  journal={Annual review of psychology},
  volume={43},
  number={1},
  pages={337--375},
  year={1992},
  publisher={Annual Reviews 4139 El Camino Way, PO Box 10139, Palo Alto, CA 94303-0139, USA}
}

@article{pinker2005so,
  title={So how does the mind work?},
  author={Pinker, Steven},
  journal={Mind \& Language},
  volume={20},
  number={1},
  pages={1--24},
  year={2005},
  publisher={Wiley Online Library}
}

@article{spelke2007core,
  title={Core knowledge},
  author={Spelke, Elizabeth S and Kinzler, Katherine D},
  journal={Developmental science},
  volume={10},
  number={1},
  pages={89--96},
  year={2007},
  publisher={Wiley Online Library}
}

@article{xu2009induction,
  title={Induction, overhypotheses, and the shape bias: Some arguments and evidence for rational constructivism},
  author={Xu, Fei and Dewar, Kathryn and Perfors, Amy},
  journal={The origins of object knowledge},
  pages={263--284},
  year={2009}
}

@misc{malviya2021tag,
      title={TAG: Task-based Accumulated Gradients for Lifelong learning}, 
      author={Pranshu Malviya and Balaraman Ravindran and Sarath Chandar},
      year={2021},
      eprint={2105.05155},
      archivePrefix={arXiv},
      primaryClass={cs.LG}
}

@article{2019_a_review_of_modularization_techniques_in_artificial_neural_networks,
  title={A review of modularization techniques in artificial neural networks},
  author={Amer, Mohammed and Maul, Tom{\'a}s},
  journal={Artificial Intelligence Review},
  volume={52},
  number={1},
  pages={527--561},
  year={2019},
  publisher={Springer}
}

@article{1999_modularity_in_neural_computing,
  title={Modularity in neural computing},
  author={Caelli, Terry and Guan, Ling and Wen, Wilson},
  journal={Proceedings of the IEEE},
  volume={87},
  number={9},
  pages={1497--1518},
  year={1999},
  publisher={IEEE}
}

@article{1998_modular_neural_network_classifiers_a_comparative_study,
  title={Modular neural network classifiers: A comparative study},
  author={Auda, Gasser and Kamel, Mohamed},
  journal={Journal of Intelligent and Robotic Systems},
  volume={21},
  number={2},
  pages={117--129},
  year={1998},
  publisher={Springer}
}

@article{1999_modular_neural_networks_a_survey,
  title={Modular neural networks: a survey},
  author={Auda, Gasser and Kamel, Mohamed},
  journal={International Journal of Neural Systems},
  volume={9},
  number={02},
  pages={129--151},
  year={1999},
  publisher={World Scientific}
}

@article{1996_on_combining_artificial_neural_nets,
  title={On combining artificial neural nets},
  author={SHARKEY, AMANDA J C},
  journal={Connection science},
  volume={8},
  number={3-4},
  pages={299--314},
  year={1996},
  publisher={Taylor \& Francis}
}

@article {2015_the_modular_and_integrative_functional_architecture_of_the_human_brain,
	author = {Bertolero, Maxwell A. and Yeo, B. T. Thomas and D{\textquoteright}Esposito, Mark},
	title = {The modular and integrative functional architecture of the human brain},
	volume = {112},
	number = {49},
	pages = {E6798--E6807},
	year = {2015},
	doi = {10.1073/pnas.1510619112},
	publisher = {National Academy of Sciences},
	issn = {0027-8424},
	URL = {https://www.pnas.org/content/112/49/E6798},
	eprint = {https://www.pnas.org/content/112/49/E6798.full.pdf},
	journal = {Proceedings of the National Academy of Sciences}
}

@article{2002_revealing_modular_organization_in_the_yeast_transcriptional_network,
  title={Revealing modular organization in the yeast transcriptional network},
  author={Ihmels, Jan and Friedlander, Gilgi and Bergmann, Sven and Sarig, Ofer and Ziv, Yaniv and Barkai, Naama},
  journal={Nature genetics},
  volume={31},
  number={4},
  pages={370--377},
  year={2002},
  publisher={Nature Publishing Group}
}

@article{2006_modularity_and_community_structure_in_networks,
  title={Modularity and community structure in networks},
  author={Newman, Mark EJ},
  journal={Proceedings of the national academy of sciences},
  volume={103},
  number={23},
  pages={8577--8582},
  year={2006},
  publisher={National Acad Sciences}
}

@article{2007_the_road_to_modularity,
  title={The road to modularity},
  author={Wagner, G{\"u}nter P and Pavlicev, Mihaela and Cheverud, James M},
  journal={Nature Reviews Genetics},
  volume={8},
  number={12},
  pages={921--931},
  year={2007},
  publisher={Nature Publishing Group}
}

@article{2008revealing_modular_architecture_of_human_brain_structural_networks_by_using_cortical_thickness_from_MRI,
  title={Revealing modular architecture of human brain structural networks by using cortical thickness from MRI},
  author={Chen, Zhang J and He, Yong and Rosa-Neto, Pedro and Germann, Jurgen and Evans, Alan C},
  journal={Cerebral cortex},
  volume={18},
  number={10},
  pages={2374--2381},
  year={2008},
  publisher={Oxford University Press}
}

@article{2009complex_brain_networks_graph_theoretical_analysis_of_structural_and_functional_systems,
  title={Complex brain networks: graph theoretical analysis of structural and functional systems},
  author={Bullmore, Ed and Sporns, Olaf},
  journal={Nature reviews neuroscience},
  volume={10},
  number={3},
  pages={186--198},
  year={2009},
  publisher={Nature Publishing Group}
}

@article{2010efficient_physical_embedding_of_topologically_complex_information_processing_networks_in_brains_and_computer_circuits,
  title={Efficient physical embedding of topologically complex information processing networks in brains and computer circuits},
  author={Bassett, Danielle S and Greenfield, Daniel L and Meyer-Lindenberg, Andreas and Weinberger, Daniel R and Moore, Simon W and Bullmore, Edward T},
  journal={PLoS comput biol},
  volume={6},
  number={4},
  pages={e1000748},
  year={2010},
  publisher={Public Library of Science}
}

@article{2010modular_and_hierarchically_modular_organization_of_brain_networks,
  title={Modular and hierarchically modular organization of brain networks},
  author={Meunier, David and Lambiotte, Renaud and Bullmore, Edward T},
  journal={Frontiers in neuroscience},
  volume={4},
  pages={200},
  year={2010},
  publisher={Frontiers}
}

@article{goodfellow2020generative,
  title={Generative adversarial networks},
  author={Goodfellow, Ian and Pouget-Abadie, Jean and Mirza, Mehdi and Xu, Bing and Warde-Farley, David and Ozair, Sherjil and Courville, Aaron and Bengio, Yoshua},
  journal={Communications of the ACM},
  volume={63},
  number={11},
  pages={139--144},
  year={2020},
  publisher={ACM New York, NY, USA}
}

@article{2005_spontaneous_evolution_of_modularity_and_network_motifs,
  title={Spontaneous evolution of modularity and network motifs},
  author={Kashtan, Nadav and Alon, Uri},
  journal={Proceedings of the National Academy of Sciences},
  volume={102},
  number={39},
  pages={13773--13778},
  year={2005},
  publisher={National Acad Sciences}
}

@article{de2019continual,
  title={Continual learning: A comparative study on how to defy forgetting in classification tasks},
  author={De Lange, Matthias and Aljundi, Rahaf and Masana, Marc and Parisot, Sarah and Jia, Xu and Leonardis, Ales and Slabaugh, Gregory and Tuytelaars, Tinne},
  journal={arXiv preprint arXiv:1909.08383},
  volume={2},
  number={6},
  year={2019}
}

@article{2013_the_evolutionary_origins_of_modularity,
  title={The evolutionary origins of modularity},
  author={Clune, Jeff and Mouret, Jean-Baptiste and Lipson, Hod},
  journal={Proceedings of the Royal Society b: Biological sciences},
  volume={280},
  number={1755},
  pages={20122863},
  year={2013},
  publisher={The Royal Society}
}

@article{2007_varying_environments_can_speed_up_evolution,
  title={Varying environments can speed up evolution},
  author={Kashtan, Nadav and Noor, Elad and Alon, Uri},
  journal={Proceedings of the National Academy of Sciences},
  volume={104},
  number={34},
  pages={13711--13716},
  year={2007},
  publisher={National Acad Sciences}
}

@inproceedings{2017_expert_gate_lifelong_learning_with_a_network_of_experts,
  title={Expert gate: Lifelong learning with a network of experts},
  author={Aljundi, Rahaf and Chakravarty, Punarjay and Tuytelaars, Tinne},
  booktitle={Proceedings of the IEEE Conference on Computer Vision and Pattern Recognition},
  pages={3366--3375},
  year={2017}
}

@InProceedings{2017_encoder_based_lifelong_learning,
author = {Rannen, Amal and Aljundi, Rahaf and Blaschko, Matthew B. and Tuytelaars, Tinne},
title = {Encoder Based Lifelong Learning},
booktitle = {Proceedings of the IEEE International Conference on Computer Vision (ICCV)},
month = {Oct},
year = {2017}
}

@inproceedings{
2020_compositional_language_continual_learning,
title={Compositional Language Continual Learning},
author={Yuanpeng Li and Liang Zhao and Kenneth Church and Mohamed Elhoseiny},
booktitle={International Conference on Learning Representations},
year={2020},
url={https://openreview.net/forum?id=rklnDgHtDS}
}

@article{2019_compositional_generalization_for_primitive_substitutions,
  title={Compositional generalization for primitive substitutions},
  author={Li, Yuanpeng and Zhao, Liang and Wang, Jianyu and Hestness, Joel},
  journal={arXiv preprint arXiv:1910.02612},
  year={2019}
}

@article{2020_ternary_feature_masks_continual_learning_without_any_forgetting,
  title={Ternary feature masks: continual learning without any forgetting},
  author={Masana, Marc and Tuytelaars, Tinne and van de Weijer, Joost},
  journal={arXiv preprint arXiv:2001.08714},
  year={2020}
}

@inproceedings{2018_piggyback_adapting_a_single_network_to_multiple_tasks_by_learning_to_mask_weights,
  title={Piggyback: Adapting a single network to multiple tasks by learning to mask weights},
  author={Mallya, Arun and Davis, Dillon and Lazebnik, Svetlana},
  booktitle={Proceedings of the European Conference on Computer Vision (ECCV)},
  pages={67--82},
  year={2018}
}

@article{2016_binarized_neural_networks_training_deep_neural_networks_with_weights_and_activations_constrained_to_plus_1_or_minus_1,
  title={Binarized neural networks: Training deep neural networks with weights and activations constrained to+ 1 or-1},
  author={Courbariaux, Matthieu and Hubara, Itay and Soudry, Daniel and El-Yaniv, Ran and Bengio, Yoshua},
  journal={arXiv preprint arXiv:1602.02830},
  year={2016}
}

@article{2015_binaryconnect_training_deep_neural_networks_with_binary_weights_during_propagations,
  title={Binaryconnect: Training deep neural networks with binary weights during propagations},
  author={Courbariaux, Matthieu and Bengio, Yoshua and David, Jean-Pierre},
  journal={arXiv preprint arXiv:1511.00363},
  year={2015}
}

@inproceedings{titsias2019functional,
    title={Functional Regularisation for  Continual Learning with Gaussian Processes},
    author={Michalis K. Titsias and Jonathan Schwarz and Alexander G. de G. Matthews and Razvan Pascanu and Yee Whye Teh},
    booktitle={International Conference on Learning Representations},
    year={2020},
    url={https://openreview.net/forum?id=HkxCzeHFDB}
}

@inproceedings{2018_overcoming_catastrophic_forgetting_with_hard_attention_to_the_task,
  title={Overcoming catastrophic forgetting with hard attention to the task},
  author={Serra, Joan and Suris, Didac and Miron, Marius and Karatzoglou, Alexandros},
  booktitle={International Conference on Machine Learning},
  pages={4548--4557},
  year={2018},
  organization={PMLR}
}

@article{2017_overcoming_catastrophic_forgetting_by_incremental_moment_matching,
  title={Overcoming catastrophic forgetting by incremental moment matching},
  author={Lee, Sang-Woo and Kim, Jin-Hwa and Jun, Jaehyun and Ha, Jung-Woo and Zhang, Byoung-Tak},
  journal={arXiv preprint arXiv:1703.08475},
  year={2017}
}

@article{2017_learning_without_forgetting,
  title={Learning without forgetting},
  author={Li, Zhizhong and Hoiem, Derek},
  journal={IEEE transactions on pattern analysis and machine intelligence},
  volume={40},
  number={12},
  pages={2935--2947},
  year={2017},
  publisher={IEEE}
}

@inproceedings{2016_binarized_neural_networks,
  title={Binarized neural networks},
  author={Hubara, Itay and Courbariaux, Matthieu and Soudry, Daniel and El-Yaniv, Ran and Bengio, Yoshua},
  booktitle={Proceedings of the 30th International Conference on Neural Information Processing Systems},
  pages={4114--4122},
  year={2016}
}

@inproceedings{2016_dynamic_network_surgery_for_efficient_DNNs,
  title={Dynamic network surgery for efficient DNNs},
  author={Guo, Yiwen and Yao, Anbang and Chen, Yurong},
  booktitle={Proceedings of the 30th International Conference on Neural Information Processing Systems},
  pages={1387--1395},
  year={2016}
}

@inproceedings{2018_packnet_adding_multiple_tasks_to_a_single_network_by_iterative_pruning,
  title={Packnet: Adding multiple tasks to a single network by iterative pruning},
  author={Mallya, Arun and Lazebnik, Svetlana},
  booktitle={Proceedings of the IEEE Conference on Computer Vision and Pattern Recognition},
  pages={7765--7773},
  year={2018}
}

@inproceedings{2015_learning_both_weights_and_connections_for_efficient_neural_networks,
  title={Learning both weights and connections for efficient neural networks},
  author={Han, Song and Pool, Jeff and Tran, John and Dally, William J},
  booktitle={Proceedings of the 28th International Conference on Neural Information Processing Systems-Volume 1},
  pages={1135--1143},
  year={2015}
}

@article{2016_dsd_dense_sparse_dense_training_for_deep_neural_networks,
  title={Dsd: Dense-sparse-dense training for deep neural networks},
  author={Han, Song and Pool, Jeff and Narang, Sharan and Mao, Huizi and Gong, Enhao and Tang, Shijian and Elsen, Erich and Vajda, Peter and Paluri, Manohar and Tran, John and others},
  journal={arXiv preprint arXiv:1607.04381},
  year={2016}
}

@article{2015_net2net_accelerating_learning_via_knowledge_transfer,
  title={Net2net: Accelerating learning via knowledge transfer},
  author={Chen, Tianqi and Goodfellow, Ian and Shlens, Jonathon},
  journal={arXiv preprint arXiv:1511.05641},
  year={2015}
}

@inproceedings{2019_random_path_selection_for_continual_learning,
 author = {Rajasegaran, Jathushan and Hayat, Munawar and Khan, Salman H and Khan, Fahad Shahbaz and Shao, Ling},
 booktitle = {Advances in Neural Information Processing Systems},
 editor = {H. Wallach and H. Larochelle and A. Beygelzimer and F. d\textquotesingle Alch\'{e}-Buc and E. Fox and R. Garnett},
 pages = {},
 publisher = {Curran Associates, Inc.},
 title = {Random Path Selection for Continual Learning},
 url = {https://proceedings.neurips.cc/paper/2019/file/83da7c539e1ab4e759623c38d8737e9e-Paper.pdf},
 volume = {32},
 year = {2019}
}

@article{2016_progressive_neural_networks,
  title={Progressive neural networks},
  author={Rusu, Andrei A and Rabinowitz, Neil C and Desjardins, Guillaume and Soyer, Hubert and Kirkpatrick, James and Kavukcuoglu, Koray and Pascanu, Razvan and Hadsell, Raia},
  journal={arXiv preprint arXiv:1606.04671},
  year={2016}
}

@article{2017_pathnet_evolution_channels_gradient_descent_in_super_neural_networks,
  title={Pathnet: Evolution channels gradient descent in super neural networks},
  author={Fernando, Chrisantha and Banarse, Dylan and Blundell, Charles and Zwols, Yori and Ha, David and Rusu, Andrei A and Pritzel, Alexander and Wierstra, Daan},
  journal={arXiv preprint arXiv:1701.08734},
  year={2017}
}

@article{bengio2019meta,
  title={A meta-transfer objective for learning to disentangle causal mechanisms},
  author={Bengio, Yoshua and Deleu, Tristan and Rahaman, Nasim and Ke, Rosemary and Lachapelle, S{\'e}bastien and Bilaniuk, Olexa and Goyal, Anirudh and Pal, Christopher},
  journal={arXiv preprint arXiv:1901.10912},
  year={2019}
}

@article{2019_an_adaptive_random_path_selection_approach_for_incremental_learning,
  title={An adaptive random path selection approach for incremental learning},
  author={Rajasegaran, Jathushan and Hayat, Munawar and Khan, Salman and Khan, Fahad Shahbaz and Shao, Ling and Yang, Ming-Hsuan},
  journal={arXiv preprint arXiv:1906.01120},
  year={2019}
}

@article{swevers1997optimal,
  title={Optimal robot excitation and identification},
  author={Swevers, Jan and Ganseman, Chris and Tukel, D Bilgin and De Schutter, Joris and Van Brussel, Hendrik},
  journal={IEEE transactions on robotics and automation},
  volume={13},
  number={5},
  pages={730--740},
  year={1997},
  publisher={IEEE}
}

@article{zhu2017fast,
  title={Fast model identification via physics engines for data-efficient policy search},
  author={Zhu, Shaojun and Kimmel, Andrew and Bekris, Kostas E and Boularias, Abdeslam},
  journal={arXiv preprint arXiv:1710.08893},
  year={2017}
}

@inproceedings{e0da8f25-d850-4a80-af21-e151cc28c4f4,
  author       = {Åström, Karl Johan and Bohlin, Torsten},
  booktitle    = {Proc. IFAC Conference on Self-Adaptive Control Systems},
  language     = {eng},
  title        = {Numerical Identification of Linear Dynamic Systems from Normal Operating Records},
  year         = {1965},
}

@book{van2012subspace,
  title={Subspace identification for linear systems: Theory—Implementation—Applications},
  author={Van Overschee, Peter and De Moor, BL},
  year={2012},
  publisher={Springer Science \& Business Media}
}

@article{zadeh1956identification,
  title={On the identification problem},
  author={Zadeh, L},
  journal={IRE Transactions on Circuit Theory},
  volume={3},
  number={4},
  pages={277--281},
  year={1956},
  publisher={IEEE}
}

@article{LJUNG20101,
title = {Perspectives on system identification},
journal = {Annual Reviews in Control},
volume = {34},
number = {1},
pages = {1-12},
year = {2010},
issn = {1367-5788},
doi = {https://doi.org/10.1016/j.arcontrol.2009.12.001},
url = {https://www.sciencedirect.com/science/article/pii/S1367578810000027},
author = {Lennart Ljung}
}

@article{chiuso2019system,
  title={System identification: A machine learning perspective},
  author={Chiuso, Alessandro and Pillonetto, Gianluigi},
  journal={Annual Review of Control, Robotics, and Autonomous Systems},
  volume={2},
  pages={281--304},
  year={2019},
  publisher={Annual Reviews}
}

@inproceedings{bhat2002computing,
  title={Computing the physical parameters of rigid-body motion from video},
  author={Bhat, Kiran S and Seitz, Steven M and Popovi{\'c}, Jovan and Khosla, Pradeep K},
  booktitle={European Conference on Computer Vision},
  pages={551--565},
  year={2002},
  organization={Springer}
}

@article{gevers2006system,
  title={System Identification without Lennart Ljung: what would have been different?},
  author={Gevers, Michel and others},
  journal={Forever Ljung in System Identification, Studentlitteratur AB, Norrtalje},
  volume={2},
  year={2006}
}

@inproceedings{ajay2019combining,
  title={Combining physical simulators and object-based networks for control},
  author={Ajay, Anurag and Bauza, Maria and Wu, Jiajun and Fazeli, Nima and Tenenbaum, Joshua B and Rodriguez, Alberto and Kaelbling, Leslie P},
  booktitle={2019 International Conference on Robotics and Automation (ICRA)},
  pages={3217--3223},
  year={2019},
  organization={IEEE}
}

@inproceedings{yu2017osi,
  author    = {Wenhao Yu and
               Jie Tan and
               C. Karen Liu and
               Greg Turk},
  title     = {Preparing for the Unknown: Learning a Universal Policy with Online
               System Identification},
  booktitle = {Robotics: Science and Systems},
  year      = {2017}
}

@article{lecun2010mnist,
  title={MNIST handwritten digit database},
  author={LeCun, Yann and Cortes, Corinna and Burges, CJ},
  journal={ATT Labs [Online]. Available: http://yann.lecun.com/exdb/mnist},
  volume={2},
  year={2010}
}

@misc{goodfellow2015empirical,
      title={An Empirical Investigation of Catastrophic Forgetting in Gradient-Based Neural Networks}, 
      author={Ian J. Goodfellow and Mehdi Mirza and Da Xiao and Aaron Courville and Yoshua Bengio},
      year={2015},
      eprint={1312.6211},
      archivePrefix={arXiv},
      primaryClass={stat.ML}
}

@article{clanuwat2018deep,
  title={Deep learning for classical japanese literature},
  author={Clanuwat, Tarin and Bober-Irizar, Mikel and Kitamoto, Asanobu and Lamb, Alex and Yamamoto, Kazuaki and Ha, David},
  journal={arXiv preprint arXiv:1812.01718},
  year={2018}
}

@article{krizhevsky2009learning,
  title={Learning multiple layers of features from tiny images},
  author={Krizhevsky, Alex and Hinton, Geoffrey and others},
  year={2009},
  publisher={Citeseer}
}

@inproceedings{deng2009imagenet,
  title={Imagenet: A large-scale hierarchical image database},
  author={Deng, Jia and Dong, Wei and Socher, Richard and Li, Li-Jia and Li, Kai and Fei-Fei, Li},
  booktitle={2009 IEEE conference on computer vision and pattern recognition},
  pages={248--255},
  year={2009},
  organization={Ieee}
}

@misc{nguyen2017variational,
    title={Variational Continual Learning},
    author={Cuong V. Nguyen and Yingzhen Li and Thang D. Bui and Richard E. Turner},
    year={2017},
    eprint={1710.10628},
    archivePrefix={arXiv},
    primaryClass={stat.ML}
}

@misc{lesort2018generative,
    title={Generative Models from the perspective of Continual Learning},
    author={Timothee Lesort and Hugo Caselles-Dupre and Michael Garcia-Ortiz and Andrei Stoian and David Filliat},
    year={2018},
    eprint={1812.09111},
    archivePrefix={arXiv},
    primaryClass={cs.LG}
}

@article{Rebuffi_2017,
   title={iCaRL: Incremental Classifier and Representation Learning},
   ISBN={9781538604571},
   url={http://dx.doi.org/10.1109/CVPR.2017.587},
   DOI={10.1109/cvpr.2017.587},
   journal={2017 IEEE Conference on Computer Vision and Pattern Recognition (CVPR)},
   publisher={IEEE},
   author={Rebuffi, Sylvestre-Alvise and Kolesnikov, Alexander and Sperl, Georg and Lampert, Christoph H.},
   year={2017},
   month={Jul}
}

@misc{castro2018endtoend,
    title={End-to-End Incremental Learning},
    author={Francisco M. Castro and Manuel J. Marín-Jimenez and Nicolás Guil and Cordelia Schmid and Karteek Alahari},
    year={2018},
    eprint={1807.09536},
    archivePrefix={arXiv},
    primaryClass={cs.CV}
}

@misc{wu2019large,
    title={Large Scale Incremental Learning},
    author={Yue Wu and Yinpeng Chen and Lijuan Wang and Yuancheng Ye and Zicheng Liu and Yandong Guo and Yun Fu},
    year={2019},
    eprint={1905.13260},
    archivePrefix={arXiv},
    primaryClass={cs.CV}
}

@misc{li2019learn,
    title={Learn to Grow: A Continual Structure Learning Framework for Overcoming Catastrophic Forgetting},
    author={Xilai Li and Yingbo Zhou and Tianfu Wu and Richard Socher and Caiming Xiong},
    year={2019},
    eprint={1904.00310},
    archivePrefix={arXiv},
    primaryClass={cs.LG}
}

@misc{lomonaco2017core50,
    title={CORe50: a New Dataset and Benchmark for Continuous Object Recognition},
    author={Vincenzo Lomonaco and Davide Maltoni},
    year={2017},
    eprint={1705.03550},
    archivePrefix={arXiv},
    primaryClass={cs.CV}
}

@misc{antoniou2020defining,
    title={Defining Benchmarks for Continual Few-Shot Learning},
    author={Antreas Antoniou and Massimiliano Patacchiola and Mateusz Ochal and Amos Storkey},
    year={2020},
    eprint={2004.11967},
    archivePrefix={arXiv},
    primaryClass={cs.CV}
}

@article{laleh2020chaotic,
  title={Chaotic Continual Learning},
  author={Laleh, Touraj and Faramarzi, Mojtaba and Rish, Irina and Chandar, Sarath},
  year={2020}
}

@misc{hayes2020remind,
      title={REMIND Your Neural Network to Prevent Catastrophic Forgetting}, 
      author={Tyler L. Hayes and Kushal Kafle and Robik Shrestha and Manoj Acharya and Christopher Kanan},
      year={2020},
      eprint={1910.02509},
      archivePrefix={arXiv},
      primaryClass={cs.LG}
}

@article{10.1093/mind/LIX.236.433,
    author = {TURING, A. M.},
    title = "{I.—COMPUTING MACHINERY AND INTELLIGENCE}",
    journal = {Mind},
    volume = {LIX},
    number = {236},
    pages = {433-460},
    year = {1950},
    month = {10},
    issn = {0026-4423},
    doi = {10.1093/mind/LIX.236.433},
    url = {https://doi.org/10.1093/mind/LIX.236.433},
    eprint = {https://academic.oup.com/mind/article-pdf/LIX/236/433/30123314/lix-236-433.pdf},
}

@article{senior2020improved,
  title={Improved protein structure prediction using potentials from deep learning},
  author={Senior, Andrew W and Evans, Richard and Jumper, John and Kirkpatrick, James and Sifre, Laurent and Green, Tim and Qin, Chongli and {\v{Z}}{\'\i}dek, Augustin and Nelson, Alexander WR and Bridgland, Alex and others},
  journal={Nature},
  volume={577},
  number={7792},
  pages={706--710},
  year={2020},
  publisher={Nature Publishing Group}
}

@article{bapst2020unveiling,
  title={Unveiling the predictive power of static structure in glassy systems},
  author={Bapst, Victor and Keck, Thomas and Grabska-Barwi{\'n}ska, A and Donner, Craig and Cubuk, Ekin Dogus and Schoenholz, Samuel S and Obika, Annette and Nelson, Alexander WR and Back, Trevor and Hassabis, Demis and others},
  journal={Nature Physics},
  volume={16},
  number={4},
  pages={448--454},
  year={2020},
  publisher={Nature Publishing Group}
}

@inproceedings{badia2020agent57,
  title={Agent57: Outperforming the atari human benchmark},
  author={Badia, Adri{\`a} Puigdom{\`e}nech and Piot, Bilal and Kapturowski, Steven and Sprechmann, Pablo and Vitvitskyi, Alex and Guo, Zhaohan Daniel and Blundell, Charles},
  booktitle={International Conference on Machine Learning},
  pages={507--517},
  year={2020},
  organization={PMLR}
}

@article{yim2020predicting,
  title={Predicting conversion to wet age-related macular degeneration using deep learning},
  author={Yim, Jason and Chopra, Reena and Spitz, Terry and Winkens, Jim and Obika, Annette and Kelly, Christopher and Askham, Harry and Lukic, Marko and Huemer, Josef and Fasler, Katrin and others},
  journal={Nature Medicine},
  volume={26},
  number={6},
  pages={892--899},
  year={2020},
  publisher={Nature Publishing Group}
}

@article{PhysRevResearch.2.033429,
  title = {Ab initio solution of the many-electron Schr\"odinger equation with deep neural networks},
  author = {Pfau, David and Spencer, James S. and Matthews, Alexander G. D. G. and Foulkes, W. M. C.},
  journal = {Phys. Rev. Research},
  volume = {2},
  issue = {3},
  pages = {033429},
  numpages = {20},
  year = {2020},
  month = {Sep},
  publisher = {American Physical Society},
  doi = {10.1103/PhysRevResearch.2.033429},
  url = {https://link.aps.org/doi/10.1103/PhysRevResearch.2.033429}
}

@article{silver2018general,
  title={A general reinforcement learning algorithm that masters chess, shogi, and Go through self-play},
  author={Silver, David and Hubert, Thomas and Schrittwieser, Julian and Antonoglou, Ioannis and Lai, Matthew and Guez, Arthur and Lanctot, Marc and Sifre, Laurent and Kumaran, Dharshan and Graepel, Thore and others},
  journal={Science},
  volume={362},
  number={6419},
  pages={1140--1144},
  year={2018},
  publisher={American Association for the Advancement of Science}
}

@article{schrittwieser2020mastering,
  title={Mastering atari, go, chess and shogi by planning with a learned model},
  author={Schrittwieser, Julian and Antonoglou, Ioannis and Hubert, Thomas and Simonyan, Karen and Sifre, Laurent and Schmitt, Simon and Guez, Arthur and Lockhart, Edward and Hassabis, Demis and Graepel, Thore and others},
  journal={Nature},
  volume={588},
  number={7839},
  pages={604--609},
  year={2020},
  publisher={Nature Publishing Group}
}

@inproceedings{
gemp2021eigengame,
title={EigenGame: {\{}PCA{\}} as a Nash Equilibrium},
author={Ian Gemp and Brian McWilliams and Claire Vernade and Thore Graepel},
booktitle={International Conference on Learning Representations},
year={2021},
url={https://openreview.net/forum?id=NzTU59SYbNq}
}

@InProceedings{Stojanov_2019_CVPR,
author = {Stojanov, Stefan and Mishra, Samarth and Thai, Ngoc Anh and Dhanda, Nikhil and Humayun, Ahmad and Yu, Chen and Smith, Linda B. and Rehg, James M.},
title = {Incremental Object Learning From Contiguous Views},
booktitle = {Proceedings of the IEEE/CVF Conference on Computer Vision and Pattern Recognition (CVPR)},
month = {June},
year = {2019}
}

@inproceedings{she2019openlorisobject,
    title={ {OpenLORIS-Object}: A Robotic Vision Dataset and Benchmark for Lifelong Deep Learning},
    author={Qi She and Fan Feng and Xinyue Hao and Qihan Yang and Chuanlin Lan and Vincenzo Lomonaco and Xuesong Shi and Zhengwei Wang and Yao Guo and Yimin Zhang and Fei Qiao and Rosa H. M. Chan},
    booktitle={2020 International Conference on Robotics and Automation (ICRA)},
    year={2020},
    pages={4767-4773},
}

@article{leibo2018psychlab,
  title={Psychlab: a psychology laboratory for deep reinforcement learning agents},
  author={Leibo, Joel Z and d'Autume, Cyprien de Masson and Zoran, Daniel and Amos, David and Beattie, Charles and Anderson, Keith and Casta{\~n}eda, Antonio Garc{\'\i}a and Sanchez, Manuel and Green, Simon and Gruslys, Audrunas and others},
  journal={arXiv preprint arXiv:1801.08116},
  year={2018}
}

@article{tomavsev2019clinically,
  title={A clinically applicable approach to continuous prediction of future acute kidney injury},
  author={Toma{\v{s}}ev, Nenad and Glorot, Xavier and Rae, Jack W and Zielinski, Michal and Askham, Harry and Saraiva, Andre and Mottram, Anne and Meyer, Clemens and Ravuri, Suman and Protsyuk, Ivan and others},
  journal={Nature},
  volume={572},
  number={7767},
  pages={116--119},
  year={2019},
  publisher={Nature Publishing Group}
}

@article{vinyals2019grandmaster,
  title={Grandmaster level in StarCraft II using multi-agent reinforcement learning},
  author={Vinyals, Oriol and Babuschkin, Igor and Czarnecki, Wojciech M and Mathieu, Micha{\"e}l and Dudzik, Andrew and Chung, Junyoung and Choi, David H and Powell, Richard and Ewalds, Timo and Georgiev, Petko and others},
  journal={Nature},
  volume={575},
  number={7782},
  pages={350--354},
  year={2019},
  publisher={Nature Publishing Group}
}

@InProceedings{Roady_2020_Stream51,
	author = {Roady, Ryne and Hayes, Tyler L. and Vaidya, Hitesh and Kanan, Christopher},
	title = {Stream-51: Streaming Classification and Novelty Detection From Videos},
	booktitle = {The IEEE/CVF Conference on Computer Vision and Pattern Recognition (CVPR) Workshops},
	month = {June},
	year = {2020}
}

@InProceedings{Abdelsalam_2021_CVPR,
    author    = {Abdelsalam, Mohamed and Faramarzi, Mojtaba and Sodhani, Shagun and Chandar, Sarath},
    title     = {IIRC: Incremental Implicitly-Refined Classification},
    booktitle = {Proceedings of the IEEE/CVF Conference on Computer Vision and Pattern Recognition (CVPR)},
    month     = {June},
    year      = {2021},
    pages     = {11038-11047}
}

@misc{lopezpaz2017gradient,
      title={Gradient Episodic Memory for Continual Learning}, 
      author={David Lopez-Paz and Marc'Aurelio Ranzato},
      year={2017},
      eprint={1706.08840},
      archivePrefix={arXiv},
      primaryClass={cs.LG}
}

@misc{lesort2019continual,
      title={Continual Learning for Robotics: Definition, Framework, Learning Strategies, Opportunities and Challenges}, 
      author={Timothee Lesort and Vincenzo Lomonaco and Andrei Stoian and Davide Maltoni and David Filliat and Natalia Díaz-Rodríguez},
      year={2019},
      eprint={1907.00182},
      archivePrefix={arXiv},
      primaryClass={cs.LG}
}

@misc{ebrahimi2020adversarial,
      title={Adversarial Continual Learning}, 
      author={Sayna Ebrahimi and Franziska Meier and Roberto Calandra and Trevor Darrell and Marcus Rohrbach},
      year={2020},
      eprint={2003.09553},
      archivePrefix={arXiv},
      primaryClass={cs.LG}
}

@InProceedings{Hou_2019_CVPR,
    author = {Hou, Saihui and Pan, Xinyu and Loy, Chen Change and Wang, Zilei and Lin, Dahua},
    title = {Learning a Unified Classifier Incrementally via Rebalancing},
    booktitle = {Proceedings of the IEEE/CVF Conference on Computer Vision and Pattern Recognition (CVPR)},
    month = {June},
    year = {2019}
}

@TECHREPORT{Sorower10aliterature,
    author = {Mohammad S Sorower},
    title = {A literature survey on algorithms for multi-label learning},
    institution = {},
    year = {2010}
}

@misc{vandeven2019scenarios,
      title={Three scenarios for continual learning}, 
      author={Gido M. van de Ven and Andreas S. Tolias},
      year={2019},
      eprint={1904.07734},
      archivePrefix={arXiv},
      primaryClass={cs.LG}
}

@misc{swaroop2019improving,
      title={Improving and Understanding Variational Continual Learning}, 
      author={Siddharth Swaroop and Cuong V. Nguyen and Thang D. Bui and Richard E. Turner},
      year={2019},
      eprint={1905.02099},
      archivePrefix={arXiv},
      primaryClass={stat.ML}
}

@misc{vandeven2019generative,
      title={Generative replay with feedback connections as a general strategy for continual learning}, 
      author={Gido M. van de Ven and Andreas S. Tolias},
      year={2019},
      eprint={1809.10635},
      archivePrefix={arXiv},
      primaryClass={cs.LG}
}

@misc{aljundi2019online,
      title={Online Continual Learning with Maximally Interfered Retrieval}, 
      author={Rahaf Aljundi and Lucas Caccia and Eugene Belilovsky and Massimo Caccia and Min Lin and Laurent Charlin and Tinne Tuytelaars},
      year={2019},
      eprint={1908.04742},
      archivePrefix={arXiv},
      primaryClass={cs.LG}
}

@inproceedings{risk_minimization, author = {Vapnik, V.}, title = {Principles of Risk Minimization for Learning Theory}, year = {1991}, isbn = {1558602224}, publisher = {Morgan Kaufmann Publishers Inc.}, address = {San Francisco, CA, USA}, booktitle = {Proceedings of the 4th International Conference on Neural Information Processing Systems}, pages = {831–838}, numpages = {8}, location = {Denver, Colorado}, series = {NIPS'91} }

@misc{farquhar2019robust,
      title={Towards Robust Evaluations of Continual Learning}, 
      author={Sebastian Farquhar and Yarin Gal},
      year={2019},
      eprint={1805.09733},
      archivePrefix={arXiv},
      primaryClass={stat.ML}
}

@article{xiao2017_online,
  title={Fashion-mnist: a novel image dataset for benchmarking machine learning algorithms},
  author={Xiao, Han and Rasul, Kashif and Vollgraf, Roland},
  journal={arXiv preprint arXiv:1708.07747},
  year={2017}
}

@article{ratcliff1990connectionist,
  title={Connectionist models of recognition memory: constraints imposed by learning and forgetting functions.},
  author={Ratcliff, Roger},
  journal={Psychological review},
  volume={97},
  number={2},
  pages={285},
  year={1990},
  publisher={American Psychological Association}
}

@article{schaul2018barbados,
  title={The barbados 2018 list of open issues in continual learning},
  author={Schaul, Tom and van Hasselt, Hado and Modayil, Joseph and White, Martha and White, Adam and Bacon, Pierre-Luc and Harb, Jean and Mourad, Shibl and Bellemare, Marc and Precup, Doina},
  journal={arXiv preprint arXiv:1811.07004},
  year={2018}
}

@incollection{bengio2013optimization,
  title={On the optimization of a synaptic learning rule},
  author={Bengio, Samy and Bengio, Yoshua and Cloutier, Jocelyn and Gescei, Jan},
  booktitle={Optimality in Biological and Artificial Networks?},
  pages={281--303},
  year={2013},
  publisher={Routledge}
}

@article{mnih2015human,
  author = {Mnih, Volodymyr and Kavukcuoglu, Koray and Silver, David and Rusu, Andrei A. and Veness, Joel and Bellemare, Marc G. and Graves, Alex and Riedmiller, Martin and Fidjeland, Andreas K. and Ostrovski, Georg and Petersen, Stig and Beattie, Charles and Sadik, Amir and Antonoglou, Ioannis and King, Helen and Kumaran, Dharshan and Wierstra, Daan and Legg, Shane and Hassabis, Demis},
  description = {Human-level control through deep reinforcement learning - nature14236.pdf},
  issn = {00280836},
  journal = {Nature},
  month = feb,
  number = 7540,
  pages = {529--533},
  publisher = {Nature Publishing Group, a division of Macmillan Publishers Limited. All Rights Reserved.},
  title = {Human-level control through deep reinforcement learning},
  url = {http://dx.doi.org/10.1038/nature14236},
  volume = 518,
  year = 2015
}

@inproceedings{chen2018gradnorm,
  title={Gradnorm: Gradient normalization for adaptive loss balancing in deep multitask networks},
  author={Chen, Zhao and Badrinarayanan, Vijay and Lee, Chen-Yu and Rabinovich, Andrew},
  booktitle={International Conference on Machine Learning},
  pages={794--803},
  year={2018},
  organization={PMLR}
}

@article{radford2019language_models_are_unsupervised_multitask_learners,
  title={Language models are unsupervised multitask learners},
  author={Radford, Alec and Wu, Jeffrey and Child, Rewon and Luan, David and Amodei, Dario and Sutskever, Ilya},
  journal={OpenAI Blog},
  volume={1},
  number={8},
  pages={9},
  year={2019}
}

@article{caruana1997multitask_learning,
  title={Multitask learning},
  author={Caruana, Rich},
  journal={Machine learning},
  volume={28},
  number={1},
  pages={41--75},
  year={1997},
  publisher={Springer}
}

@inproceedings{zhang2014facial_landmark_detection_by_deep_multitask_learning,
  title={Facial landmark detection by deep multi-task learning},
  author={Zhang, Zhanpeng and Luo, Ping and Loy, Chen Change and Tang, Xiaoou},
  booktitle={European conference on computer vision},
  pages={94--108},
  year={2014},
  organization={Springer}
}

@article{regularizing_deep_multi_task_networks_using_orthogonal_gradients,
  title={Regularizing Deep Multi-Task Networks using Orthogonal Gradients},
  author={Suteu, Mihai and Guo, Yike},
  journal={arXiv preprint arXiv:1912.06844},
  year={2019}
}

@article{sussmann1992uniqueness,
  title={Uniqueness of the weights for minimal feedforward nets with a given input-output map},
  author={Sussmann, H{\'e}ctor J},
  journal={Neural networks},
  volume={5},
  number={4},
  pages={589--593},
  year={1992},
  publisher={Elsevier}
}

@article{adapting_auxiliary_losses_using_gradient_similarity,
  title={Adapting auxiliary losses using gradient similarity},
  author={Du, Yunshu and Czarnecki, Wojciech M and Jayakumar, Siddhant M and Pascanu, Razvan and Lakshminarayanan, Balaji},
  journal={arXiv preprint arXiv:1812.02224},
  year={2018}
}

@article{feudal_networks_for_hierarchical_reinforcement_learning,
  title={Feudal networks for hierarchical reinforcement learning},
  author={Vezhnevets, Alexander Sasha and Osindero, Simon and Schaul, Tom and Heess, Nicolas and Jaderberg, Max and Silver, David and Kavukcuoglu, Koray},
  journal={arXiv preprint arXiv:1703.01161},
  year={2017}
}

@incollection{hecht1992theory,
  title={Theory of the backpropagation neural network},
  author={Hecht-Nielsen, Robert},
  booktitle={Neural networks for perception},
  pages={65--93},
  year={1992},
  publisher={Elsevier}
}

@inproceedings{babyai,
title={Baby{AI}: First Steps Towards Grounded Language Learning With a Human In the Loop},
author={Maxime Chevalier-Boisvert and Dzmitry Bahdanau and Salem Lahlou and Lucas Willems and Chitwan Saharia and Thien Huu Nguyen and Yoshua Bengio},
booktitle={International Conference on Learning Representations},
year={2019},
url={https://openreview.net/forum?id=rJeXCo0cYX},
}

@inproceedings{end_to_end_multi_task_learning_with_attention,
  title={End-to-end multi-task learning with attention},
  author={Liu, Shikun and Johns, Edward and Davison, Andrew J},
  booktitle={Proceedings of the IEEE Conference on Computer Vision and Pattern Recognition},
  pages={1871--1880},
  year={2019}
}

@inproceedings{learning_modular_neural_network_policies_for_multi_task_and_multi_robot_transfer,
  title={Learning modular neural network policies for multi-task and multi-robot transfer},
  author={Devin, Coline and Gupta, Abhishek and Darrell, Trevor and Abbeel, Pieter and Levine, Sergey},
  booktitle={2017 IEEE International Conference on Robotics and Automation (ICRA)},
  pages={2169--2176},
  year={2017},
  organization={IEEE}
}

@article{ruder2017overview,
  title={An overview of multi-task learning in deep neural networks},
  author={Ruder, Sebastian},
  journal={arXiv preprint arXiv:1706.05098},
  year={2017}
}

@inproceedings{pentina2015curriculum,
  title={Curriculum learning of multiple tasks},
  author={Pentina, Anastasia and Sharmanska, Viktoriia and Lampert, Christoph H},
  booktitle={Proceedings of the IEEE Conference on Computer Vision and Pattern Recognition},
  pages={5492--5500},
  year={2015}
}

@inproceedings{bengio2009curriculum,
  title={Curriculum learning},
  author={Bengio, Yoshua and Louradour, J{\'e}r{\^o}me and Collobert, Ronan and Weston, Jason},
  booktitle={Proceedings of the 26th annual international conference on machine learning},
  pages={41--48},
  year={2009}
}

@article{xiong2018guided,
  title={Guided policy search for sequential multitask learning},
  author={Xiong, Fangzhou and Sun, Biao and Yang, Xu and Qiao, Hong and Huang, Kaizhu and Hussain, Amir and Liu, Zhiyong},
  journal={IEEE Transactions on Systems, Man, and Cybernetics: Systems},
  volume={49},
  number={1},
  pages={216--226},
  year={2018},
  publisher={IEEE}
}

@article{zhang2017survey,
  title={A survey on multi-task learning},
  author={Zhang, Yu and Yang, Qiang},
  journal={arXiv preprint arXiv:1707.08114},
  year={2017}
}

@article{weiss2016survey,
  title={A survey of transfer learning},
  author={Weiss, Karl and Khoshgoftaar, Taghi M and Wang, DingDing},
  journal={Journal of Big data},
  volume={3},
  number={1},
  pages={1--40},
  year={2016},
  publisher={SpringerOpen}
}

@article{pan2009survey,
  title={A survey on transfer learning},
  author={Pan, Sinno Jialin and Yang, Qiang},
  journal={IEEE Transactions on knowledge and data engineering},
  volume={22},
  number={10},
  pages={1345--1359},
  year={2009},
  publisher={IEEE}
}

@incollection{torrey2010transfer,
  title={Transfer learning},
  author={Torrey, Lisa and Shavlik, Jude},
  booktitle={Handbook of research on machine learning applications and trends: algorithms, methods, and techniques},
  pages={242--264},
  year={2010},
  publisher={IGI global}
}

@inproceedings{ying2018transfer,
  title={Transfer learning via learning to transfer},
  author={Ying, Wei and Zhang, Yu and Huang, Junzhou and Yang, Qiang},
  booktitle={International conference on machine learning},
  pages={5085--5094},
  year={2018},
  organization={PMLR}
}

@inproceedings{bengio2012deep,
  title={Deep learning of representations for unsupervised and transfer learning},
  author={Bengio, Yoshua},
  booktitle={Proceedings of ICML workshop on unsupervised and transfer learning},
  pages={17--36},
  year={2012},
  organization={JMLR Workshop and Conference Proceedings}
}

@article{zhuang2020comprehensive,
  title={A comprehensive survey on transfer learning},
  author={Zhuang, Fuzhen and Qi, Zhiyuan and Duan, Keyu and Xi, Dongbo and Zhu, Yongchun and Zhu, Hengshu and Xiong, Hui and He, Qing},
  journal={Proceedings of the IEEE},
  volume={109},
  number={1},
  pages={43--76},
  year={2020},
  publisher={IEEE}
}

@inproceedings{tan2018survey,
  title={A survey on deep transfer learning},
  author={Tan, Chuanqi and Sun, Fuchun and Kong, Tao and Zhang, Wenchang and Yang, Chao and Liu, Chunfang},
  booktitle={International conference on artificial neural networks},
  pages={270--279},
  year={2018},
  organization={Springer}
}

@inproceedings{dai2009eigentransfer,
  title={Eigentransfer: a unified framework for transfer learning},
  author={Dai, Wenyuan and Jin, Ou and Xue, Gui-Rong and Yang, Qiang and Yu, Yong},
  booktitle={Proceedings of the 26th Annual International Conference on Machine Learning},
  pages={193--200},
  year={2009}
}

@inproceedings{zamir2018taskonomy,
  title={Taskonomy: Disentangling task transfer learning},
  author={Zamir, Amir R and Sax, Alexander and Shen, William and Guibas, Leonidas J and Malik, Jitendra and Savarese, Silvio},
  booktitle={Proceedings of the IEEE conference on computer vision and pattern recognition},
  pages={3712--3722},
  year={2018}
}

@inproceedings{stickland2019bert,
  title={Bert and pals: Projected attention layers for efficient adaptation in multi-task learning},
  author={Stickland, Asa Cooper and Murray, Iain},
  booktitle={International Conference on Machine Learning},
  pages={5986--5995},
  year={2019},
  organization={PMLR}
}

@inproceedings{houlsby2019parameter,
  title={Parameter-efficient transfer learning for NLP},
  author={Houlsby, Neil and Giurgiu, Andrei and Jastrzebski, Stanislaw and Morrone, Bruna and De Laroussilhe, Quentin and Gesmundo, Andrea and Attariyan, Mona and Gelly, Sylvain},
  booktitle={International Conference on Machine Learning},
  pages={2790--2799},
  year={2019},
  organization={PMLR}
}

@article{schmidt2019recent,
  title={Recent advances and applications of machine learning in solid-state materials science},
  author={Schmidt, Jonathan and Marques, M{\'a}rio RG and Botti, Silvana and Marques, Miguel AL},
  journal={npj Computational Materials},
  volume={5},
  number={1},
  pages={1--36},
  year={2019},
  publisher={Nature Publishing Group}
}

@article{bhuvaneswari2021deep,
  title={Deep learning for material synthesis and manufacturing systems: a review},
  author={Bhuvaneswari, V and Priyadharshini, M and Deepa, C and Balaji, D and Rajeshkumar, L and Ramesh, M},
  journal={Materials Today: Proceedings},
  year={2021},
  publisher={Elsevier}
}

@article{mirhoseini2021graph,
  title={A graph placement methodology for fast chip design},
  author={Mirhoseini, Azalia and Goldie, Anna and Yazgan, Mustafa and Jiang, Joe Wenjie and Songhori, Ebrahim and Wang, Shen and Lee, Young-Joon and Johnson, Eric and Pathak, Omkar and Nazi, Azade and others},
  journal={Nature},
  volume={594},
  number={7862},
  pages={207--212},
  year={2021},
  publisher={Nature Publishing Group}
}

@article{devlin2018bert,
  title={Bert: Pre-training of deep bidirectional transformers for language understanding},
  author={Devlin, Jacob and Chang, Ming-Wei and Lee, Kenton and Toutanova, Kristina},
  journal={arXiv preprint arXiv:1810.04805},
  year={2018}
}

@article{brown2020language,
  title={Language models are few-shot learners},
  author={Brown, Tom B and Mann, Benjamin and Ryder, Nick and Subbiah, Melanie and Kaplan, Jared and Dhariwal, Prafulla and Neelakantan, Arvind and Shyam, Pranav and Sastry, Girish and Askell, Amanda and others},
  journal={arXiv preprint arXiv:2005.14165},
  year={2020}
}

@article{silver2017mastering,
  title={Mastering the game of go without human knowledge},
  author={Silver, David and Schrittwieser, Julian and Simonyan, Karen and Antonoglou, Ioannis and Huang, Aja and Guez, Arthur and Hubert, Thomas and Baker, Lucas and Lai, Matthew and Bolton, Adrian and others},
  journal={nature},
  volume={550},
  number={7676},
  pages={354--359},
  year={2017},
  publisher={Nature Publishing Group}
}

@inproceedings{he2016deep,
  title={Deep residual learning for image recognition},
  author={He, Kaiming and Zhang, Xiangyu and Ren, Shaoqing and Sun, Jian},
  booktitle={Proceedings of the IEEE conference on computer vision and pattern recognition},
  pages={770--778},
  year={2016}
}

@article{krizhevsky2017imagenet,
  title={Imagenet classification with deep convolutional neural networks},
  author={Krizhevsky, Alex and Sutskever, Ilya and Hinton, Geoffrey E},
  journal={Communications of the ACM},
  volume={60},
  number={6},
  pages={84--90},
  year={2017},
  publisher={ACM New York, NY, USA}
}

@article{zhang2020resnest,
  title={Resnest: Split-attention networks},
  author={Zhang, Hang and Wu, Chongruo and Zhang, Zhongyue and Zhu, Yi and Zhang, Zhi and Lin, Haibin and Sun, Yue and He, Tong and Mueller, Jonas and Manmatha, R and others},
  journal={arXiv preprint arXiv:2004.08955},
  year={2020}
}

@inproceedings{vaswani2017attention,
  title={Attention is all you need},
  author={Vaswani, Ashish and Shazeer, Noam and Parmar, Niki and Uszkoreit, Jakob and Jones, Llion and Gomez, Aidan N and Kaiser, {\L}ukasz and Polosukhin, Illia},
  booktitle={Advances in neural information processing systems},
  pages={5998--6008},
  year={2017}
}

@article{bahdanau2014neural,
  title={Neural machine translation by jointly learning to align and translate},
  author={Bahdanau, Dzmitry and Cho, Kyunghyun and Bengio, Yoshua},
  journal={arXiv preprint arXiv:1409.0473},
  year={2014}
}

@article{miller2016key,
  title={Key-value memory networks for directly reading documents},
  author={Miller, Alexander and Fisch, Adam and Dodge, Jesse and Karimi, Amir-Hossein and Bordes, Antoine and Weston, Jason},
  journal={arXiv preprint arXiv:1606.03126},
  year={2016}
}

@article{sodhani2020toward,
  title={Toward training recurrent neural networks for lifelong learning},
  author={Sodhani, Shagun and Chandar, Sarath and Bengio, Yoshua},
  journal={Neural computation},
  volume={32},
  number={1},
  pages={1--35},
  year={2020},
  publisher={MIT Press}
}

@article{masana2020class,
  title={Class-incremental learning: survey and performance evaluation},
  author={Masana, Marc and Liu, Xialei and Twardowski, Bartlomiej and Menta, Mikel and Bagdanov, Andrew D and van de Weijer, Joost},
  journal={arXiv preprint arXiv:2010.15277},
  year={2020}
}

@article{rajpurkar2016squad,
  title={Squad: 100,000+ questions for machine comprehension of text},
  author={Rajpurkar, Pranav and Zhang, Jian and Lopyrev, Konstantin and Liang, Percy},
  journal={arXiv preprint arXiv:1606.05250},
  year={2016}
}

@incollection{thrun1996explanation,
  title={Explanation-based neural network learning},
  author={Thrun, Sebastian},
  booktitle={Explanation-Based Neural Network Learning},
  pages={19--48},
  year={1996},
  publisher={Springer}
}

@inproceedings{xie2017aggregated,
  title={Aggregated residual transformations for deep neural networks},
  author={Xie, Saining and Girshick, Ross and Doll{\'a}r, Piotr and Tu, Zhuowen and He, Kaiming},
  booktitle={Proceedings of the IEEE conference on computer vision and pattern recognition},
  pages={1492--1500},
  year={2017}
}

@article{zhang2018natural,
  title={Natural environment benchmarks for reinforcement learning},
  author={Zhang, Amy and Wu, Yuxin and Pineau, Joelle},
  journal={arXiv preprint arXiv:1811.06032},
  year={2018}
}

@article{zhang2020learning,
  title={Learning Robust State Abstractions for Hidden-Parameter Block MDPs},
  author={Zhang, Amy and Sodhani, Shagun and Khetarpal, Khimya and Pineau, Joelle},
  journal={arXiv preprint arXiv:2007.07206},
  year={2020}
}

@inproceedings{cobbe2020leveraging,
  title={Leveraging procedural generation to benchmark reinforcement learning},
  author={Cobbe, Karl and Hesse, Chris and Hilton, Jacob and Schulman, John},
  booktitle={International conference on machine learning},
  pages={2048--2056},
  year={2020},
  organization={PMLR}
}

@article{wang2021generalizing,
  title={Generalizing to Unseen Domains: A Survey on Domain Generalization},
  author={Wang, Jindong and Lan, Cuiling and Liu, Chang and Ouyang, Yidong and Zeng, Wenjun and Qin, Tao},
  journal={arXiv preprint arXiv:2103.03097},
  year={2021}
}

@article{rosenbaum2017routing,
  title={Routing networks: Adaptive selection of non-linear functions for multi-task learning},
  author={Rosenbaum, Clemens and Klinger, Tim and Riemer, Matthew},
  journal={arXiv preprint arXiv:1711.01239},
  year={2017}
}

@article{narvekar2018learning,
  title={Learning curriculum policies for reinforcement learning},
  author={Narvekar, Sanmit and Stone, Peter},
  journal={arXiv preprint arXiv:1812.00285},
  year={2018}
}

@article{portelas2020automatic,
  title={Automatic curriculum learning for deep rl: A short survey},
  author={Portelas, R{\'e}my and Colas, C{\'e}dric and Weng, Lilian and Hofmann, Katja and Oudeyer, Pierre-Yves},
  journal={arXiv preprint arXiv:2003.04664},
  year={2020}
}

@article{wang2020survey,
  title={A Survey on Curriculum Learning},
  author={Wang, Xin and Chen, Yudong and Zhu, Wenwu},
  journal={arXiv preprint arXiv:2010.13166},
  year={2020}
}

@inproceedings{dong2017multi,
  title={Multi-task curriculum transfer deep learning of clothing attributes},
  author={Dong, Qi and Gong, Shaogang and Zhu, Xiatian},
  booktitle={2017 IEEE Winter Conference on Applications of Computer Vision (WACV)},
  pages={520--529},
  year={2017},
  organization={IEEE}
}

@inproceedings{narvekar2017curriculum,
  title={Curriculum Learning in Reinforcement Learning.},
  author={Narvekar, Sanmit},
  booktitle={IJCAI},
  pages={5195--5196},
  year={2017}
}

@inproceedings{weinshall2018curriculum,
  title={Curriculum learning by transfer learning: Theory and experiments with deep networks},
  author={Weinshall, Daphna and Cohen, Gad and Amir, Dan},
  booktitle={International Conference on Machine Learning},
  pages={5238--5246},
  year={2018},
  organization={PMLR}
}

@article{narvekar2020curriculum,
  title={Curriculum learning for reinforcement learning domains: A framework and survey},
  author={Narvekar, Sanmit and Peng, Bei and Leonetti, Matteo and Sinapov, Jivko and Taylor, Matthew E and Stone, Peter},
  journal={arXiv preprint arXiv:2003.04960},
  year={2020}
}

@inproceedings{sarafianos2017curriculum,
  title={Curriculum learning for multi-task classification of visual attributes},
  author={Sarafianos, Nikolaos and Giannakopoulos, Theodore and Nikou, Christophoros and Kakadiaris, Ioannis A},
  booktitle={Proceedings of the IEEE International Conference on Computer Vision Workshops},
  pages={2608--2615},
  year={2017}
}

@article{murugesan2017self,
  title={Self-paced multitask learning with shared knowledge},
  author={Murugesan, Keerthiram and Carbonell, Jaime},
  journal={arXiv preprint arXiv:1703.00977},
  year={2017}
}

@article{elman1993learning,
  title={Learning and development in neural networks: The importance of starting small},
  author={Elman, Jeffrey L},
  journal={Cognition},
  volume={48},
  number={1},
  pages={71--99},
  year={1993},
  publisher={Elsevier}
}

@article{peterson2004day,
  title={A day of great illumination: BF Skinner's discovery of shaping},
  author={Peterson, Gail B},
  journal={Journal of the experimental analysis of behavior},
  volume={82},
  number={3},
  pages={317--328},
  year={2004},
  publisher={Wiley Online Library}
}

@article{skinner1958reinforcement,
  title={Reinforcement today.},
  author={Skinner, Burrhus F},
  journal={American Psychologist},
  volume={13},
  number={3},
  pages={94},
  year={1958},
  publisher={American Psychological Association}
}

@InProceedings{pmlr-v97-hacohen19a, title = {On The Power of Curriculum Learning in Training Deep Networks}, author = {Hacohen, Guy and Weinshall, Daphna}, booktitle = {Proceedings of the 36th International Conference on Machine Learning}, pages = {2535--2544}, year = {2019}, editor = {Kamalika Chaudhuri and Ruslan Salakhutdinov}, volume = {97}, series = {Proceedings of Machine Learning Research}, month = {09--15 Jun}, publisher = {PMLR}, url = { http://proceedings.mlr.press/v97/hacohen19a.html } }
\end{document}